\documentclass[preprint,12pt,authoryear]{elsarticle}

\usepackage{xcolor}
\usepackage{lineno}
\usepackage[font=small]{caption, subcaption}
\usepackage{amssymb}
\usepackage{amsfonts}
\usepackage{bbm}
\usepackage{amsmath}
\usepackage{adjustbox}
\usepackage{makecell}
\usepackage{booktabs}
\usepackage[hyphens,spaces,obeyspaces]{url}
\usepackage{courier}
\usepackage{pdflscape}

\usepackage{algpseudocode,algorithm,algorithmicx}

\newcommand{\cardinality}[1]{| {#1} |}
\DeclareMathOperator*{\argmax}{arg\,max}

\newcommand\RedactForReview[1]{#1}

\newcommand{\actionWalk}[0]{\texttt{Walk}}
\newcommand{\actionRight}[0]{\texttt{Turn Right}}
\newcommand{\actionLeft}[0]{\texttt{Turn Left}}
\newcommand{\actionLight}[0]{\texttt{Activate Light}}
\newcommand{\actionJump}[0]{\texttt{Jump}}
\newcommand{\actionLegend}[0]{Legend: \texttt{S}: \actionLight, \texttt{W}: \actionWalk, \texttt{J}: \actionJump, \texttt{R}: \actionRight, \texttt{L}: \actionLeft.
}

\journal{Cognition}

\begin{document}

\begin{frontmatter}

\title{Exploring the hierarchical structure of human plans via program generation}

\RedactForReview{

\author[1]{Carlos G. Correa\corref{cor1}}
\ead{cgcorrea@princeton.edu}
\cortext[cor1]{Corresponding author}

\author[2]{Sophia Sanborn}
\author[3,4]{Mark K. Ho}
\author[3]{Frederick Callaway}
\author[1,3]{Nathaniel D. Daw}
\author[3,4]{Thomas L. Griffiths}

\affiliation[1]{organization={Princeton Neuroscience Institute, Princeton University}, country={USA}}
\affiliation[2]{organization={Department of Ophthalmology, Stanford University}, country={USA}}
\affiliation[3]{organization={Department of Psychology, Princeton University}, country={USA}}
\affiliation[4]{organization={Department of Computer Science, Princeton University}, country={USA}}

}

\begin{abstract}
Human behavior is often assumed to be hierarchically structured,
made up of abstract actions that can be decomposed into concrete actions.
However, behavior is typically measured as a sequence of actions, which makes it difficult to infer its hierarchical structure.
In this paper, we explore how people form hierarchically structured plans, using an experimental paradigm with observable hierarchical representations: participants create programs that produce sequences of actions in a language with explicit hierarchical structure.
This task lets us test two well-established principles of human behavior: utility maximization (i.e.~using fewer actions) and minimum description length (MDL; i.e.~having a shorter program).
We find that humans are sensitive to both metrics, but that both accounts fail to predict a qualitative feature of human-created programs, namely that people prefer programs with \textit{reuse} over and above the predictions of MDL.
We formalize this preference for reuse by extending the MDL account into a generative model over programs, modeling hierarchy choice as the induction of a grammar over actions.
Our account can explain the preference for reuse and provides better predictions of human behavior, going beyond simple accounts of compressibility to highlight a principle that guides hierarchical planning.

\end{abstract}

\begin{highlights}
\item A process-tracing paradigm is used to observe hierarchies that support planning.
\item Participants prefer reuse beyond a simple measure of compressibility.
\item We formalize this preference in a model of hierarchy choice as grammar induction.
\item In model comparison, our approach is a better account for behavior than alternatives.
\end{highlights}


\end{frontmatter}

\section{Introduction}

Human behavior has rich hierarchical structure,
in which actions are grouped into abstract, higher-level actions that are used to accomplish tasks \citep{miller1960plans,newell1972human,klir1991architecture}.
However, these internal representations are not observable, so they are typically inferred from behavioral signatures of hierarchy \citep{rosenbaum1983hierarchical}.
Consider the problem of \textit{making tomato sauce for dinner}. The task can be made successively more concrete as \textit{prepare the tomatoes} then \textit{prepare one tomato} then \textit{slice the tomato}. Some of these actions have highly repetitive structure, like preparing a tomato by repeatedly slicing it. While action hierarchies arise naturally from practiced behavior, they can still provide a benefit in novel settings by providing a compact representation for behavior and by radically decreasing the complexity of search for a solution.

What guides the generation of hierarchical plans and action hierarchies, particularly in novel settings?
While past studies try to infer this structure from behavioral signatures of hierarchy,
it is challenging to address this question directly because the hierarchical structure of behavior is not observable.
For example, in the sequence-learning literature, hierarchies are inferred based on elevated response times or error rates when switching between abstract actions \citep{rosenbaum1983hierarchical,verwey1996buffer,acuna2014multifaceted}. Similar methods are applied when investigating hierarchical planning, though additional behavioral attributes can be measured such as the particular choice of action sequence or self-reports about behavioral hierarchy \citep{solway2014optimal,huys_interplay_2015,tomov2020discovery,correa2023humans}.

\begin{figure}[t]
\begin{center}

\includegraphics[width=\textwidth]{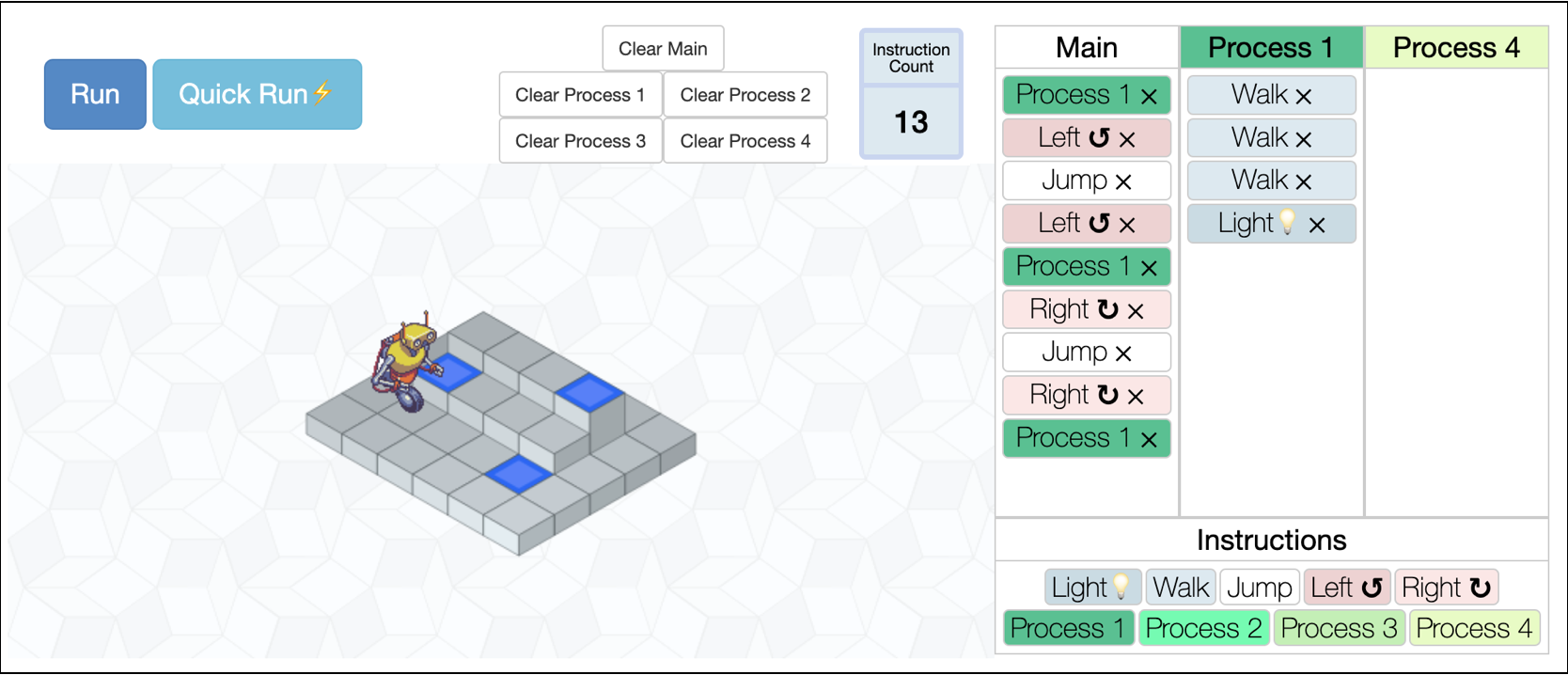}

\caption{
The interface for Lightbot, a process-tracing experiment for complex, hierarchical plans. Participants create programs by dragging instructions from the lower right to define a program. The program is executed by the robot from the initial subroutine (``Main''), and the task is completed when all blue squares in the environment are activated.
An example program that solves the task is shown. The program includes a single subroutine (``Process 1'') with four actions (\actionWalk, \actionWalk, \actionWalk, \actionLight).
Participants can use up to four subroutines (referred to as processes in the experiment), but for brevity only two are pictured.
The \emph{Run} button executes the program with an animation of the robot taking each action. The program is executed without animation with the \emph{Quick Run} button. The program length is displayed as an \emph{Instruction Count}, and the white buttons at the top are used to clear all instructions from a subroutine for ease of program editing.
}
\label{fig:intro-program-writing-ui}
\end{center}
\end{figure}

In order to make hierarchical plans observable, we use a process-tracing paradigm where participants are tasked with creating hierarchically structured plan-like \emph{programs} in an environment called Lightbot.\footnotemark{}
In the task, participants create a program to guide a robot that starts in a specific location (Fig.~\ref{fig:intro-program-writing-ui}) and must visit all unlit lights (blue squares) to activate them.
The programs must contain simple actions (e.g., \actionWalk, \actionLight, \actionRight), but can also have hierarchical structure by reusing sequences of actions as \emph{subroutines}.
These programs serve as an explicit representation of a hierarchical plan, where abstract actions correspond to subroutines.
In previous studies, this experimental paradigm has been used to demonstrate that people prefer programs that are short or can be compressed to be short \citep{sanborn2018representational}.

\footnotetext{Lightbot (\url{https://lightbot.com}) is an educational game developed by Danny Yaroslavski, released as a Flash game in 2008 and a mobile application in 2013.}

This process-tracing paradigm facilitates the comparison of different accounts of hierarchy selection.
For example, given the preference for compressibility in this domain \citep{sanborn2018representational}, a natural theory is that hierarchy selection is based on the principle of minimum description length (MDL; \citealp{chater1999search}). MDL and related information-theoretic frameworks have been previously used to explain hierarchical representations in sequence learning \citep{planton2021theory} and planning \citep{solway2014optimal,maisto2015divide,lai2022chunking}. Another natural theory is that of utility maximization, e.g. solving the problem in the fewest steps.
In this work, we find evidence that people favor both shorter programs and those with fewer actions.
We highlight example programs that these theories cannot account for, but are consistent with a preference for reusing subroutines above and beyond the predictions of either account.
We formalize this preference as a prior belief that subroutine use should be biased towards subroutines that have been used often in the current task.
This prior is central to our computational-level theory \citep{marr1982vision} of human planning as \textit{grammar induction}, an approach drawn from linguistics where it
has been used to model word learning as joint inference of a grammar over words and sentences using those words \citep{goldwater2009bayesian,johnson2007adaptor}.
Inspired by this analogy, we use this approach to model planning as joint inference of a grammar over abstract actions and a plan using those actions.
Here the preference for reuse arises from a clustering prior over the abstract (subroutine-level) actions \citep{johnson2007adaptor}, resembling approaches used to model clustering and reuse in categorization \citep{anderson1991adaptive}, reinforcement learning \citep{gershman2010context}, and causal learning \citep{kemp2010learning}.

In this article, we study how people select hierarchical plans by directly analyzing the programs they create in our process-tracing experiment.
First, we develop our framework of plan inference using grammar induction and detail our generative model over hierarchical plans, showing that our theory is a natural approach to predict qualitative aspects of behavior missed by simpler accounts.
Then, we analyze behavioral data from an online experiment using our process-tracing paradigm.
We examine qualitative examples and perform model comparison, finding that participant programs are best predicted by our grammar induction model.
Analyzing other behavioral signatures unique to our experiment, we also find that people prefer hierarchies that simplify task solution.

\section{Background}

Our approach to modeling complex, hierarchical planning draws on several distinct literatures, which we briefly summarize in this section.
Common approaches for studying hierarchical structure in behavior have used paradigms like sequence learning or frameworks like hierarchical reinforcement learning.
While these studies have often focused on characterizing hierarchy discovery following learning from experience, several have focused on the hierarchies used when planning, as we do in this paper.
Theories of hierarchy discovery require specifying a space of possible hierarchies, which we accomplish through Bayesian program induction.

\subsection{Sequence learning}
A long-standing question is how sequences are learned by humans and other animals.
An influential finding is that sequence memorization is vastly improved when information is grouped into larger \emph{chunks} \citep{miller1956magical}.
Methodologically, mental representations of hierarchical structure are often inferred from behavior on the basis of elevated response times \citep{verwey1996buffer}, errors \citep{rosenbaum1983hierarchical,acuna2014multifaceted}, or decreased speed in motor trajectories \citep{ramkumar2016chunking}.
Recent normative theories of sequence learning explain chunk selection through the optimization of a speed-accuracy trade-off \citep{dezfouli2012habits,wu2023chunking}, MDL \citep{planton2021theory}, and idealized search costs \citep{ramkumar2016chunking}. There are also theories of action segmentation that model reuse by using a similar framework (an Adaptor Grammar; \citealp{johnson2007adaptor}) as our approach \citep{buchsbaum2015inferring}.
Another broad approach to modeling sequence learning is statistical learning \citep{perruchet2006implicit}\textemdash for example, modeling action chunking using a hierarchical non-parametric model of sequence statistics \citep{elteto2022tracking,elteto2023habits}.

\subsection{Hierarchical reinforcement learning}

Hierarchical reinforcement learning adapts the framework of reinforcement learning to a setting with hierarchically structured policies. In one prominent framework, this is formalized by augmenting a task with abstract actions, called ``options'', which each consist of a behavioral policy and conditions for initiation and termination \citep{sutton1999between,stolle2002learning,botvinick2009}.
Much of the research on option discovery has focused on \textit{subgoal}-based options, with
approaches including partitioning a task into subtasks \citep{tomov2020discovery,solway2014optimal,mcnamee2016efficient} or identifying ``bottleneck'' states within the state space (i.e.~states that are commonly passed on the way to the ultimate goal;  \citealp{simsek2009skill}).

Many approaches have been taken to understand how people choose behavioral hierarchies, including some based on task partitioning \citep{solway2014optimal,tomov2020discovery}, policy compression \citep{solway2014optimal,maisto2015divide,lai2022chunking}, and minimizing planning costs \citep{huys_interplay_2015,correa2023humans}. The approach we introduce in this paper based on an Adaptor Grammar \citep{johnson2007adaptor} is similar to a model used to examine planning behavior in prior work \citep{huys_interplay_2015}.
Studies have investigated error-based learning of hierarchical policies, finding evidence in both neural data \citep{ribasfernandes2011neural} and behavioral patterns \citep{eckstein2020computational}.

\subsection{Bayesian program induction}

Bayesian program induction is an approach to inferring programs $\pi$ consistent with some observed data $d$ \citep{goodman2008rational,piantadosi2012bootstrapping,lake2015human,rule2020child}. Concretely, by specifying a \emph{prior} probability distribution over programs $p(\pi)$, and the \emph{likelihood} of the observed data under that program $p(d|\pi)$, Bayes' theorem can be used to define the \emph{posterior} probability of programs conditioned on the data
\begin{align*}
p(\pi | d)
& \propto p(d|\pi) p(\pi). \\
\end{align*}

This approach has been fruitfully used to model many aspects of cognition, including concept learning \citep{goodman2008rational,franken2022algorithms}, number word learning \citep{piantadosi2012bootstrapping}, writing characters with motor actions \citep{lake2015human}, and sequential decision making \citep{maisto2015divide,wingate2011bayesian}.
While the computational demands of inference usually restrict its application to simple domains, modern machine learning methods like neural networks have made it possible to scale these methods to more difficult domains, while still producing interpretable, human-like representations \citep{ellis2021dreamcoder,poesia2023peano}. Another crucial strategy for making inference tractable is library learning \citep{ellis2021dreamcoder,poesia2023peano,zhao2023model}, where program fragments are reused in order to accelerate inference.
Our grammar induction model builds upon this idea.

\subsection{Minimum description length}

Minimum description length (MDL) is an information-theoretic approach
that focuses on minimizing the description length of mental representations and is formally related to Bayesian inference \citep{chater1999search}.
In the setting of sequence learning, an MDL representation is one that provides a minimal, hierarchical compression of the sequence.
This approach has been applied broadly in studying sequence learning.
For example, \citet{planton2021theory} propose that compressed, hierarchical representations of sequences support learning.
\citet{maisto2015divide} incorporate a prior over hierarchical policies related to the description length of plans under those hierarchies.
\citet{lai2022chunking} extend a theory of policy compression to account for the information-theoretic savings from ignoring perceptual information while executing an action chunk.
\citet{solway2014optimal} note their theory identifies hierarchies that minimize the description length of optimal behavior, even though it is primarily formulated as a Bayesian account.

\section{Theories of program creation in Lightbot}

Having introduced these literatures, we now turn to our approach. We draw on ideas from sequence learning and hierarchical reinforcement learning to study the hierarchical structure underlying action sequences, and propose a modeling framework based on principles from program induction and MDL.

Our approach to studying complex, hierarchical plans has two key components. First, we use a process-tracing experimental paradigm in order to make explicit the internal hierarchical representations participants use for planning.
Instead of asking participants to simply solve the problem, we ask participants to create a program that will be interpreted by an agent in the experiment. Critically, this program may be hierarchically structured through the use of subroutines.
Second, we model these plans with a framework based on models of grammar learning from computational linguistics \citep{goldwater2009bayesian,johnson2007adaptor}. In this section, we introduce the experiment and motivate our choice of modeling framework.

Our process-tracing paradigm is based on the popular Lightbot game. The experiment interface and an example program are shown in Fig.~\ref{fig:intro-program-writing-ui}. In this task, research participants control a robot in a gridworld-like environment by supplying a program for it to execute, with the goal of having the robot light all the blue squares.
The program is composed by dragging and dropping instructions in the interface shown in Fig.~\ref{fig:intro-program-writing-ui}. Participants create programs that include five primitive actions (\actionWalk, \actionJump, \actionLeft, \actionRight, \actionLight) as well as calls to subroutines that can be defined and reused.
Subroutines are defined in the same way (by dragging and dropping instructions), so they also consist of actions and subroutine calls, which makes it possible to submit programs that correspond to hierarchically structured plans. For simplicity, the number of subroutines is fixed to four, which means no action must be taken to create an empty subroutine.
When participants submit a program for execution, the robot starts from the initial subroutine and runs instructions in order, so that primitive actions immediately take effect and subroutine calls execute the respective subroutine. Thus, executing a program results in a deterministic sequence of actions, which we refer to as the \emph{execution trace}. A video of program creation and execution is included in the supplementary materials.
We formally define the task, programs, and their execution in the next section (see Section~\ref{sec:modeling-framework}).

\begin{figure}[t]
\begin{center}

{
\setlength{\tabcolsep}{.3em}

\begin{tabular}{ccccc}
\toprule
\makecell{Trace} &                                                                   \makecell{Program} &  \makecell{Step \\ Count} &  \makecell{Program \\ Length} \\
\midrule

\makecell{\includegraphics[scale=.45]{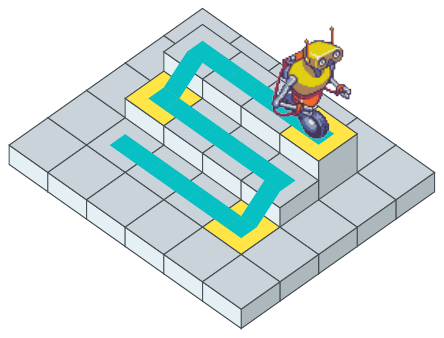}} & \makecell{\includegraphics[scale=.3]{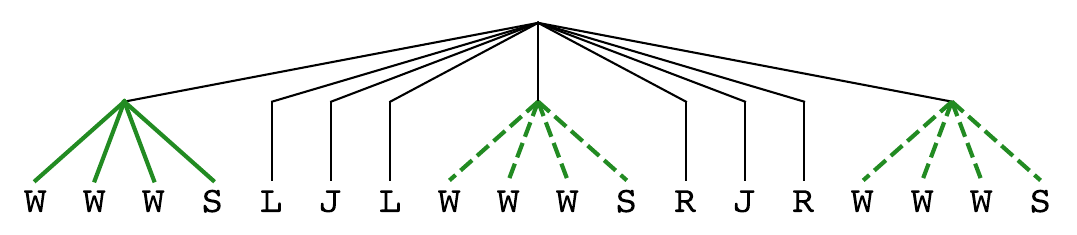}} &                                                      18 &                            13 \\
\makecell{\includegraphics[scale=.45]{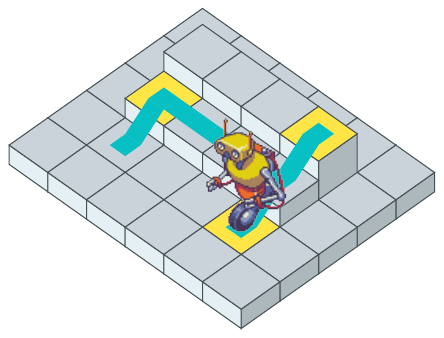}} & \makecell{\includegraphics[scale=.3]{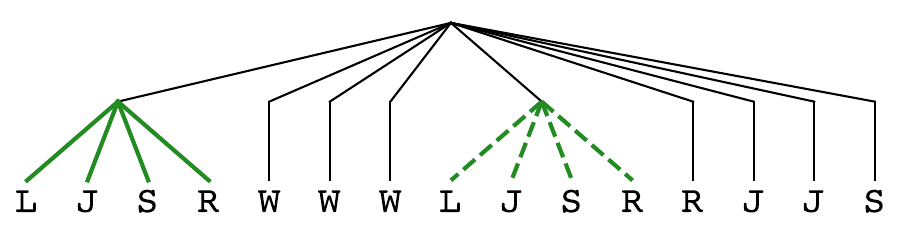}} &                                                       15 &                            13 \\
\bottomrule

\end{tabular}
}

\caption{
Two example programs that solve the task are shown in Fig.~\ref{fig:intro-program-writing-ui}. The top row contains the same program as Fig.~\ref{fig:intro-program-writing-ui}, demonstrating the execution trace, or sequence of resulting actions, in blue.
Table rows correspond to distinct programs. Table columns show a program execution trace, a tree representation of the program, the step count (number of actions) resulting from program execution, and the program length.
The tree representation of a program shows how the sequence of actions relates to the program structure. Actions are executed in sequence, from left to right.
Subroutines are shown with green lines that connect the subroutine call to its constituent instructions. The first use of a subroutine has solid lines, while subsequent uses have dashed lines.
In the experiment, participants only see an animation of the robot, not the execution trace in blue.
\actionLegend
}
\label{fig:intro-program-examples}
\end{center}
\end{figure}

In the example in Fig.~\ref{fig:intro-program-writing-ui}, the robot is shown in the start state. The example program in Fig.~\ref{fig:intro-program-writing-ui} results in an S-shaped execution trace and accomplishes the goal of activating the lights, as shown in the top row of Fig.~\ref{fig:intro-program-examples}. Notably, the program has a subroutine that contains the instructions \actionWalk, \actionWalk, \actionWalk, \actionLight.
If the program were written without a subroutine, the program length would be 18, equivalent to the execution trace length. However, by using this subroutine, the program is much shorter, using only 13 instructions.
In order to make program structure clearer in this article, we display programs as trees, as in the second column of Fig.~\ref{fig:intro-program-examples}, described further in the figure caption.

We formalize the process of planning as Bayesian program induction over hierarchical plans, so the posterior probability of a plan depends on the likelihood that it solves a task as well as the belief assigned to it \emph{a priori}.
By framing planning as an inference problem, we can examine different assumptions about program structure in the form of prior distributions and see how they influence the posterior distribution of programs that solve the task.
This is a notable departure from non-hierarchical planning, which typically finds a sequence of actions that solve a task.
By inferring a hierarchy, this also departs from accounts of hierarchical planning that use fixed hierarchies.
A key component of the posterior is the likelihood that we assign to a program.
Because participants are required to create programs that complete a task, we define this likelihood to require that execution of a program reaches the goal (i.e. having all lights activated).

A prior over programs is the other key component needed to define a posterior. Our modeling approach compares different priors over programs, which we introduce here.
The simplest prior ignores program structure, preferring those that result in shorter execution traces. This model corresponds to a softmax choice rule \citep{luce1959choice} over plans based on the number of resulting actions, similar to standard modeling approaches in reinforcement learning.
An alternative prior might prefer shorter programs, a reasonable approach given that program length directly corresponds to the number of motor actions participants take when writing programs. We quantify program length as the \textit{number of instructions} used in the written description of the program, as opposed to the \textit{number of actions} the program yields in the execution trace. We treat program length as description length, and consider this an MDL prior.

Both of these priors are sensible approaches to modeling plan inference in Lightbot, but are inconsistent with our experimental results (reported in detail below).
For example, in Fig.~\ref{fig:intro-program-examples} the top row features a commonly-written program that contains a single subroutine (\actionWalk, \actionWalk, \actionWalk, \actionLight). This subroutine is used three times and mirrors the repeated environmental structure. The program in the bottom row, written less frequently, uses a subroutine only twice.
Despite experimental participants preferring the top program over the bottom one, it would not be preferred by either of the priors outlined above.

We introduce a theory to explain this by having previous use of a subroutine in the current task inform future use\textemdash in other words, subroutine choice is biased to reuse subroutines. So, the top program in Fig.~\ref{fig:intro-program-examples} is more intuitive because it has greater repeated use of the subroutine.
Our theory of grammar induction is a prior over Lightbot programs, including subroutines, where each use of a subroutine increases the probability that it will be used again.
By contrast, the MDL prior assigns the same probability to programs of equal length, equally preferring flat sequences of actions and highly nested programs.
By accounting for this bias towards reuse, our prior can explain qualitative patterns in the use of subroutines in human behavioral data, as we explore further below.

\section{Modeling framework}
\label{sec:modeling-framework}

\subsection{Task formalism}

We formalize Lightbot as an undiscounted, deterministic Markov Decision Process (MDP), specified by a state set $\mathcal{S}$, initial state $s_0 \in \mathcal{S}$, action set $\mathcal{A}$, transition function $T : \mathcal{S} \times \mathcal{A} \rightarrow \mathcal{S}$, and reward function $R : \mathcal{S} \times \mathcal{S} \rightarrow \mathbb{R}$.
A Lightbot task is defined by a 2-dimensional grid of squares, which each has an associated integer height and type (light square or not).
The state set $\mathcal{S}$ has elements that track the environment state (each light is either active or unactive) and the agent state (a location and orientation).
The action set $\mathcal{A}$ consists of the five primitive actions (\actionWalk, \actionJump, \actionLeft, \actionRight, \actionLight). The goal states $\mathcal{G} \subset \mathcal{S}$ are those where all light squares are activated.

The transition function $T$ is straightforward: \actionLeft{} and \actionRight{} change the agent orientation, by rotating either counterclockwise or clockwise, respectively; \actionLight{} activates a light if present at the agent location. A valid use of \actionWalk{} and \actionJump{} both move the agent forward, changing the location based on the orientation. A valid use of \actionWalk{} requires the current and next location to have matching heights, while a valid use of \actionJump{} requires the next location to either have a height greater by one unit or smaller by any amount. The agent does not move for invalid uses of either action.

The goal in Lightbot is to activate all light squares, so upon reaching a goal state, $g \in \mathcal{G}$, from a non-goal state, $s \notin \mathcal{G}$, the agent is rewarded, $R(s, g)=1$. Otherwise, the agent receives no reward, $R(s, s')=0$. The goal states $g\in\mathcal{G}$ are absorbing, so for all actions $a$, $T(g, a)=g$.

\subsection{Program formalism}

We now introduce notation for the programs that participants submit in our process-tracing paradigm.
We define a \emph{program} $\pi$ as a tuple of subroutines $( \rho^0, \rho^1, \ldots )$, where each subroutine $\rho^i$ consists of a sequence of instructions $(\rho^i_0, \rho^i_1, \ldots)$ and has length $\cardinality{\rho^i}$. Each \emph{instruction} $\rho^i_j$ is either an action in the MDP, so that $\rho^i_j \in \mathcal{A}$, or a \emph{subroutine call} $\rho^i_j \in \{ \rho^1, \rho^2, \ldots \}$.
The program length is the length of all subroutines, $\cardinality{\pi} = \sum_i \cardinality{\rho^i}$.

To execute a program, we start at an initial state $s_0 \in \mathcal{S}$ and begin executing the initial subroutine $\rho^0$.
Executing a subroutine $\rho^i$ in turn requires sequentially executing its constituent instructions $\rho^i_j$. When the current instruction is an action, so $\rho^i_j \in \mathcal{A}$, the action is taken, resulting in a state transition, so $s_{t+1}=T(s_t, a_t=\rho^i_j)$ and $t=t+1$, and a corresponding reward, $R(s_t, s_{t+1})$. Otherwise, the instruction $\rho^i_j$ is a subroutine call, which does not directly result in a state transition but goes on to execute the referenced subroutine. When execution of the referenced subroutine is completed, execution returns to the original subroutine that made the call and continues on to the next instruction.
To match the experimental interface, recursive calls are permitted however $\rho^0$ is excluded from the list of valid subroutine calls.
The execution of a program can be visualized as a tree (Fig.~\ref{fig:intro-program-examples}), where leaf nodes correspond to executed actions, in order from left to right, and internal nodes correspond to subroutine calls.

Programs are executed until they reach a goal state, which results in an execution trace $\tau(\pi) = (a_0, a_1, \ldots, a_{T-1})$ and state trajectory $(s_0, s_1, \ldots, s_T)$, where the final state is a goal, $s_T \in \mathcal{G}$.
This constraint ensures that programs that reach a goal always halt, even if the program is recursive.
The value of a program is $V(\pi) = \sum_t R(s_t, s_{t+1})$. Since there is no reward until the goal is reached,
$V(\pi)=1$ when a program reaches the goal and $V(\pi)=0$ otherwise.
An algorithm for program evaluation is included in the appendix (see Alg.~\ref{alg:recur-exec-new}).

\subsection{Inference}
\label{sec:inference}

\newcommand{\posterior}[1][]{p(\pi \mid \programOptimal#1)}
\newcommand{\likelihood}[1][\pi]{p(\programOptimal \mid #1)}
\newcommand{\prior}[1][\pi]{p(#1)}
\newcommand{\programOptimalVar}{\Omega}
\newcommand{\programOptimal}{\programOptimalVar = 1}

We formulate the choice of programs as the result of Bayesian program induction, adapted to this sequential decision setting by recasting planning as inference \citep{levine2018rlinference,wingate2011bayesian,toussaint2006probabilistic}. Concretely, we define a posterior distribution over programs that combines a prior distribution over programs $\prior$ with a likelihood that indicates whether a program solves the task.

\subsubsection{Inferring programs}

We first introduce a binary random variable $\programOptimalVar$ so that $\programOptimal$ means a program solves the task. We condition on programs that solve the task in order to define a posterior distribution over programs, $\posterior$. Applying Bayes' theorem, we can define this posterior in terms of a likelihood that reflects whether a program solves the task, $\likelihood$, and a prior over programs, $\prior$,
\begin{align*}
\posterior &= \frac{\likelihood \prior}{Z} \\
\end{align*}
where $Z$ is a normalizing constant $Z=\sum_{\pi'} \likelihood[\pi'] \prior[\pi']$,  with $\pi'$ ranging over all possible programs.

We define the likelihood to identify programs that solve the task, or equivalently, reach the goal. Since the value of a program is an indicator of reaching the goal, we can simply define our likelihood as
$$
\likelihood = V(\pi) =
\begin{cases}
1 & \text{ if $\pi$ reaches the goal} \\
0 & \text{ otherwise} \\
\end{cases}.
$$

A key benefit of this inference-based formulation is that it can integrate prior beliefs about the structure of programs alongside the requirement that a program solves the task.

\subsubsection{Approximate inference}

Computing the normalizing constant $Z$ requires the intractable task of enumerating all programs.
As a tractable alternative, we approximate the normalization constant $Z$ using a large corpus of programs, intended to span a large space of compact programs with short traces.
To generate these programs, we first search for a corpus of the shortest execution traces $(a_0, a_1, \ldots)$ that solve a given task.
Then, we generate programs for each trace by taking each combination of non-trivial subroutines and rewriting the trace, assuming subroutines are used as much as possible.
We describe trace search and program generation in more detail in the appendix.

Given a collection of programs $\Pi$, we can compute an approximation to the normalization constant $Z_{\Pi}$ by summing over the programs in $\Pi$, formally
$$
Z_{\Pi}=\sum_{\pi' \in \Pi} \likelihood[\pi'] \prior[\pi'].
$$
This lets us define an approximate posterior
\begin{equation}
\posterior \approxeq Z_{\Pi}^{-1} \likelihood \prior.
\label{eq:approx_post}
\end{equation}

We note that $Z_{\Pi}$ is a lower bound to $Z$, $Z_{\Pi} < Z$, since $Z$ is a sum of positive terms and $Z_{\Pi}$ is a sum of a subset of those terms.

Our approximation scheme is motivated by the difficulty of generating a large number of programs that solve the task.
In particular, the number of traces grows exponentially with trace length. For example, the number of traces of length 18 (like the trace in the top row of Fig.~\ref{fig:intro-program-examples}) is $\prod^{18} 5 = 5^{18} \approx 3.8 \times 10^{12}$, or nearly 4 trillion.
This large space of possible traces is what led us to our scheme to first identify traces that reach the goal, then to subsequently construct programs consistent with those traces.

Importantly, we are not claiming that people perform this approximation scheme when solving the task.
Our model is a computational-level account \citep{marr1982vision} of program writing, intended to examine which factors guide how people select programs.
By identifying the factors that influence program writing, we hope to facilitate future studies that could explore which algorithms people use to solve this challenging inference problem.

\subsection{Baseline models}

Having outlined our framework for Bayesian program induction, we formalize some simple priors over programs.
As a baseline, we consider a model that minimizes trace length. This prior ignores program structure and penalizes the resulting trace length. Since this is equivalent to having a cost for every step, we refer to it as the \emph{step cost} prior:
$$
\prior \propto \exp \left\{ -\cardinality{\tau(\pi)} \right\}.
$$

Some prominent accounts of task decomposition have incorporated description length into their objectives to influence choice over subgoals. In one account, minimizing description length guides the choice among task decompositions, used to compress fixed, optimal policies \citep{solway2014optimal}. In another account, the description length of plans guides subgoal choice in the form of a prior over subgoals in an inference-based hierarchical planner \citep{maisto2015divide}.

Based on this work, we additionally consider a prior that minimizes the description length of a program, which we refer to as the \emph{MDL} prior. We formalize the description length (DL) as the program length, so $\operatorname{DL}(\pi) = \cardinality{\pi}$. The description length is simple and idealized, assuming that all instructions have the same description length. We use this to define a prior, so that shorter lengths are preferred:
$$
p(\pi) \propto \exp \left\{ -\cardinality{\pi} \right\}.
$$

As noted above, the simple approaches introduced in this section struggle to explain intuitive solutions to Lightbot problems, as in Fig.~\ref{fig:intro-program-examples}. The critical missing piece is that these approaches are insensitive to differences in hierarchical structure observed when holding program length or trace length constant. We fill this explanatory gap with the model we introduce next.

\subsection{Grammar induction}

In this section, we recast hierarchical planning as grammar induction. Here, we propose a generative model in which new subroutines augment a grammar over actions.
The key idea behind our model is that past use of subroutines should inform the probability assigned to future use of subroutines. We formalize this idea using Adaptor Grammars \citep{johnson2007adaptor}.

To start, we define a prior over a single subroutine $\rho^i$ that is a sequence of low-level actions, so that $\rho^i_j \in \mathcal{A}$ for all $j$\textemdash that is, the subroutine does not contain other subroutines.
We let the length of the sequence range dynamically in standard fashion, by having a constant probability that construction of the sequence terminates, $p_{end}$, and sampling actions from a uniform distribution, $p(\rho^i_j=a)=\frac{1}{\cardinality{\mathcal{A}}}$. The overall probability of an action sequence is thus
\begin{equation}
p(\rho^i) = p_{end} p(\rho^i_0=a) \prod_{j=1} (1 - p_{end}) p(\rho^i_j=a) .
\label{eq:sr}
\end{equation}
This constant probability of sequence termination means that action sequence lengths are geometrically-distributed with parameter $p_{end}$.

We augment this prior to generate subroutine calls in addition to actions by (1) sampling a subroutine call with probability $p_{call}$ or an action with probability $1-p_{call}$ and then (2) either generating a new subroutine for the call or reusing one that was previously defined.
Permitting the reuse of subroutines means that the resulting grammar is no longer \textit{context-free}, since the probability of a subroutine varies based on previous subroutine calls. To formalize this kind of long-range dependency, we instead use an \textit{Adaptor Grammar} \citep{johnson2007adaptor,huys_interplay_2015}, which augments a standard context-free grammar by allowing the grammar's productions to be dependent on the history of previous productions. This provides a formal basis for a generative model of reuse, where previous productions can become more probable in the future. We use a Dirichlet Process (DP; \citealp{aldous1985exchangeability}) to adapt subroutine generation for our grammar over Lightbot programs.
We briefly highlight some key aspects of this account of subroutine use:
new subroutines are generated,
subroutines can be reused,
and neither the number of subroutines nor the subroutines themselves are predefined.

The DP is a probability distribution over clusters of data that incorporates a new element by either assigning it to (1) a new cluster or (2) an existing cluster, biased towards larger clusters.
We likewise sample subroutine calls, as either (1) a new subroutine or (2) an existing subroutine, biased towards often-used subroutines.
Formally, having already drawn $n$ subroutine calls to $m$ distinct subroutines with a total of $n_k$ calls to subroutine $k$, we define a distribution over the next subroutine call $z_{n+1}$, given the history of drawn subroutine calls $z_{1:n}$

\begin{equation}
p( z_{n+1} \mid z_{1:n} ) =
\begin{cases}
\frac{\alpha}{n+\alpha} & \text{if } z_{n+1}=m+1\\
\frac{n_k}{n+\alpha} & \text{if } 1 \le z_{n+1} = k \le m
\end{cases}
\label{eq:sr-call}
\end{equation}

The first case deals with a new subroutine call, which requires that a new subroutine be generated. In the second case, an existing subroutine is selected proportionally to how often its been used in the past, $n_k$. When an existing subroutine is selected, it is reused and generation of that subroutine is avoided.
The DP has a single parameter, $\alpha \ge 0$, which is usually referred to as the concentration parameter since it controls the relative probability of creating a subroutine. As $\alpha \to 0$ reuse of an existing subroutine becomes more likely and as $\alpha \to \infty$ creation of a new subroutine becomes more likely. Since the most probable subroutines are those used most often in the past, the DP results in so-called \emph{rich-get-richer} dynamics, where frequently used subroutines are most likely in the future.
The rich-get-richer dynamics of the DP are a key prediction that differentiates our grammar induction account from the alternative priors\textemdash as noted above, this preference for reuse can account for the example in Fig.~\ref{fig:intro-program-examples}.

Having introduced the DP, we can define $p(\rho^i_j=a)$ to handle both low-level actions and subroutines as
\begin{equation}
p(\rho^i_j=a) =
\begin{cases}
(1-p_{call}) \frac{1}{\cardinality{\mathcal{A}}} & \text{ if } a \in \mathcal{A} \\
p_{call} p(z_{n+1}=a|z_{1:n}) & \text{ if } a \in \{\rho^1, \rho^2, \ldots\}. \\
\end{cases}
\label{eq:sr-instruction}
\end{equation}
where we elide the dependence on past subroutine calls to simplify notation.
Along with the above Eq.~\ref{eq:sr}, we can define the prior probability of a Lightbot program as
\begin{equation}
p(\pi=\{\rho^1, \rho^2, \ldots\}) = \prod_i p(\rho^i)
\label{eq:program}
\end{equation}

Putting these pieces together, the grammar induction prior generates sequences of instructions, terminating with probability $p_{end}$ (Eq.~\ref{eq:sr}). Instructions are either subroutine calls with probability $p_{call}$ or actions with probability $1 - p_{call}$ (Eq.~\ref{eq:sr-instruction}).
Subroutine calls are drawn from the Adaptor Grammar based on a DP, with new subroutines generated as an instruction sequence and existing subroutines reused in proportion to the number of past calls (Eq.~\ref{eq:sr-call}). An algorithm for sampling is included in the appendix (see Alg.~\ref{alg:grammar-induction}).

In the MDL account, using a subroutine more times can make a program shorter, so a preference for shorter programs can indirectly lead to a preference for subroutine reuse.
The grammar induction model has an analogous preference because, in general, it assigns higher probability to programs with fewer instructions or subroutines. However, a distinctive prediction of the grammar induction model is the explicit preference to use subroutines based on how often they've been used in the past, which is above and beyond that of the implicit preference that comes from avoiding generation.
While the MDL account only indirectly encourages hierarchical structure, the grammar induction model also does so in a more explicit way by assigning higher probability to hierarchies with greater reuse.

Adaptor grammars have been used to study linguistic phenomena like word segmentation \citep{goldwater2009bayesian} and inference of discourse structure \citep{luong2013parsing}, and have also been extended to study forms of generalization that require greater abstraction, like past tense and derivational morphemes in English \citep{odonnell2015productivity} and in non-linguistic topics like causal concept bootstrapping \citep{zhao2023model}.
Adaptor grammars have also been used to model action sequences, by predicting action segmentation \citep{buchsbaum2015inferring} and planning behavior \citep{huys_interplay_2015}. Adaptor grammars are also formally related to approaches used in probabilistic programming languages to express non-parametric distributions \citep{goodman2008church}.

\section{Experiment}

\subsection{Methods}

\begin{figure}[t]
\centering

\newcommand\exptask[2]{\begin{subfigure}[t]{#2\textwidth}
    \centering
    \caption{}
    \label{fig:#1}
    \includegraphics[scale=.3]{figures/results/#1}
\end{subfigure}}

\exptask{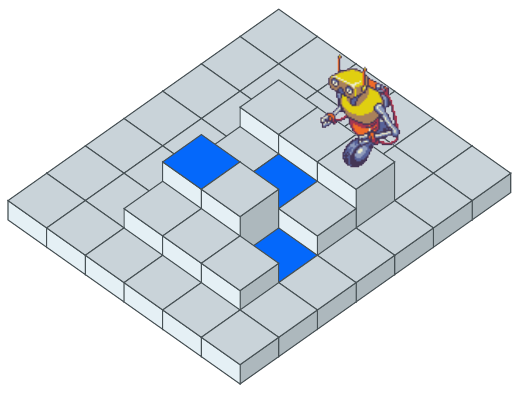}{.24}
\exptask{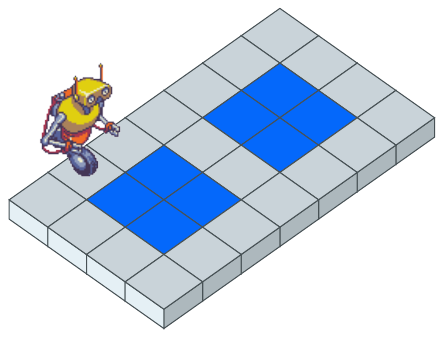}{.24}
\exptask{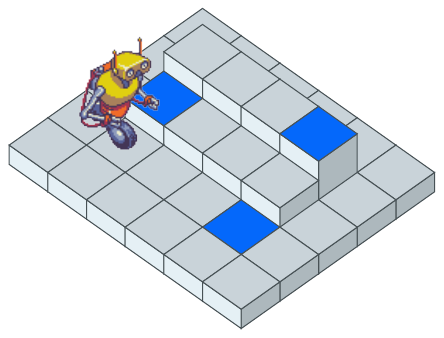}{.24}
\exptask{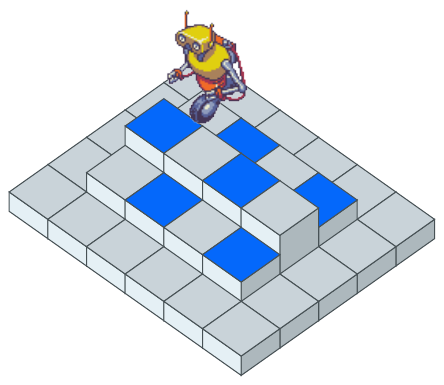}{.24}

\exptask{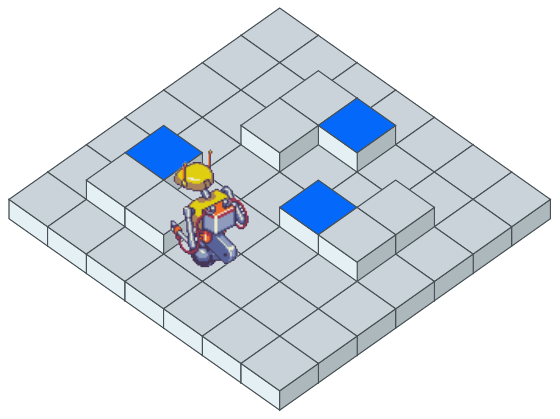}{.24}
\exptask{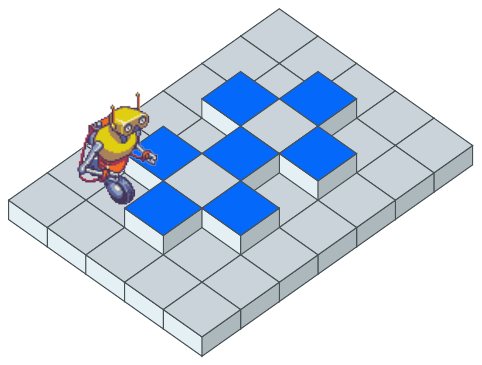}{.24}
\exptask{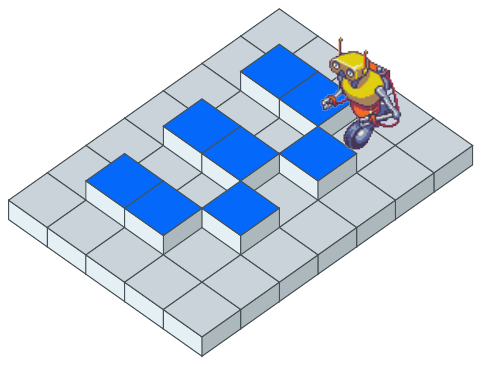}{.24}
\exptask{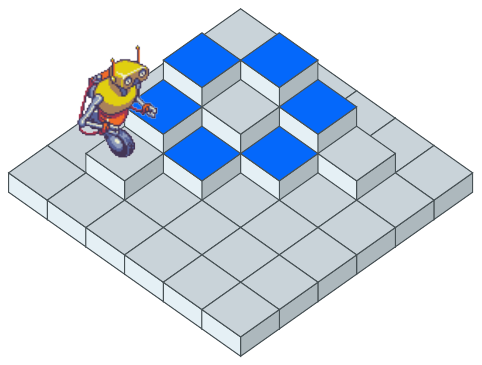}{.24}

\exptask{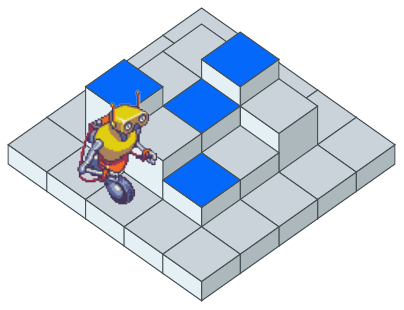}{.24}
\exptask{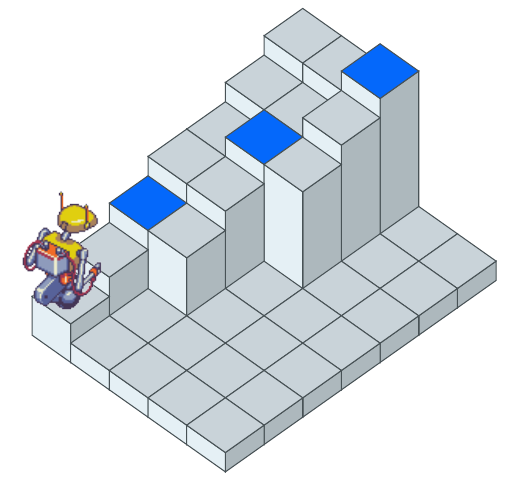}{.24}

\caption{
The tasks participants completed in the experiment.
}
\label{fig:exp-task}
\end{figure}

\subsubsection{Procedure}

Participants were given an extensive tutorial to ensure they learned how to control Lightbot using the five primitive actions, as well as to ensure they understood how to create and use programs. The tutorial described the functionality of each action, included demonstrations, and the opportunity to solve practice problems to ensure understanding. Screenshots of the entirety of the tutorial are included in the appendix.

In the remainder of the experiment, participants wrote programs to solve 10 Lightbot tasks and were incentivized to write short programs to encourage subroutine use. Participants received instructions about the incentive and were shown multiple examples of subroutine use. The program editing interface had a counter showing program length (Fig.~\ref{fig:intro-program-writing-ui}).
Importantly, our instructions focus on describing how the program length influences the received bonus, showing examples of how subroutines can result in both shorter and longer programs than the trajectory.
Participants who wrote the shortest program for a task received the full bonus of \$0.40 and those who wrote a longer program received a proportionally smaller bonus, calculated after the experiment to ensure the average per-task bonus across participants was \$0.20.
Since participants solved 10 tasks, the total bonus ranged from \$0.00 to \$4.00, with an average bonus of \$2.00.
Since the task incentivizes short programs (by encouraging participants to minimize their program length), this could bias participants towards behavior more consistent with the MDL prior. However, because the instructions do not explicitly encourage subroutine reuse, we assume minimal experimentally induced bias towards the grammar induction prior (specifically, its preference for reuse).

After being instructed about the incentives, participants went on to write programs to solve each of 10 tasks (Fig.~\ref{fig:exp-task}), with the program editing interface shown in Fig.~\ref{fig:intro-program-writing-ui}.
When starting a new task, the program editing interface was cleared of the program created for the previous task (i.e. subroutines from past tasks were not available for use).
During each task, participants were given the option to skip the problem and forfeit a bonus for the current problem after 3.5 minutes elapsed.
We recorded the programs submitted by participants, in addition to other measures related to task difficulty, like the number of program evaluations and time elapsed.
In a closing survey, participants were asked about their programming experience and whether they had previously played games similar to the experiment.

\subsubsection{Participants}

The experiment was run on the Prolific platform,
requiring that participants
were English-speakers,
were located in the United States of America,
had not participated in pilot studies for Lightbot,
had an approval rate of $95\%+$,
and had at least 25 past submissions on the platform.
Participants consented to the experimental procedures beforehand, as approved by the Institutional Review Board of \RedactForReview{Princeton University}.
A total of 193 participants (age $M=38.39$ $SD=12.10$, range: 19--86, 72 female, 3 with demographic data missing) completed the entire study,
in an average of 59.55 minutes ($SD=25.63$, range: 17.20--125.56).
171 participants (89\%) were included because they satisfied the inclusion criteria, having no more than 3 college courses related to computer programming.
As noted, participants could skip tasks after 3.5 minutes. 148 participants (87\%) skipped no tasks, 16 participants (9\%) skipped 1 tasks, 3 participants (2\%) skipped 2 tasks, and 4 participants (2\%) skipped 3+ tasks. In the below analysis, we analyze all fully completed tasks, even from participants who may have skipped other tasks.

As noted above, participants were asked about their programming experience.
In response to ``How much experience do you have with computer programming?'' 65\% of participants responded with ``None,'' 24\% responded with ``Between 1 and 3 college courses (or equivalent),'' and 11\% responded with ``More than 3 college courses (or equivalent).''
In response to ``Have you played Lightbot or another similar programming game before?'' 98\% of participants responded with ``No'' and 2\% responded with ``Yes''.
In the appendix, we found that programming experience had little relationship to task performance, focusing on response times, number of program evaluations, and how often participants wrote the most common program.

An important question about our process-tracing paradigm is how it influences participant behavior. We quantify one aspect of this in the appendix, where we find that number of created subroutines and number of program evaluations are positively correlated.

\subsubsection{Parameter fitting}

Given a behavioral dataset of Lightbot programs $\pi$ across all tasks and the number of times they were produced by research participants $n_{\pi}$, we want to optimize for parameters $\theta$ in the model based on the posterior probability they assign to our observed behavior
$$
\mathcal{L}(\theta) = \prod_{\pi} \posterior[, \theta]^{n_{\pi} }.
$$
For models with nuisance parameters $\psi$, we marginalize them out
$$
\mathcal{L}(\theta) = \int_{\psi} \prod_{\pi} \posterior[,\theta,\psi]^{n_{\pi}} p(\psi) d\psi.
$$
In both these equations, parameters are explicitly passed to the posterior for clarity, but were left implicit in Section~\ref{sec:inference}.
We estimate parameters by maximizing the probability assigned to observed behavior, so our objective for the parameters is
$$
\argmax_{\theta} \log \{ \mathcal{L}(\theta) \}
$$
Because of the intractability of computing the posterior, $\posterior$, we instead compute the approximation in Eq.~\ref{eq:approx_post}.

To predict participant programs, we use six models.
One baseline account is a null model of \emph{random choice}, where the prior is uninformative, $\prior \propto 1$, so uniform probability is assigned to programs.
We also test the \emph{step cost}, \emph{MDL}, and \emph{grammar induction} priors introduced above.
Finally, given the efficacy of the step cost prior in explaining behavior, we combine it with the other two priors for the \emph{grammar induction + step cost} and \emph{MDL + step cost} models. These combined models define a new composite prior that is the product of the two original priors.

Parameters are displayed with fitted values in Table~\ref{tab:parameters}, but are briefly summarized here.
A relative weight for the prior is fit, which is used to scale the log prior, $\log \prior$, and are named as follows: $\beta_{\text{StepCost}}$, $\beta_{\text{MDL}}$, $\beta_{\text{GrammarInduction}}$.
For the combined models that include the step cost prior, we also fit a weight of $\beta_{StepCost}$.
The grammar induction model additionally fits the concentration parameter of the DP, $\alpha$, and the probability of a subroutine, $p_{call}$.
In addition, the grammar induction model has the nuisance parameter $\psi=\{p_{end}\}$, which we approximately marginalize for the values
$p_{end} \in \{0.1, 0.2, 0.3, 0.4, 0.5, 0.6, 0.7, 0.8, 0.9\}$,
reflecting a uniform prior.

Parameter fitting was performed once from an initialization using a default value, and 50 times with sampled values.
The parameters with best fit from these 51 parameter fitting executions are used to make predictions below.
Probabilities (e.g., $p_{call}$) were constrained to be in $(0, 1)$ and either fit from an initial default value of $\frac{1}{2}$ or a value sampled uniformly from $(0, 1)$.
All other parameters were constrained to be positive and either fit from an initial default value of 1 or a value sampled from an $Exponential(2)$ distribution.

\subsubsection{Preprocessing of participant programs}

\begin{table}[t]
\caption{
The steps taken in preprocessing participant programs. Shown in order, with descriptions and the number of programs modified by that preprocessing step.
}
\label{tab:preprocessing}
\begin{center}
\begin{tabular}{lp{7cm}ll}
\toprule
& Description & \makecell{Number \\ modified} & \makecell{Percent \\ modified} \\
\midrule
Step 1 & Ensure subroutines are used as often as possible & 221/1668 &     13.25\% \\
Step 2 & Remove instructions that never have an effect\textsuperscript{\textdagger} or are never executed & 241/1668 &     14.45\% \\
Step 3 & Remove subroutines that are never called & 0/1668 & 0.00\% \\
Step 4 & Inline subroutines only called once & 461/1668 &     27.64\% \\
Step 5 & Inline subroutines that only have a single instruction &  34/1668 &      2.04\% \\
Step 6 & Order turns and lights to match constraints on program search (see Appendix) &  67/1668 &      4.02\% \\
Step 7 & Ensure subroutines have a canonical ordering determined by their execution order & 338/1668 &     20.26\% \\
\midrule
& Modified by any step & 868/1668 &     52.04\% \\
\bottomrule
\end{tabular}
\textsuperscript{\textdagger}We remove instructions that are actions $\rho^i_j \in \mathcal{A}$ if for all $t$ where $a_t=\rho^i_j$ the action results in a transition to the same state, so $s_t = s_{t+1} = T(s_t, a_t)$.
\end{center}
\end{table}

In order to facilitate analysis, participant programs are preprocessed in order to ensure they are present in our corpus of generated programs as often as possible.
Preprocessing steps and the number of modified programs are reported in Table~\ref{tab:preprocessing}.
To generate programs for our approximate posterior, we focus on structurally different programs, while avoiding consideration of variation that could arise from certain process-level details (like whether a subroutine was used in every possible location).
This informs how we preprocess participant programs, in order to ensure that participant-generated programs can be matched to model-predicted programs as often as possible.
For example, model-generated programs always make maximal use of subroutines, so we preprocess participant programs to similarly make maximal use of existing subroutines.
While we recognize that these features can provide signatures of the process that participants use to solve the task, we leave further analyses to future studies.
Before preprocessing, only 41\% of participant programs were generated by our methods. After preprocessing, this increased to 78\%. Focusing on programs written by at least 2 participants, preprocessing helped increase this coverage from 82\% to 94\%.

\subsubsection{Data Availability}

\RedactForReview{
Code for the experiment is available at \linebreak\url{https://github.com/cgc/cocosci-lightbot/tree/v0.4}. The data and analysis code are available at \url{https://github.com/cgc/lightbot-grammar-induction}.
}

\subsection{Results}

\subsubsection{Participant programs are inconsistent with alternative accounts}

\begin{figure}[t]
\centering
{
\setlength{\tabcolsep}{.3em}
\begin{tabular}{ccccc}
\toprule

\makecell{trace}                            &  \makecell{\includegraphics[scale=.35]{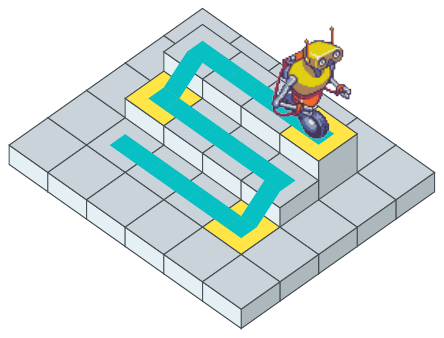}} &  \makecell{\includegraphics[scale=.35]{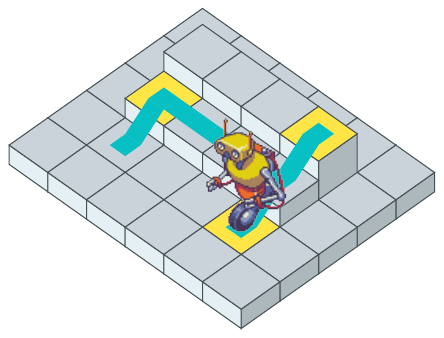}} \\
\midrule
\makecell{program}                          &   \makecell{\includegraphics[scale=0.2]{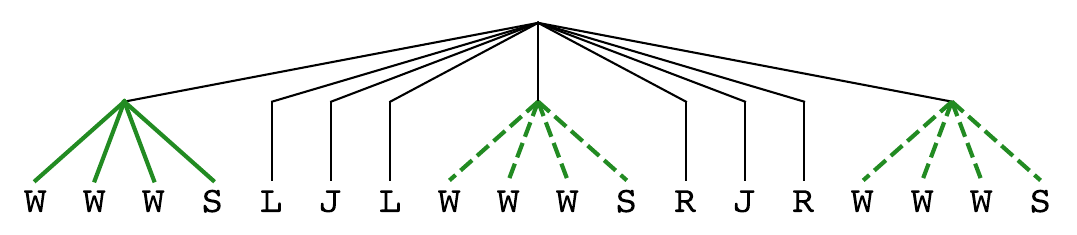}} &   \makecell{\includegraphics[scale=0.2]{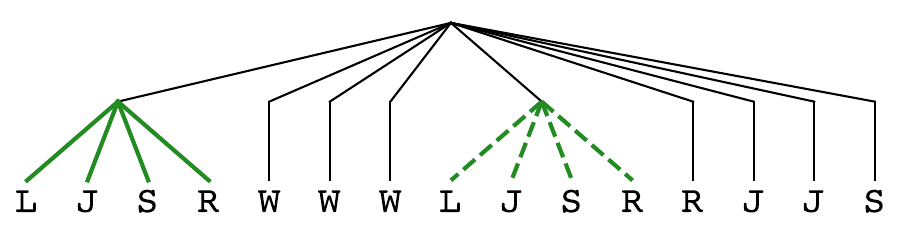}} \\
\midrule
\makecell{participant count}                &                                                                                      59 &                                                                                       1 \\
\midrule
\makecell{step count}                       &                                                                                      18 &                                                                                      15 \\
\midrule
\makecell{program length}                   &                                                                                      13 &                                                                                      13 \\
\midrule
\makecell{grammar induction \\ prior (log)} &                                                                                  -31.97 &                                                                                  -33.17 \\

\bottomrule
\end{tabular}
}
\caption{
Comparing a common participant program (left) to an uncommon one (right) for one task, revisiting the example from Fig.~\ref{fig:intro-program-examples}.
The two programs have equivalent program lengths. The more common program has a larger step count.
The grammar induction prior assigns higher probability to the left program (it is 3.32 times more likely), so it can explain the overall trend of participant preferences.
Subroutines are shown with green lines that connect the subroutine call to its constituent instructions. First use of a subroutine has solid lines, while subsequent uses have dashed lines.
The log of the grammar induction prior is shown, with the values for its three parameters set to $\alpha=1$, $p_{end}=\frac{1}{10}$, and $p_{call}=\frac{1}{2}$.
}
\label{fig:sample-programs0}
\end{figure}

\begin{figure}[t]
\centering
{
\setlength{\tabcolsep}{.3em}
\begin{tabular}{ccccc}
\toprule

\makecell{trace}                            &  \makecell{\includegraphics[scale=.35]{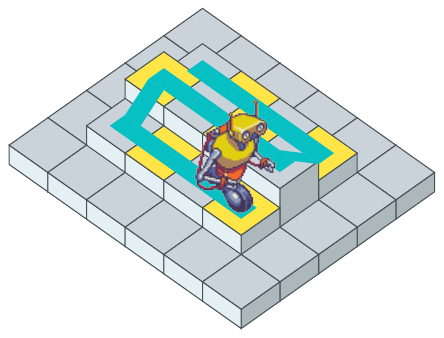}} &  \makecell{\includegraphics[scale=.35]{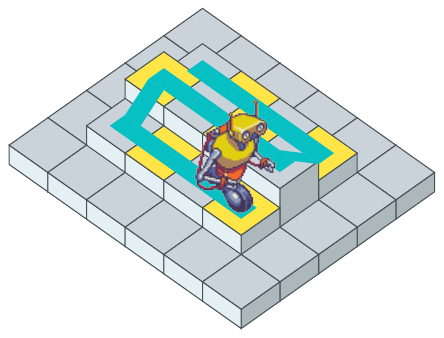}} &  \makecell{\includegraphics[scale=.35]{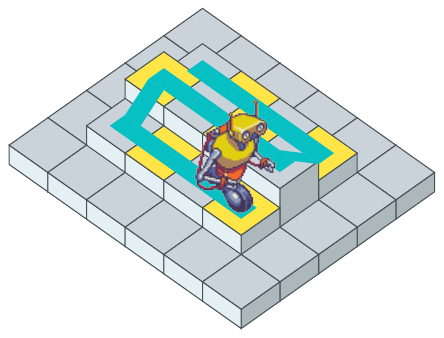}} \\
\midrule
\makecell{program}                          &  \makecell{\includegraphics[scale=0.14]{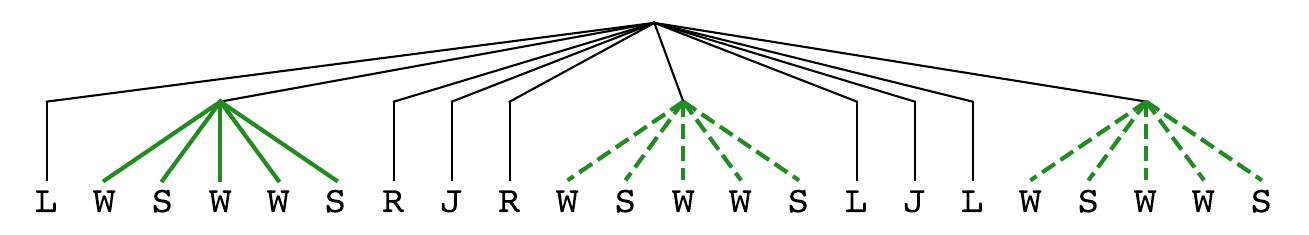}} &  \makecell{\includegraphics[scale=0.14]{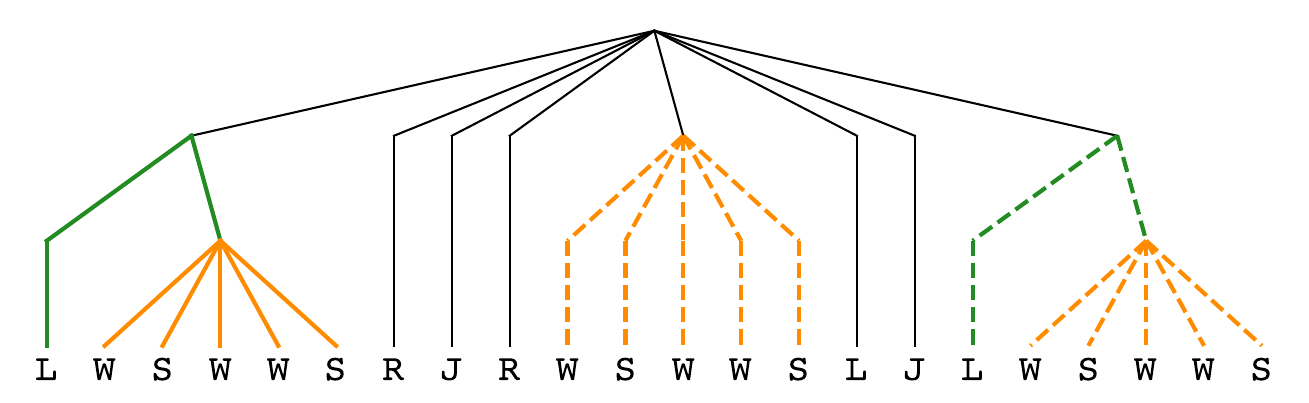}} &  \makecell{\includegraphics[scale=0.14]{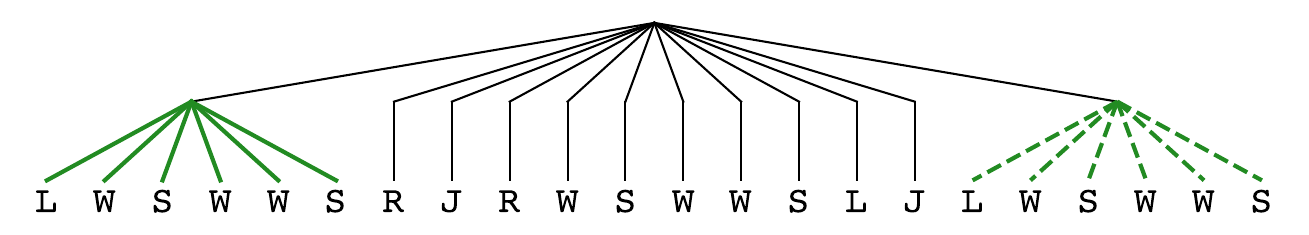}} \\
\midrule
\makecell{part. count}                      &                                                                                          50 &                                                                                           3 &                                                                                           3 \\
\midrule
\makecell{step count}                       &                                                                                          22 &                                                                                          22 &                                                                                          22 \\
\midrule
\makecell{prog. length}                     &                                                                                          15 &                                                                                          15 &                                                                                          18 \\
\midrule
\makecell{grammar induction \\ prior (log)} &                                                                                      -36.78 &                                                                                      -39.45 &                                                                                      -45.21 \\

\bottomrule
\end{tabular}
}
\caption{
Examining a common program (left) to two uncommon programs (middle, right) that all share the same trace.
Here, programs are matched by step count, since they share a trace. Participant choice is not explained by program length, but can be explained by the grammar induction model.
See Fig.~\ref{fig:sample-programs0} for more detail about the figure.
}
\label{fig:sample-programs1}
\end{figure}

\begin{figure}[t]
\centering
{
\setlength{\tabcolsep}{.3em}
\begin{tabular}{ccccc}
\toprule

\makecell{trace}                            &  \makecell{\includegraphics[scale=.35]{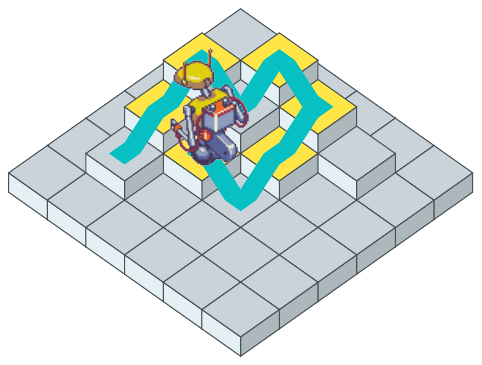}} &  \makecell{\includegraphics[scale=.35]{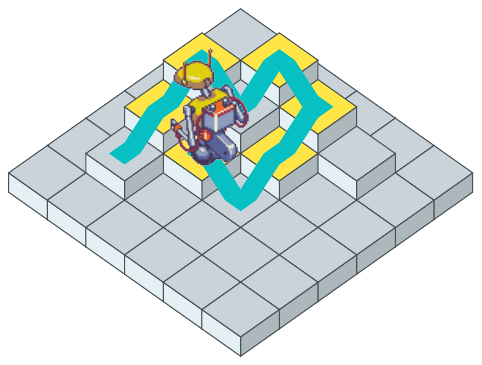}} \\
\midrule
\makecell{program}                          &   \makecell{\includegraphics[scale=0.2]{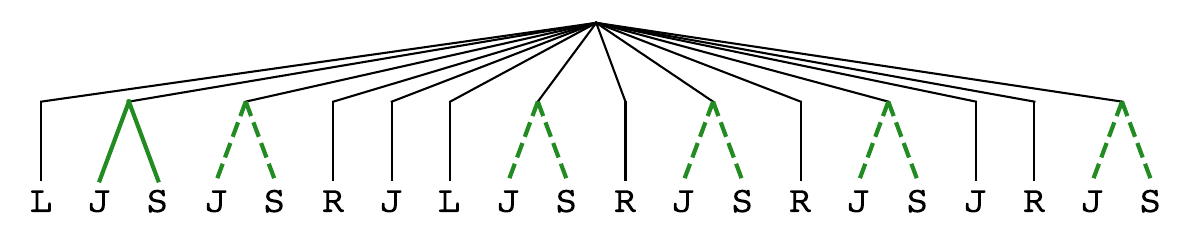}} &   \makecell{\includegraphics[scale=0.2]{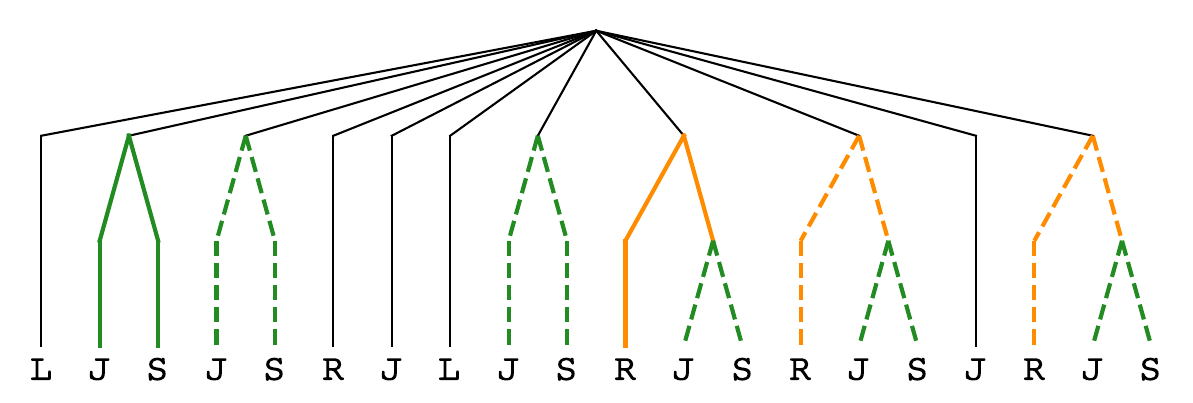}} \\
\midrule
\makecell{participant count}                &                                                                                          19 &                                                                                           3 \\
\midrule
\makecell{step count}                       &                                                                                          20 &                                                                                          20 \\
\midrule
\makecell{program length}                   &                                                                                          16 &                                                                                          15 \\
\midrule
\makecell{grammar induction \\ prior (log)} &                                                                                      -35.06 &                                                                                      -37.49 \\

\bottomrule
\end{tabular}
}
\caption{
A case where a common participant program is inconsistent with the predictions of program length. As in Fig.~\ref{fig:sample-programs1}, the programs have the same trace, so they are matched by step cost. In contrast, now the uncommon figure has a shorter program length. The grammar induction prior can predict this pattern of behavior because of the bias towards reuse.
See Fig.~\ref{fig:sample-programs0} for more detail about the figure.
}
\label{fig:sample-programs2}
\end{figure}

In this section, we qualitatively analyze some example participant programs in Figs.~\ref{fig:sample-programs0}-\ref{fig:sample-programs2}, comparing common programs to other programs that were uncommon.
The programs in these figures are examples that rule out a straightforward account of human choice based on solely optimizing step cost or program length.
In particular, these examples can be viewed as evidence of errors, since participants are generating programs that are in conflict with objectives they might have (i.e. minimizing program length or step cost).
We illustrate how our model aligns with these patterns of choice behavior, but defer statistical comparisons to the next section where parameters are fit and models are compared.
We primarily focus on whether the priors assign higher probability to participant behavior,
but avoid consideration of the magnitudes of probability since our models in the next section fit a parameter that scales the log prior.

We first examine Fig.~\ref{fig:sample-programs0}, which revisits the programs introduced in Fig.~\ref{fig:intro-program-examples}.
Both programs have the same length, but differ in their cost. In particular, the most common program (left) has a large cost, while the uncommon program (right) has a smaller cost. This pattern of choice can't be explained solely on the basis of either program length or cost, but is consistent with the predictions of the grammar induction prior: The common program has greater subroutine use, which has higher probability under the grammar induction prior, due to the preference for reuse in the DP.

We now examine cases where the trace is held constant, but the program varies.
By examining a fixed trace, we can directly compare the structural features of programs while holding action-related features constant, such as step cost.
For example, Fig.~\ref{fig:sample-programs1} shows three programs that generate the most common trace for the problem shown.
The common program (left) has a single subroutine. Notably, this program can be written in several ways since two of the three uses of a subroutine are preceded by the instruction \actionLeft.
This preceding instruction can be incorporated by creating an additional subroutine (middle). This program has the same program length, but is written by fewer participants. Under the grammar induction prior, this program is less probable because it has less reuse.
An intermediate program between these two is also shown (right), where the original subroutine is instead rewritten to include the preceding instruction, reducing the number of places the subroutine can be called.

Fig.~\ref{fig:sample-programs2} shows a similar example of programs that generate the most common trace, for a different task. As in the previous example, the common program (left) has a shorter subroutine that appears more often, while the uncommon program (right) has more subroutines that are used less often. Notably, in this case the common program is actually \emph{longer}. The grammar induction prior can capture this pattern for the same reason as above\textemdash by way of the DP, it prefers greater reuse of a single subroutine.

Taken together, these qualitative results challenge a simple account of choice solely on the basis of cost or program length. This is particularly notable since participants have a monetary incentive to minimize the length of their programs in the experiment.
Distinctive about the grammar induction account is that, above and beyond the length savings associated with the reuse of a subroutine, we expect additional preference for reuse of subroutines, as predicted by the rich-get-richer dynamics of reuse from the DP. Next, we move to a quantitative comparison of models in predicting behavior.

\subsubsection{Predicting participant programs}

\begin{table}[t]
\caption{
Table of fitted models, with log likelihood, BIC, parameter count, and fitted parameters.
}
\label{tab:parameters}
\begin{center}
\begin{tabular}{lcccl}
\toprule
{} & \makecell{Log \\ likelihood} &      BIC &  \makecell{Param. \\ count} &                                                                                                                  Parameters \\
\midrule
\makecell{random choice}                             &                     -21709.7 &  43419.4 &                           0 &                                                                                                              \makecell[l]{} \\
\midrule \makecell{MDL}                              &                     -16787.3 &  33582.0 &                           1 &                                                                                     \makecell[l]{$\beta_{\text{MDL}}=0.98$} \\
\midrule \makecell{grammar induction}                &                     -13482.3 &  26986.9 &                           3 &                                    \makecell[l]{$\alpha=4.23$ \\ $\beta_{\text{GrammarInduction}}=0.48$ \\ $p_{call}=0.12$} \\
\midrule \makecell{step cost}                        &                     -15880.5 &  31768.5 &                           1 &                                                                                \makecell[l]{$\beta_{\text{StepCost}}=1.31$} \\
\midrule \makecell{MDL + step cost}                  &                     -15267.2 &  30549.3 &                           2 &                                                   \makecell[l]{$\beta_{\text{MDL}}=0.53$ \\ $\beta_{\text{StepCost}}=0.85$} \\
\midrule \makecell{grammar induction \\ + step cost} &                     -12825.2 &  25680.1 &                           4 &  \makecell[l]{$\alpha=2.52$ \\ $\beta_{\text{GrammarInduction}}=0.38$ \\ $p_{call}=0.09$ \\ $\beta_{\text{StepCost}}=0.52$} \\
\bottomrule
\end{tabular}
\end{center}
\end{table}

\begin{figure}[t]
\centering

\begin{subfigure}[T]{.40\textwidth}
    \caption{}
    \includegraphics[scale=.5]{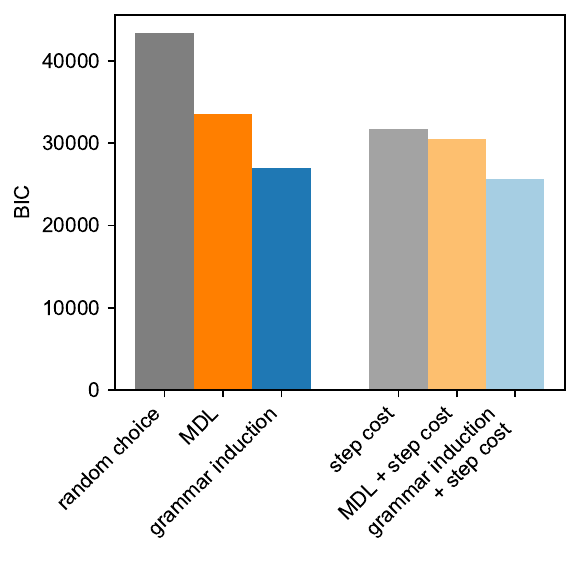}
    \label{fig:model-induction:across-mdp}
\end{subfigure}
\begin{subfigure}[T]{.55\textwidth}
    \caption{}
    \includegraphics[scale=.5]{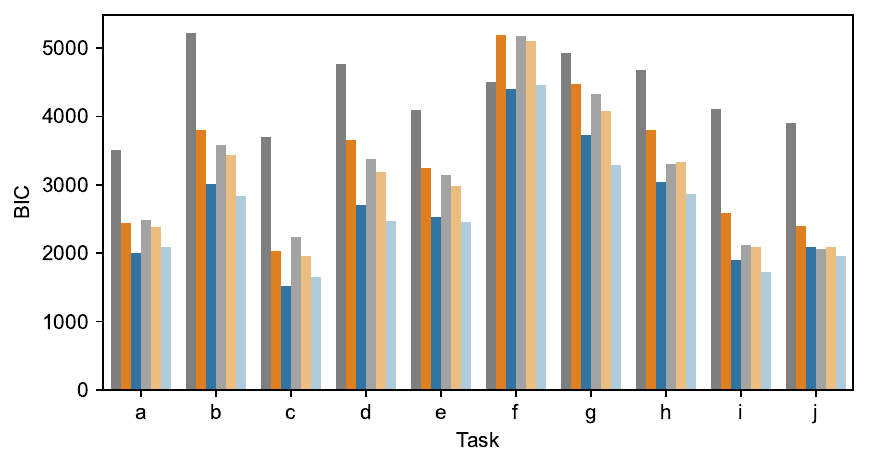}
    \label{fig:model-induction:within-mdp}
\end{subfigure}

\caption{
Model comparison of accounts of participant program creation. a) Plot of BIC of experimental data as predicted by each of the models, after parameter fitting. Models with smaller BIC are a better account of behavior. b) BIC of data under each model, but split by task. Parameters are the same as in a), so they are the best fit for all tasks. The color of the models is also the same as in a). Task letter is a reference to the subfigure in Fig.~\ref{fig:exp-task}.
}
\label{fig:model-induction}
\end{figure}

To more comprehensively examine human behavior, we fit models to the full set of programs people wrote, estimating parameters with maximum likelihood estimation using the procedures described above (see Tab.~\ref{tab:parameters} for fitted parameters).
We compare models by using the Bayesian information criterion (BIC; \citealp{schwarz1978bic}), which penalizes models based on their number of fit parameters.
We first compare the three priors, based on the model BICs shown in Fig.~\ref{fig:model-induction:across-mdp}. All improve on a random choice model (Likelihood-ratio test for step cost: $\chi^2(1)=11658.3$, $p < .001$, MDL: $\chi^2(1)=9844.8$, $p < .001$, grammar induction: $\chi^2(3)=16454.7$, $p < .001$), and the best fit to behavior is the grammar induction model based on the BIC. Surprisingly, the step cost model is a better fit to behavior than the MDL model\textemdash this could mean that participants first identified short traces and then compressed them. In order to control for the influence of this kind of strategy, we also include the combined models described above, in order to compare the MDL and grammar induction models.
Compared to a baseline step cost model, we still find an improvement in fit in the MDL + step cost (Likelihood-ratio test, $\chi^2(1)=1226.6$, $p < .001$) and grammar induction + step cost (Likelihood-ratio test, $\chi^2(3)=6110.6$, $p < .001$) models. Among all tested models, the grammar induction + step cost model is the most predictive of behavior.

While our inclusion of step costs was primarily meant to ensure our results hold after controlling for step costs, we also found that their inclusion improved upon baseline accounts (Likelihood-ratio test for adding step cost to MDL: $\chi^2(1)=3040.1$, $p < .001$; to grammar induction: $\chi^2(1)=1314.1$, $p < .001$).
We explore several qualitative examples demonstrating the importance of step costs in participant programs in the appendix.

Using these parameters (which were fit across all tasks), we examine the model's predictions for each task in Fig.~\ref{fig:model-induction:within-mdp}, 
finding patterns that are broadly consistent with the group summary.
In particular, we find that one of the grammar induction models is still the best predictor for each individual task, compared to the alternative accounts.

We test for the influence of two confounds in the Appendix: 1) Does programming experience have any influence on model comparison? 2) Does program preprocessing have any influence on model comparison? We find that controlling for either produces qualitatively similar results as reported here.

These analyses have focused on predicting behavior for all participants and tasks. However, an important direction for future studies is to better understand differences for individual participants and tasks.
We report several analyses in the appendix that provide interesting directions for future study.
On a per-task basis, we find that task difficulty is related to higher solution variability.
We also develop a theoretical measure that is related to solution variability, based on the idea that competing objectives (i.e. step cost versus grammar induction) could lead to higher variability.
We also quantify individual differences between participants by relating subroutine use and program length, finding variability in how participants trade them off.

\subsubsection{Common programs are easier to write}

\begin{figure}[t]
\centering

\begin{subfigure}[t]{.45\textwidth}
    \includegraphics[width=\textwidth]{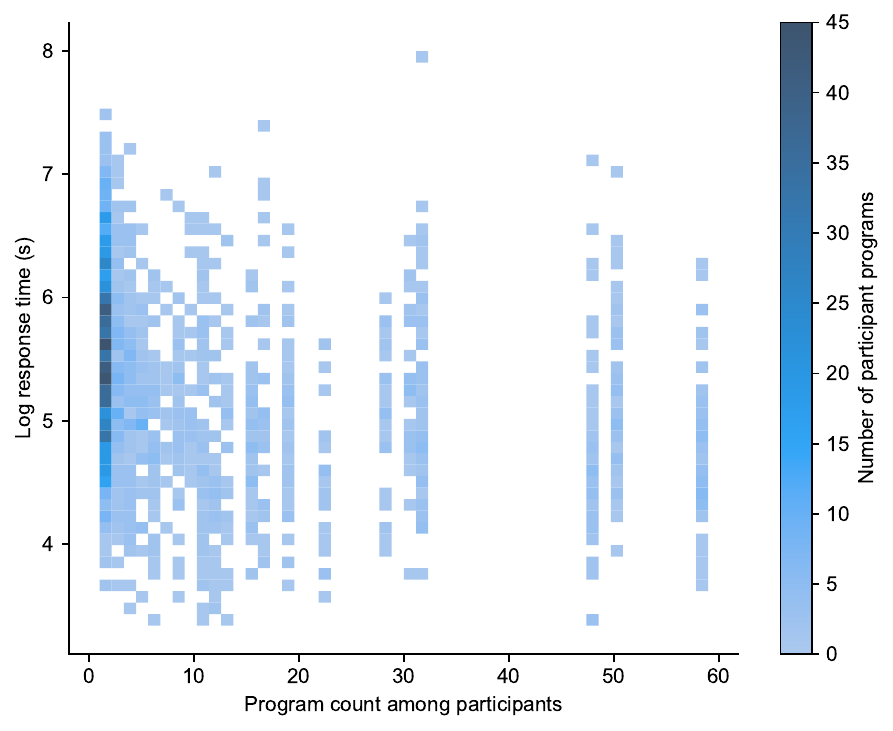}
    \caption{}
    \label{fig:program-ease:time}
\end{subfigure}
\begin{subfigure}[t]{.45\textwidth}
    \includegraphics[width=\textwidth]{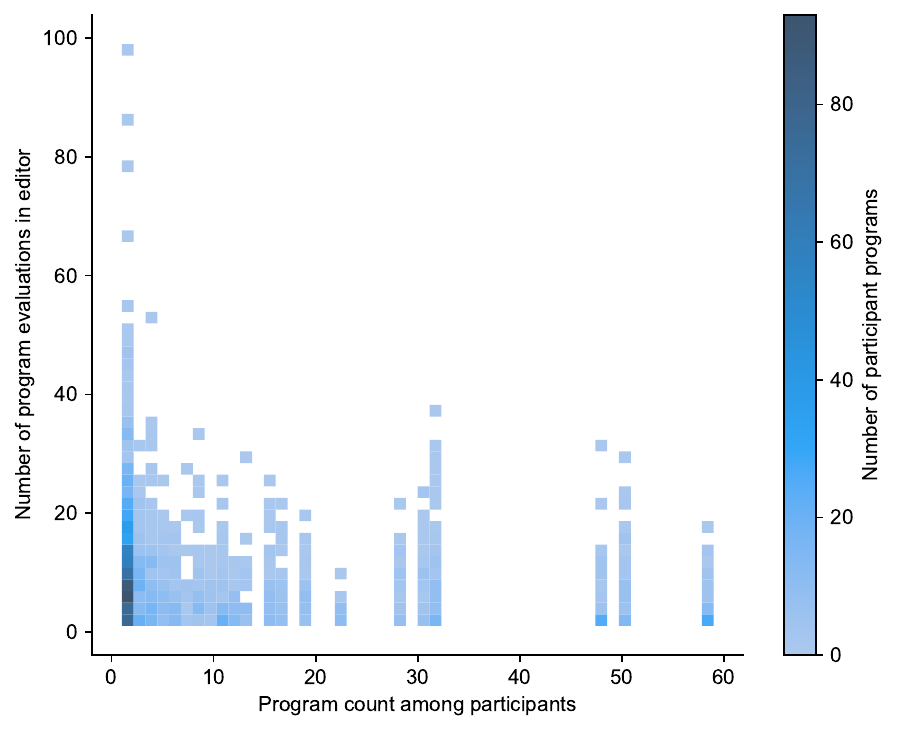}
    \caption{}
    \label{fig:program-ease:runs}
\end{subfigure}

\caption{
Participants write common programs faster and with fewer program evaluations.
Bivariate histogram of task responses for each participant on each task, 
with responses binned along horizontal axis by program count and along vertical axis by either a) log-transformed response times (s) or b) the number of program evaluations.
A related regression analysis is reported in the main text, which uses program count to predict response times and number of program evaluations while controlling for program length and cost, and finds results consistent with these plots.
}
\label{fig:program-ease}
\end{figure}

The following analyses explore how judicious use of hierarchy can make learning and planning easier.
We used hierarchical linear mixed-effects models to test whether people who wrote common programs had an easier time completing the task.
We fit models with \texttt{lmer} and used a baseline model with fixed effects for the intercept, program cost, and program length and random effects for per-participant and per-task intercepts.
To test whether common programs were easier to write, we used a likelihood-ratio test to see if the addition of a regressor for program count would improve model fit when added to a baseline model.

One way to quantify the difficulty of the task is the length of time it takes participants to complete it, a measure that is not specific to our process-tracing paradigm.
We find that participants who wrote more common programs were faster at writing their programs ($\beta=-.008$, $\chi^2(1)=69.73$, $p < .001$, Fig.~\ref{fig:program-ease:time}).
This suggests that people prefer hierarchies that are simpler to use for planning, consistent with prior findings.
Another measure of difficulty, which is uniquely measurable in our paradigm, is the number of times participants test the execution of their program in the editor.
We found that, in the process of writing their programs, those with more common programs executed their program fewer times ($\beta=-.053$, $\chi^2(1)=16.48$, $p < .001$, Fig.~\ref{fig:program-ease:runs}).
In order to rule out any effect of programming experience, we also ran these analyses in the subset of participants with no programming experience. The results are similar as above, and reported in the appendix.
So, people prefer hierarchies that are easier to reason about\textemdash in this case, people writing common programs can avoid explicit program execution in the editor, perhaps by instead relying on mental simulation.

\section{Discussion}

In this paper, we studied the human ability to construct complex, hierarchical plans. In order to do so, we used a process-tracing experimental paradigm to collect data about hierarchical plans and developed a framework for program inference to analyze these plans.
We ran a behavioral experiment and found that our model of grammar induction was better at predicting behavior compared to alternative accounts that simply assumed participants were efficient as measured by program length or trace length.
Key to the grammar induction model is the idea that previous use of subroutines should inform future use. In addition, we used our process-tracing paradigm to examine how the choice of hierarchy can simplify planning. We found that participants who wrote common programs were faster at solving the task and required less use of program execution in the editor.

Our model was motivated by the idea that previous usage should bias reuse. But when is this bias towards reuse sensible?
It might be a natural solution when tasks typically have repeated subtasks, since it might lead to more efficient planning or learning \citep{wen2020efficiency}.
Another explanation could rest on the theoretical observation that the DP is well-approximated by a particular kind of information-theoretic cost \citep{dasgupta2022clustering}\textemdash this would correspond to a representational cost for the distribution over subroutine calls.
One final rationale relates to the idealized computational complexity of search, which is influenced by two factors: the possible choices at each state and the depth of a solution. Abstract actions can accelerate search by making it possible to rapidly reach deeper parts of the search tree. However, they might negatively impact the complexity of search by increasing the number of possible choices. A bias towards reuse could mitigate this increase in complexity by focusing attention on promising subroutines (c.f. \citealp{elteto2023habits}).
What psychological mechanisms might implement a bias towards reuse?
Our theory could be cognitively implemented by some kind of learning or memorization\textemdash it may reflect a simple bias towards recently considered subroutines or instead reflect an online process of inferring appropriate subroutines based on prior history.
We hope future research can explore how these rationalizations and potential mechanisms relate to our account, and how they can be developed into theories of reuse that our model could be quantitatively compared to.
For example, future accounts could develop a generative model of tasks to explore how repeated substructure relates to participant solutions, inspired by the approach in \citet{wen2020efficiency}.

While we focused on reuse in the context of a single task, future research could examine patterns of reuse across tasks. Some existing research has examined this in how people learn program-based causal relations,
showing that a compositional concept is difficult to learn without an appropriate curriculum \citep{rule2018learning}
and the concept learned from experience can be dependent on the sequence of experienced tasks \citep{zhao2023model}.
Lightbot could be used to examine similar questions by permitting the transfer of subroutines between tasks. Having explicit representations of subroutines as they are transferred, adapted, and used in future tasks might provide an exciting process-tracing approach for testing theories of the influence of curriculum and experience on learning.

However, adapting our account to study across-task reuse would introduce issues that we have not yet considered.
In particular, an ever-expanding library of subroutines would grow unwieldy, making the cost of search high since many subroutines would need to be considered at each step.
The rich-get-richer dynamics of our model also pose an issue because a heavily-used subroutine might remain probable long after it is useful.
These issues are addressable by decaying use counts over time, or ensuring that contextual information (like the current task or task state) is used to prioritize consideration of subroutines.
These ideas could be incorporated into our model by generalizing our approach to subroutine sampling to be distance dependent \citep{blei2011distance}, so that subroutine use would be driven by similarities to past uses. Through appropriate definition of similarity between subroutine uses, like temporal recency or task state similarity, this formalism could be used to test complex hypotheses about how time, task, and task state influence the consideration of subroutines.
Other ideas could be drawn from past work, which has identified that
people prioritize possibilities that have been probable or effective in the past \citep{mattar2018prioritized,bear2020what,morris2021generating}.

Since our paradigm is a process-tracing paradigm, we cannot be sure about the relationship between the explicit representation we can measure (the hierarchical plans people submit) and the internal representations people use to act.
Hierarchical structure has often been studied by using paradigms like sequence learning, where participants perform a fixed sequence many times. By contrast, participants never perform the sequence of actions directly in our paradigm, and instead must simulate it mentally or watch it be executed by the robot.
In addition, participants are not trained on a fixed sequence; they instead solve the task a single time and simply submit their program. Because behavior in our experiment is not driven by bottom-up statistics in the same way as sequential learning, our behavioral paradigm might isolate a distinct aspect of behavior, namely structural biases of hierarchical planning.

It is possible that the explicitly hierarchical interface of the task encouraged participants to use hierarchical structure in a way that they would not in real-world planning problems.
While contemporary studies using laboratory tasks have identified behavioral and neural signatures of hierarchical plan representations in navigation tasks \citep{solway2014optimal,huys_interplay_2015,balaguer2016neural},
the structure of Lightbot could encourage subroutine reuse. Although the explicit incentives of the task (both bonus payment and number of clicks necessary) encourage minimizing total program length, the mere existence of explicitly reusable subroutines could push people towards reuse at the expense of longer programs, perhaps through a demand effect.
Testing our account in a more naturalistic setting is thus a critical\textemdash although empirically challenging\textemdash direction for future work.

Our approach is a computational-level account \citep{marr1982vision}, which naturally leads to the question of the algorithmic processes people use to approximately implement our theory, particularly because of the intractability of Bayesian inference. 
We hope that our experimental findings justify further investigation into possible process-level accounts
and behavioral signatures that could be used to distinguish among them.
Many existing approaches to inference explore how varying the parameterization of inference algorithms can lead to a human-like bias. For example, some studies have explored how low-capacity algorithms might recapitulate behavior \citep{daw2007pigeon,lake2020people} or how different proposal distributions explain auto-correlation in participant hypotheses \citep{franken2022algorithms}. These results are often interpreted as evidence that human-like biases might arise from rational use of cognitive resources.
Future research might pursue process models that are more consistent with findings in the sequence learning literature\textemdash for example, it has been observed that larger action sequences might form through concatenation of existing chunks \citep{tosatto2022evolution}, which can be used to guide the model of program creation.
In general, future work could extend our model to investigate how process-level and algorithmic changes can better align with observed behavior.

Our grammar induction prior, as a generative model for programs,
raises the straightforward possibility of an algorithmic implementation based on inference.
In contrast, the MDL account we compare to is not formulated as a probabilistic model, so it requires non-trivial effort to adapt it to an inference-based algorithm.
Either account could be implemented by a process of retrospectively compressing an identified action sequence, by first planning and then compressing. With appropriate weighting, this could form a valid algorithm for the models we have introduced. However, a more interesting direction for the future could examine the bias induced with varied path-finding (e.g., what heuristic is used to guide the search for solutions?) and compression algorithms (e.g., do cognitive constraints influence the set of programs that can be considered?).
More specific to the grammar induction prior are sequential inference algorithms (like sequential Monte Carlo samplers; \citealp{doucet2001sequential}), which suggest a process in which subroutines are generated prospectively, as part of the sequential generation of a program.
This suggests an interesting algorithmic question for future studies, which could look for behavioral indications about whether participants generate hierarchical programs prospectively, or instead retrospectively compress already-identified plans. However, despite the access to representations that our paradigm offers, it seems difficult to distinguish between these two accounts with our reported experiments, since the program participants produce does not immediately indicate whether the trajectory or hierarchy was constructed first.

A related concern that is agnostic to the inference algorithm is the computational cost of evaluating the terms that comprise the unnormalized posterior, namely the prior and likelihood. In particular, the likelihood in our model requires evaluating a program in the task. Evaluating a policy in the general case of stochastic environments requires integrating over all possible outcomes which can be computationally costly, driving people to use heuristics when estimating utilities \citep{lieder2018overrepresentation}, like relying on individual memories of past trials \citep{duncan2016memory}. While participants have extensive training on the effect of instructions in Lightbot and have access to a program simulator, effectively using mental resources and available time requires judicious use of simulation \citep{hamrick2015think,ullman2023resource}. We found that the most common programs participants wrote required less program evaluation, which suggests that the computational cost of evaluating programs may influence how participants choose among programs.
A previous study used the Lightbot domain to examine how execution-related properties of programs influence prior beliefs about programs \citep{ho2018human}\textemdash a natural extension of that analysis could examine whether these computational properties influence the process of search directly, or only by way of influencing program evaluation.
Future work could use explicit representations of hierarchy, like in our experiment, in order to finely probe how differences in hierarchy influence mental simulation.

Hierarchy is a key organizational strategy that humans use to structure their behavior in order to make planning and learning efficient \citep{newell1972human,botvinick2009}, but is difficult to study, requiring indirect measures to infer internal hierarchical representations \citep{rosenbaum1983hierarchical,verwey1996buffer,huys_interplay_2015}.
In this article, we used a process-tracing paradigm to observe the hierarchical representations used to solve the task, which allowed us to identify that people have a bias towards reuse captured by our generative model of grammar induction.
We hope our approach can inspire future efforts to externalize the representations underlying complex mental processes such as planning.

\section{Acknowledgements}

\RedactForReview{
This research was supported by John Templeton Foundation grant 61454 awarded to TLG and NDD (https://www.templeton.org/), U.S. Air Force Office of Scientific Research grant FA 9550-18-1-0077 awarded to TLG 
\linebreak
(https://www.afrl.af.mil/AFOSR/), and U.S. Army Research Office grant ARO W911NF-16-1-0474 awarded to NDD (https://www.arl.army.mil/who-we-are/directorates/aro/). The funders had no role in study design, data collection and analysis, decision to publish, or preparation of the manuscript.
}

\appendix

\setcounter{table}{0}
\setcounter{figure}{0}

\section{Appendix}

\subsection{Algorithms}

In this section, we include an algorithm for program execution (see Alg.~\ref{alg:recur-exec-new}) and a generative algorithm that returns samples from the grammar induction prior (see Alg.~\ref{alg:grammar-induction}).

\renewcommand{\algorithmicrequire}{\textbf{Input:}}
\renewcommand{\algorithmicensure}{\textbf{Output:}}

\begin{algorithm}
\caption{
    A recursive algorithm for program execution.
}\label{alg:recur-exec-new}
\begin{algorithmic}
\Require \\
Current subroutine $\rho^i$ \\
Start time $t$
\Ensure End time $t$
\For{$\rho^i_j$ in $\rho^i$}
\If{$\rho^i_j \in \mathcal{A}$}
    \Comment{Instruction $\rho^i_j$ is the action $a_t$}
    \State $s_{t+1} \gets T(s_t, \rho^i_j)$
    \If{$s_{t+1} \in \mathcal{G}$}
        \State Halt program execution because a goal has been reached
    \EndIf
    \State $t \gets t + 1$
\Else
    \Comment{Instruction $\rho^i_j$ is some subroutine $\rho^k$}
    \State $t \gets recurse(\rho^i_j, t)$
    \Comment{Recursively execute the subroutine}
\EndIf
\EndFor
\end{algorithmic}
\end{algorithm}

\begin{algorithm}
\caption{Algorithm for sampling from the grammar induction prior. For simplicity, the programs generated by this algorithm can have an unlimited number of subroutines, though we only examine programs with four subroutines in the text.}\label{alg:grammar-induction}
\begin{algorithmic}[1]
\Ensure Program $\pi=(\rho^0, \rho^1, \ldots)$
\Function{Instruction}{\null}
    \If{$true \sim Bernoulli(p_{call})$}
        \State $k \sim p( z_{n+1} \mid z_{1:n} )$ \Comment{Sample subroutine, Eq.~\ref{eq:sr-call}}
        \If{$\rho^k$ is not defined} \Comment{Generate if new subroutine}
            \State define $\rho^k$
            \Comment{Defining since recursive program is possible}
            \State $\rho^k \sim $ \Call{Subroutine}{\null}
        \EndIf
        \State \Return $\rho^k$
    \Else

        \State $a \sim Uniform(\mathcal{A})$
        \Comment{Sample from actions $\mathcal{A}$}
        \State \Return $a$
    \EndIf
\EndFunction
\Function{Subroutine}{\null}
    \State Initialize $\rho$ as an empty subroutine
    \State $j \gets 0$
\Repeat
        \State $\rho_j \sim $ \Call{Instruction}{\null}
        \State $j \gets j + 1$
\Until{$true \sim Bernoulli(p_{end})$}
    \State \Return $\rho$
\EndFunction

\State $\rho^0 \sim $ \Call{Subroutine}{\null}
\Comment{Initiate sampling of program}

\end{algorithmic}
\end{algorithm}

\subsection{Trace search}

Our strategy for trace search features three components: 1) find a large corpus of traces 2) by using heuristic search methods, 3) while avoiding trivial traces.

To minimize bias in the programs we generate, we search for the shortest traces, taking a minimum of $m=1000$ traces.
To avoid any effects of tie-breaking, we continue searching to ensure we find all traces of equivalent cost to the $m^{th}$ trace.

We search in the space of traces using a variant of A* search \citep{hart1968formal} that is adapted to find the best $m$ traces and continue searching to avoid tie-breaking effects.
Our algorithm closely resembles m-A* search \citep{flerova2016searching}, which finds the top $m$ solutions for a problem.
To ensure search efficiency, we use a heuristic based on shortest-path lengths.
In particular, while the number of traces grows exponentially in trace length, the number of task states is finite (though exponential in the number of lights), making it reasonable to compute the shortest path to a goal from any state.
So, the heuristic cost function simply returns the shortest path length from the trace's final state to any goal.
By design, the heuristic is monotone (which we also empirically verify),
so this ensures the results are optimal given the assumptions of A* search \citep{russell2021artificial,flerova2016searching}.

In order to accelerate search, we avoid traces with trivial action sequences.
We required traces to consist of actions that resulted in changes to the state, so $T(s_t, a_t) \ne s_t$.
We also excluded certain sequences that are redundant or symmetric: three left turns, three right turns, a right turn followed by or preceding a left turn, and light instructions after turns.

\subsection{Program generation}

We expand an execution trace into many possible programs that generate the trace by rewriting the trace using all combinations of possible subroutines. Subroutines are only considered if they are called in at least two places in the resulting program and contain at least two instructions. To match the experimental interface, programs could only contain at most four subroutines. Trace rewriting uses a greedy algorithm that rewrites to maximize subroutine use, considering one subroutine at a time from longest to shortest.

We add special cases to generate programs consistent with those written by some participants. One case optionally adds post-goal actions to the trace that are consistent with some candidate subroutine. Though post-goal actions are extraneous, when produced by a subroutine, they can still result in an overall reduction in program length. A second case involves recursive subroutines---subroutines that call themselves repeatedly until the goal is reached. We generate these by proposing all trace suffixes as subroutines. A final case combines these: recursive programs with post-goal actions.

Since these programs are used to approximate the posterior, we minimize bias by ensuring that participant programs are in this corpus of programs as much as possible.
We do so by searching for traces in order of ascending trace length, taking a minimum of $n=1000$ traces but continuing to search for traces of equivalent cost to the $n^{th}$ trace. Using A* search to find multiple solutions requires searching over state trajectories, as opposed to simply searching over states. The cost of the $n^{th}$ trace is the maximum trace cost we consider, so participants with programs of greater cost have programs that are not in this set. Using these methods, 78\% of participant programs were generated and 94\% of participant programs written by at least one other participant were generated. Missing programs largely correspond to programs excluded for the purpose of efficiency, as noted above: greater than maximum trace cost, use of actions that result in no state change, use of redundant instruction sequences, and post-goal actions due to a subroutine used only once in the pre-goal program.

\subsection{Examining how step cost influences program creation}

The model that is the best fit to behavior in the text (grammar induction + step cost) incorporates the step cost prior, which minimizes trace length. While the qualitative examples in the main text focused extensively on providing support for the reuse-based grammar induction account, we did not provide examples showing support for the step cost prior.
In this section, we review some examples that show how step costs inform participant choices, leading them to underuse a subroutine.

\begin{landscape}
\begin{figure}[t]
\begin{center}
\begin{tabular}{c|cc|cc|c}
\toprule
\makecell{trace}                            &  \makecell{\includegraphics[scale=.35]{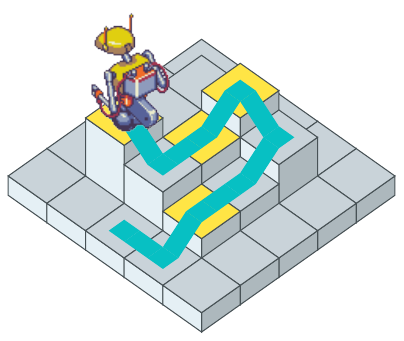}} &  \makecell{\includegraphics[scale=.35]{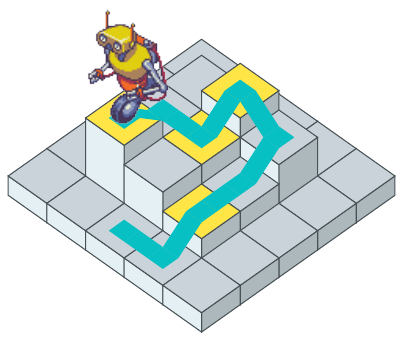}} &  \makecell{\includegraphics[scale=.35]{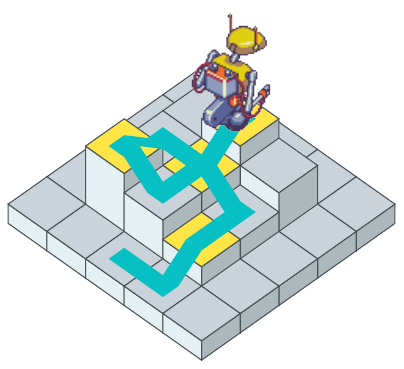}} &  \makecell{\includegraphics[scale=.35]{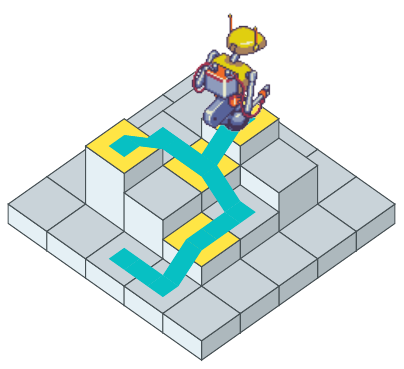}} &  \makecell{\includegraphics[scale=.35]{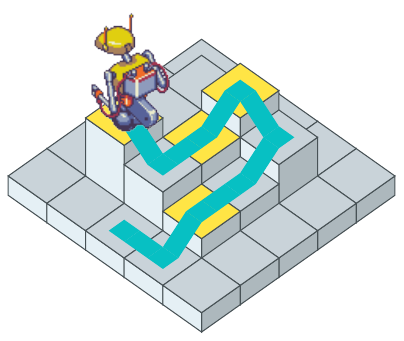}} \\
\midrule
\makecell{program}                          &   \makecell{\includegraphics[scale=0.15]{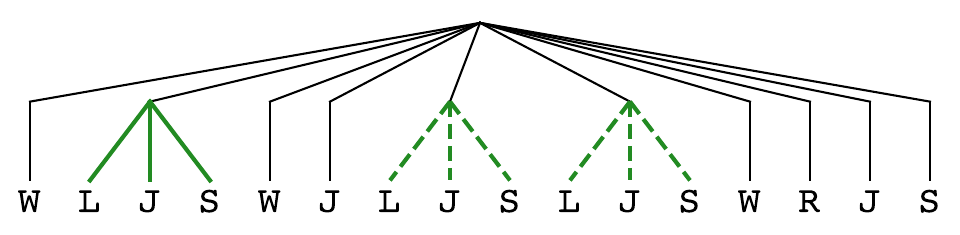}} &   \makecell{\includegraphics[scale=0.15]{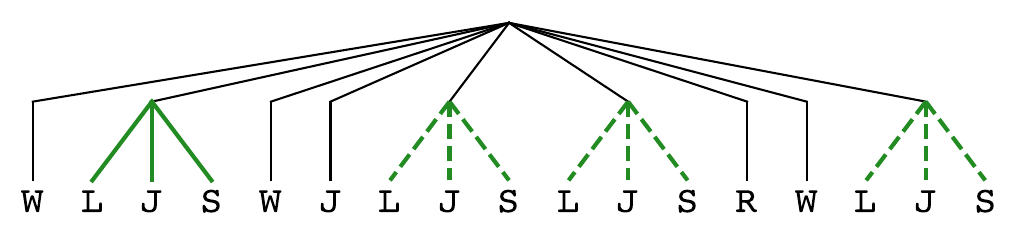}} &   \makecell{\includegraphics[scale=0.15]{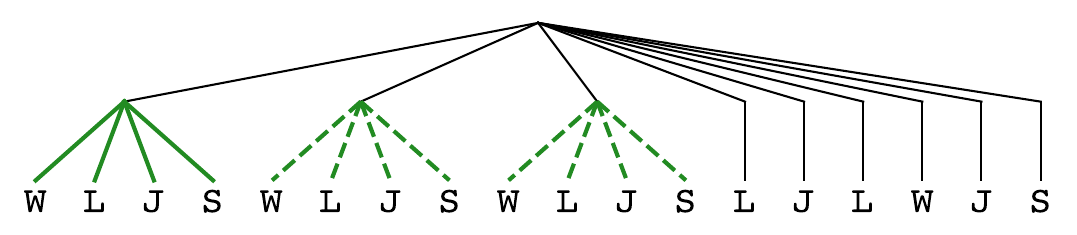}} &   \makecell{\includegraphics[scale=0.15]{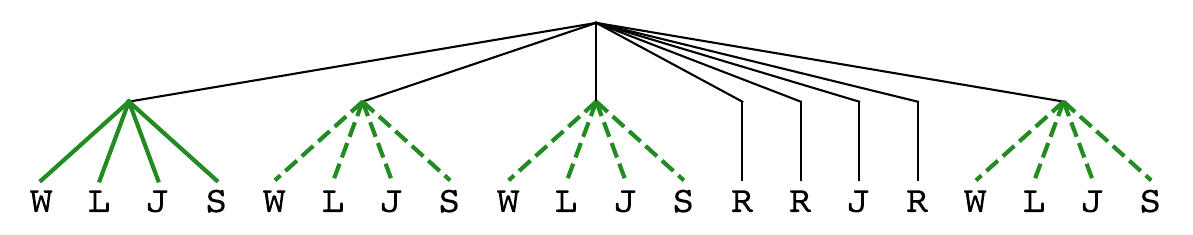}} &   \makecell{\includegraphics[scale=0.15]{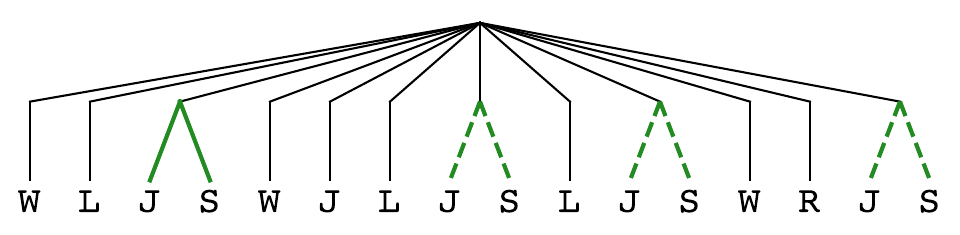}} \\
\midrule
\makecell{part. count}                      &                                                                                          10 &                                                                                           0 &                                                                                           5 &                                                                                           0 &                                                                                           9 \\
\midrule
\makecell{step count}                       &                                                                                          16 &                                                                                          17 &                                                                                          18 &                                                                                          20 &                                                                                          16 \\
\midrule
\makecell{prog. length}                     &                                                                                          13 &                                                                                          12 &                                                                                          13 &                                                                                          12 &                                                                                          14 \\
\midrule
\makecell{grammar induction \\ prior (log)} &                                                                                      -31.97 &                                                                                      -28.24 &                                                                                      -31.97 &                                                                                      -28.24 &                                                                                      -33.05 \\
\bottomrule
\end{tabular}
\caption{
Example programs demonstrating the influence of step costs on participant programs.
The most common hierarchical program (first column) uses a subroutine (\actionLeft, \actionJump, \actionLight) three times.
A program discovered by program search (second column) has the same subroutine and uses it four times.
This pattern of choice is inconsistent with MDL and grammar induction, but is explained by step count.
Third and fourth columns are a similar example, with the subroutine \actionWalk, \actionLeft, \actionJump, \actionLight.
Fifth column shows that participants will use a shorter subroutine (\actionJump, \actionLight) four times.
See Fig.~\ref{fig:sample-programs0} for more detail about this figure.
}
\label{fig:sample-programs3}
\end{center}
\end{figure}
\end{landscape}

The first two columns of Fig.~\ref{fig:sample-programs3} show programs with a shared subroutine (\actionLeft, \actionJump, \actionLight), where the program in the first column was created by participants and the one in the second column was found by our program search methods. The two programs differ very slightly: the one participants wrote has a subroutine used three times (instead of four), a longer program length, but a shorter trace length. Looking closely at the resulting trace, the programs are identical until they diverge in their approach to the final light. Participants take a more direct route to the final light (\actionWalk, \actionRight, \actionJump, \actionLight), instead of taking a longer route that requires one additional turn but can use the subroutine (\actionRight, \actionWalk, \actionLeft, \actionJump, \actionLight).
Participant preferences between these programs are inconsistent with the MDL or grammar induction accounts, but can be explained by the step cost prior.

The third and fourth columns of Fig.~\ref{fig:sample-programs3} provide a very similar example. The two programs share a subroutine (\actionWalk, \actionLeft, \actionJump, \actionLight) and only differ in their approach to the final light. As above, the program that participants created has three subroutine uses (instead of four), a longer program, and a shorter trace. This example can also be explained by the step cost prior.

We also include a fifth column which has the simplest subroutine (\actionJump, \actionLight), to show that participants will readily use a subroutine four times in this task.

These examples seem to suggest that the process of subroutine use might be informed by step costs, particularly since participants never create the shorter programs (in the second and fourth columns).

While our examples in the main text either have entirely different traces (Fig.~\ref{fig:sample-programs0}) or identical traces (Fig.~\ref{fig:sample-programs1}, Fig.~\ref{fig:sample-programs2}), this example highlights differences in participant choices near the end of a program. We think a promising direction for future experiments and analyses could focus on examples like this, where the trace length and grammar induction/MDL models make different predictions.

\subsection{Controlling for effects of program preprocessing}

\begin{figure}[t]
\centering

\begin{subfigure}[T]{.40\textwidth}
    \caption{}
    \includegraphics[scale=.5]{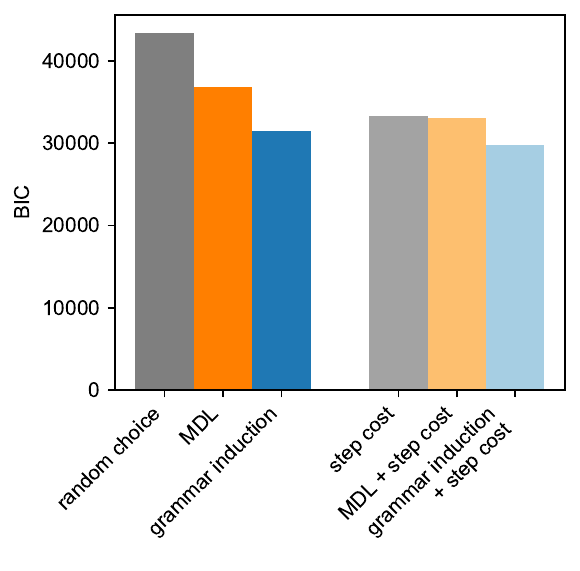}
    \label{fig:model-induction:no_canon:across-mdp}
\end{subfigure}
\begin{subfigure}[T]{.55\textwidth}
    \caption{}
    \includegraphics[scale=.5]{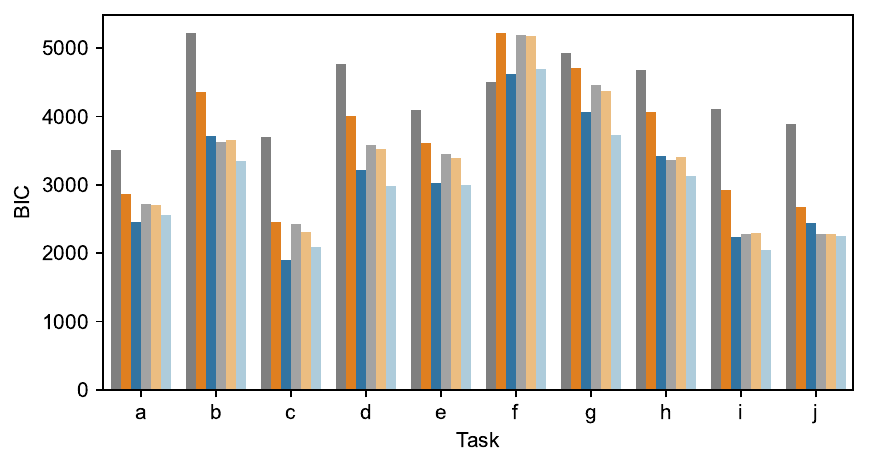}
    \label{fig:model-induction:no_canon:within-mdp}
\end{subfigure}

\caption{
Model comparison of accounts, without program preprocessing.
a) Plot of BIC of experimental data as predicted by each of the models, after parameter fitting. Models with smaller BIC are a better account of behavior. b) BIC of data under each model, but split by task. Parameters are the same as in a), so they are the best fit for all tasks. The color of the models is also the same as in a). Task letter is a reference to the subfigure in Fig.~\ref{fig:exp-task}.
}
\label{fig:model-induction:no_canon}
\end{figure}

\begin{table}[t]
\caption{
Table of fitted models to data without program preprocessing, with log likelihood, BIC, parameter count, and fitted parameters.
}
\label{tab:parameters:no_canon}
\begin{center}
\begin{tabular}{lcccl}
\toprule
{} & \makecell{Log \\ likelihood} &      BIC &  \makecell{Param. \\ count} &                                                                                                                   Parameters \\
\midrule
\makecell{random choice}                             &                     -21710.4 &  43420.7 &                           0 &                                                                                                               \makecell[l]{} \\
\midrule \makecell{MDL}                              &                     -18430.9 &  36869.3 &                           1 &                                                                                      \makecell[l]{$\beta_{\text{MDL}}=0.82$} \\
\midrule \makecell{grammar induction}                &                     -15706.6 &  31435.5 &                           3 &                                    \makecell[l]{$\alpha=68.12$ \\ $\beta_{\text{GrammarInduction}}=0.38$ \\ $p_{call}=0.28$} \\
\midrule \makecell{step cost}                        &                     -16665.0 &  33337.4 &                           1 &                                                                                 \makecell[l]{$\beta_{\text{StepCost}}=1.18$} \\
\midrule \makecell{MDL + step cost}                  &                     -16514.3 &  33043.4 &                           2 &                                                    \makecell[l]{$\beta_{\text{MDL}}=0.26$ \\ $\beta_{\text{StepCost}}=0.96$} \\
\midrule \makecell{grammar induction \\ + step cost} &                     -14905.1 &  29839.8 &                           4 &  \makecell[l]{$\alpha=45.48$ \\ $\beta_{\text{GrammarInduction}}=0.26$ \\ $p_{call}=0.17$ \\ $\beta_{\text{StepCost}}=0.59$} \\
\bottomrule
\end{tabular}
\end{center}
\end{table}

Our primary analysis in the main text compares models based on how well they predict programs. As reported, these programs have been preprocessed. Because preprocessing modifies programs in a way that could favor the grammar induction account, in particular since it maximizes the use of existing subroutines, we run a variant of the analysis that avoids any preprocessing of programs.

Our results are very similar to those reported in the main text.
The BIC, log likelihood, and fitted parameters for each model are reported in Table~\ref{tab:parameters:no_canon}.
We found that all models were more predictive of behavior than a random choice model (Likelihood-ratio test
for step cost: $\chi^2(1)=10090.8$, $p < .001$,
MDL: $\chi^2(1)=6558.9$, $p < .001$,
grammar induction: $\chi^2(3)=12007.5$, $p < .001$) or a step cost model
(Likelihood-ratio test
for MDL: $\chi^2(1)=301.4$, $p < .001$,
grammar induction: $\chi^2(3)=3519.8$, $p < .001$).
We found that the grammar induction model (with or without step costs) was most predictive of behavior (Fig.~\ref{fig:model-induction:no_canon:across-mdp}).
These qualitative results generally held on a task-specific basis (Fig.~\ref{fig:model-induction:no_canon:within-mdp}).

Of note, however, is that the $\alpha$ parameter is much higher for the grammar induction models than the analysis reported in the main text. While a bias towards reuse among existing subroutines is generally present in models due to the rich-get-richer dynamics of subroutine sampling, a high $\alpha$ leads to a preference for subroutine creation. This could be driven by the number of subroutines used only once, which are inlined by program preprocessing (Step 4 in Table~\ref{tab:preprocessing}), but present in this dataset.

\subsection{Controlling for programming experience}

In our model comparison in the main text, we included participants with programming experience equivalent to between 1 and 3 college courses. In order to control for any influence of this prior experience, we run our primary analyses in the subset of participants ($N=125$) that have no programming experience.

\begin{figure}[t]
\centering

\begin{subfigure}[T]{.40\textwidth}
    \caption{}
    \includegraphics[scale=.5]{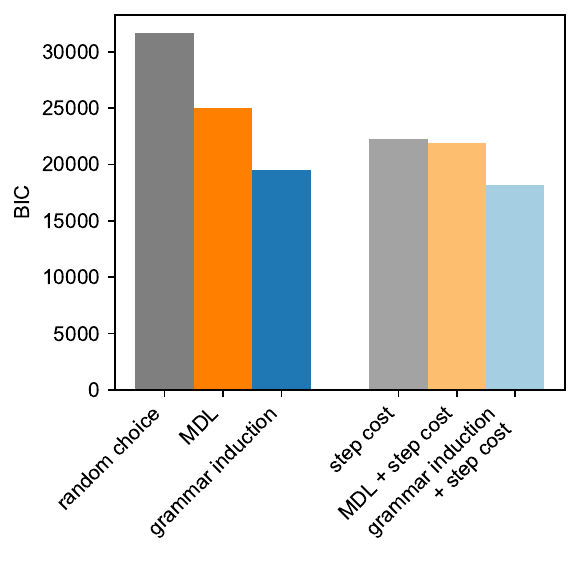}
    \label{fig:model-induction:no_prog_exp:across-mdp}
\end{subfigure}
\begin{subfigure}[T]{.55\textwidth}
    \caption{}
    \includegraphics[scale=.5]{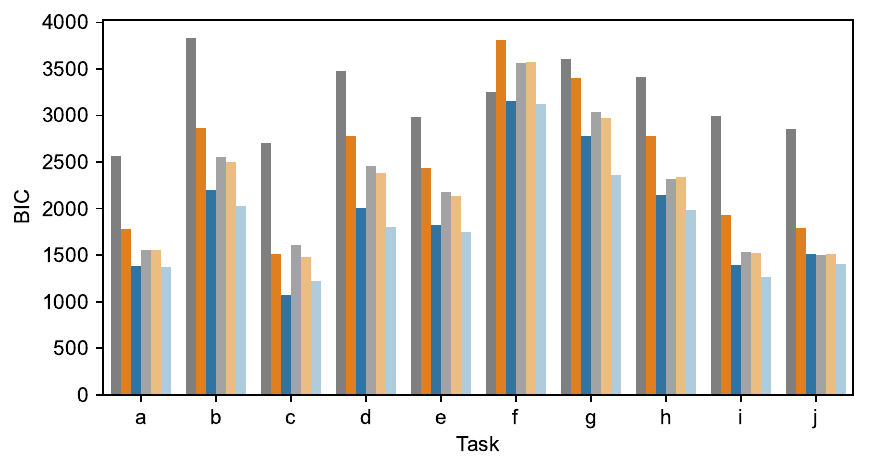}
    \label{fig:model-induction:no_prog_exp:within-mdp}
\end{subfigure}

\caption{
Model comparison of accounts, controlling for past programming experience.
a) Plot of BIC of experimental data as predicted by each of the models, after parameter fitting. Models with smaller BIC are a better account of behavior. b) BIC of data under each model, but split by task. Parameters are the same as in a), so they are the best fit for all tasks. The color of the models is also the same as in a). Task letter is a reference to the subfigure in Fig.~\ref{fig:exp-task}.
}
\label{fig:model-induction:no_prog_exp}
\end{figure}

We found extremely similar results as our analysis in the main text.
Table~\ref{tab:parameters:no_prog_exp} lists the BIC, log likelihood, and fit parameters for each model.
We found that models were an improvement over random choice
(Likelihood-ratio test for
step cost: $\chi^2(1)=9416.8$, $p < .001$,
MDL: $\chi^2(1)=6636.6$, $p < .001$,
grammar induction: $\chi^2(3)=12160.8$, $p < .001$)
and also an improvement over a baseline of step cost
(Likelihood-ratio test for
MDL + step cost: $\chi^2(1)=377.8$, $p < .001$,
grammar induction + step cost: $\chi^2(3)=4110.4$, $p < .001$).
Judged by BIC, the grammar induction model (with or without step cost) was the most predictive of behavior (Fig.~\ref{fig:model-induction:no_prog_exp:across-mdp}).
These patterns generally held when examined on a per-task basis (Fig.~\ref{fig:model-induction:no_prog_exp:within-mdp}).

\begin{table}[t]
\caption{
Table of fitted models to data excluding participants with programming experience, with log likelihood, BIC, parameter count, and fitted parameters.
}
\label{tab:parameters:no_prog_exp}
\begin{center}
\begin{tabular}{lcccl}
\toprule
{} & \makecell{Log \\ likelihood} &      BIC &  \makecell{Param. \\ count} &                                                                                                                  Parameters \\
\midrule
\makecell{random choice}                             &                     -15840.5 &  31680.9 &                           0 &                                                                                                              \makecell[l]{} \\
\midrule \makecell{MDL}                              &                     -12522.2 &  25051.5 &                           1 &                                                                                     \makecell[l]{$\beta_{\text{MDL}}=0.95$} \\
\midrule \makecell{grammar induction}                &                      -9760.1 &  19541.5 &                           3 &                                    \makecell[l]{$\alpha=4.15$ \\ $\beta_{\text{GrammarInduction}}=0.48$ \\ $p_{call}=0.10$} \\
\midrule \makecell{step cost}                        &                     -11132.1 &  22271.3 &                           1 &                                                                                \makecell[l]{$\beta_{\text{StepCost}}=1.38$} \\
\midrule \makecell{MDL + step cost}                  &                     -10943.2 &  21900.6 &                           2 &                                                   \makecell[l]{$\beta_{\text{MDL}}=0.36$ \\ $\beta_{\text{StepCost}}=1.07$} \\
\midrule \makecell{grammar induction \\ + step cost} &                      -9076.9 &  18182.2 &                           4 &  \makecell[l]{$\alpha=2.16$ \\ $\beta_{\text{GrammarInduction}}=0.35$ \\ $p_{call}=0.06$ \\ $\beta_{\text{StepCost}}=0.64$} \\
\bottomrule
\end{tabular}

\end{center}
\end{table}

\subsection{Does programming experience predict task performance?}

\begin{figure}[t]
\centering

\begin{subfigure}[t]{.30\textwidth}
    \includegraphics[width=\textwidth]{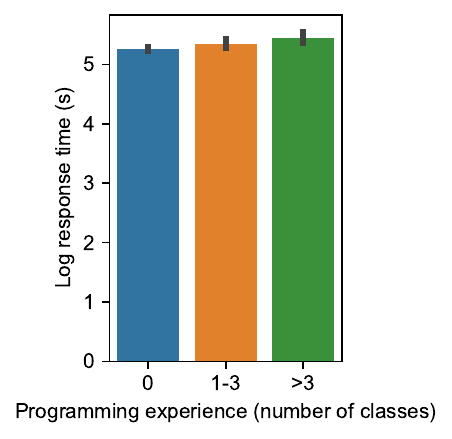}
    \caption{}
    \label{fig:prog_exp_predict:log_elapsed_seconds}
\end{subfigure}
\begin{subfigure}[t]{.30\textwidth}
    \includegraphics[width=\textwidth]{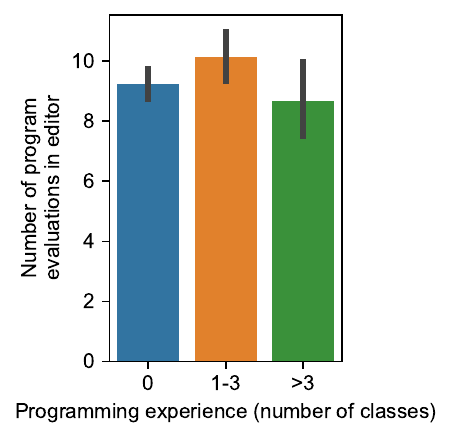}
    \caption{}
    \label{fig:prog_exp_predict:number_exec}
\end{subfigure}
\begin{subfigure}[t]{.30\textwidth}
    \includegraphics[width=\textwidth]{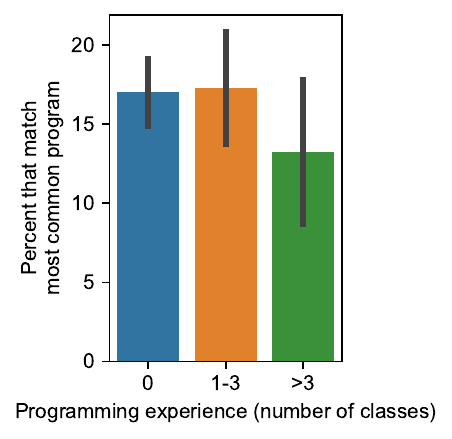}
    \caption{}
    \label{fig:prog_exp_predict:produced_most_common}
\end{subfigure}

\caption{
Plotting measures of task performance for different levels of programming experience.
Values are a) log-transformed response times (s), b) the number of program evaluations, and c) the percentage of programs that match the most common program.
Error bars show the 95\% confidence interval of the mean, estimated by bootstrapping.
}
\label{fig:prog_exp_predict}
\end{figure}

Another question of interest is whether programming experience has an influence on measures of task performance, like how quickly tasks were completed.
In order to test these questions, we analyze all 193 participants that completed the task, which included the 22 participants with programming experience equivalent to $>3$ college courses, who were excluded from analyses in the main text.
For each measure of task performance, we used a likelihood-ratio test to see whether adding a fixed effect for programming experience would significantly improve upon a null model, which predicted the measure with a fixed intercept and random per-participant intercepts.
We found that programming experience did not improve predictions of task performance for any measures we considered, which consisted of
response times ($\chi^2(2)=3.04$, $p = .218$; Fig.~\ref{fig:prog_exp_predict:log_elapsed_seconds}),
number of program evaluations ($\chi^2(2)=0.94$, $p = .625$; Fig.~\ref{fig:prog_exp_predict:number_exec}),
and percentage of programs that match the common program ($\chi^2(2)=2.08$, $p = .353$; Fig.~\ref{fig:prog_exp_predict:produced_most_common}).
These findings suggest that prior programming instruction or experience may have minimal impact on task performance in Lightbot.

\begin{figure}[t]
\centering

\begin{subfigure}[t]{.45\textwidth}
    \includegraphics[width=\textwidth]{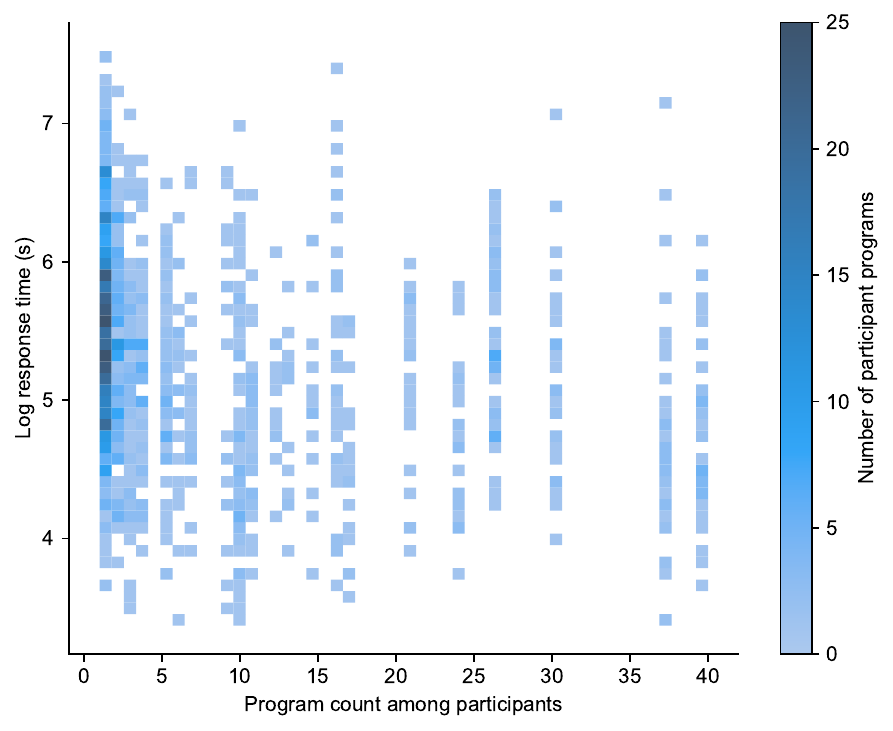}
    \caption{}
    \label{fig:program-ease:no_prog_exp:time}
\end{subfigure}
\begin{subfigure}[t]{.45\textwidth}
    \includegraphics[width=\textwidth]{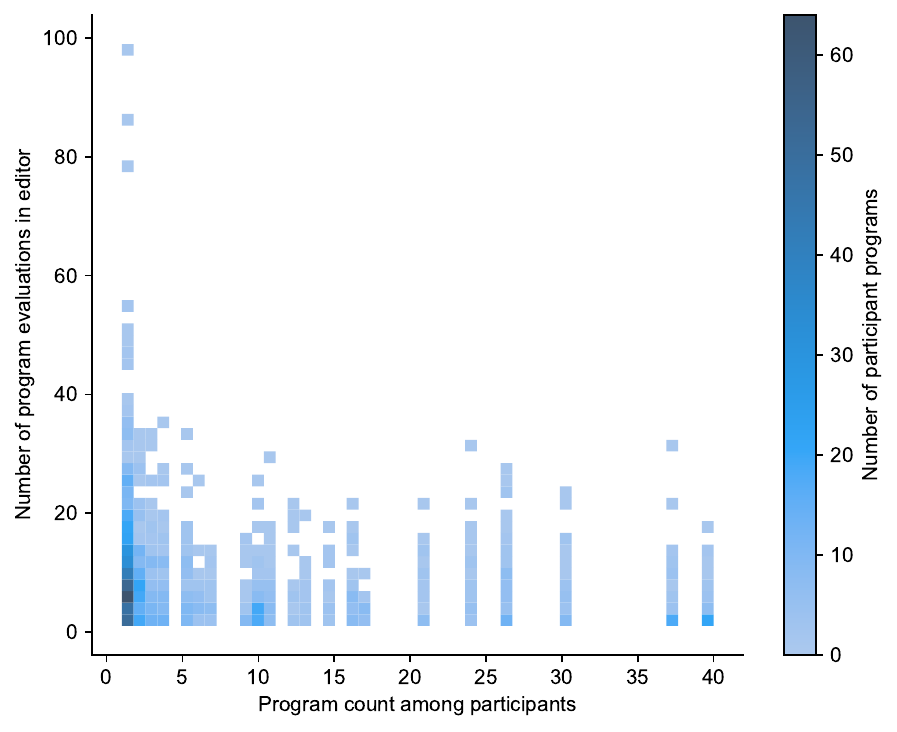}
    \caption{}
    \label{fig:program-ease:no_prog_exp:runs}
\end{subfigure}

\caption{
Participants without programming experience write common programs faster and with fewer program evaluations.
Bivariate histogram of task responses for each participant on each task, 
with responses binned along horizontal axis by program count and along vertical axis by either a) log-transformed response times (s) or b) the number of program evaluations.
A related regression analysis is reported in the text, which uses program count to predict response times and number of program evaluations while controlling for program length and cost, and finds results consistent with these plots.
}
\label{fig:program-ease:no_prog_exp}
\end{figure}

Another test that could be impacted by our inclusion of those with programming experience is our analysis of performance characteristics of those who wrote common programs (in Fig.~\ref{fig:program-ease}). In particular, in the main text we found that participants, when writing common programs, completed the task more quickly and with fewer program evaluations. Running the same analysis with the 125 participants that had no programming experience, we found similar results of faster responses ($\beta=-0.011$, $\chi^2(1)=43.77$, $p < .001$, Fig.~\ref{fig:program-ease:no_prog_exp:time}) and fewer program evaluations ($\beta=-0.054$, $\chi^2(1)=6.34$, $p = .012$, Fig.~\ref{fig:program-ease:no_prog_exp:runs}).

\subsection{Examining variance in task solutions}

\begin{figure}[t]
\centering
\includegraphics[scale=.7]{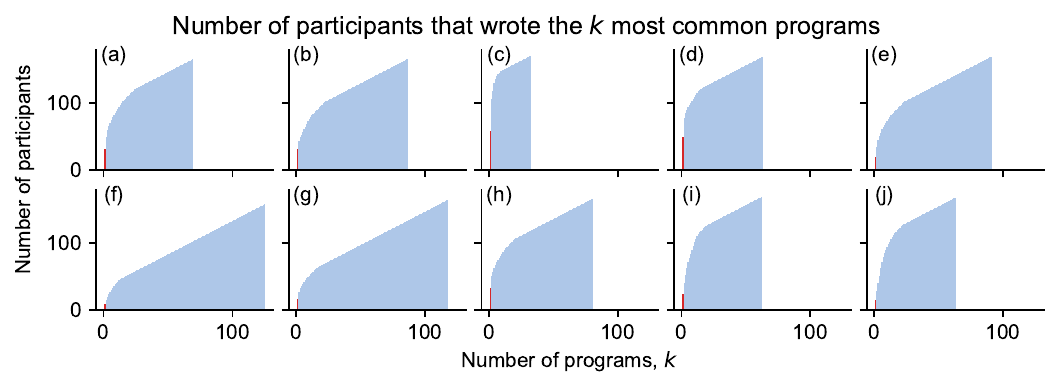}
\caption{
Plotting the cumulative number of participants that wrote the $k$ most common programs. The bar corresponding to the most common program is the leftmost bar in each subfigure, at left and in red.
Task label is a reference to subfigure in Fig.~\ref{fig:exp-task}.
}
\label{fig:program-histogram}
\end{figure}

\begin{landscape}\begin{table}[t]
\caption{
Measures of task properties, difficulty, and solution variance.
Each column is a different task, with letter referencing a subfigure in Fig.~\ref{fig:exp-task}.
Rows report task properties (light count), difficulty (mean across participants of log response time, mean across participants of number of program evaluations, number of times task was skipped), and solution variance (number of unique solutions, and number of participants that wrote the most common program).
}
\label{tab:task-desc}
\begin{center}
\begin{tabular}{lllllllllll}
\toprule
Task &                          a &                          b &                          c &                          d &                          e &                        f &                          g &                          h &                          i &                         j \\
\midrule
Light count &                          3 &                          8 &                          3 &                          6 &                          3 &                        7 &                          8 &                          6 &                          4 &                         3 \\
\midrule \makecell[l]{Mean log response \\ time (seconds)} &                       5.69 &                       5.71 &                       4.87 &                       5.41 &                       5.27 &                     5.72 &                       5.50 &                       5.22 &                       4.88 &                      4.61 \\
\midrule \makecell[l]{Mean program \\ evaluations} &                       9.31 &                      12.55 &                       5.42 &                       9.92 &                       8.69 &                    14.80 &                      12.24 &                       9.60 &                       7.14 &                      5.60 \\
\midrule Times skipped &                          5 &                          5 &                          0 &                          1 &                          1 &                       13 &                          7 &                          5 &                          2 &                         3 \\
\midrule \makecell[l]{Number of \\ unique solutions} &                         69 &                         86 &                         32 &                         63 &                         91 &                      125 &                        117 &                         80 &                         62 &                        63 \\
\midrule \makecell[l]{Number of participants \\ with most common program} & \makecell[l]{32 \\ (19\%)} & \makecell[l]{31 \\ (19\%)} & \makecell[l]{59 \\ (35\%)} & \makecell[l]{50 \\ (29\%)} & \makecell[l]{19 \\ (11\%)} & \makecell[l]{8 \\ (5\%)} & \makecell[l]{16 \\ (10\%)} & \makecell[l]{32 \\ (19\%)} & \makecell[l]{23 \\ (14\%)} & \makecell[l]{15 \\ (9\%)} \\
\bottomrule
\end{tabular}
\end{center}
\end{table}
    \end{landscape}

We examine the variability in how participants solve tasks, focusing on two questions: How does this variability relate to task performance? Can we use our models to predict which tasks will elicit greater variability?
We measure variance in problem solutions in two ways: the number of unique solutions, and the number of participants that wrote the most common program. The quantities we analyze in this section are reported in Table~\ref{tab:task-desc} and a histogram of program frequencies is in Fig.~\ref{fig:program-histogram}.

\begin{figure}[t]
\centering
\includegraphics[scale=.6]{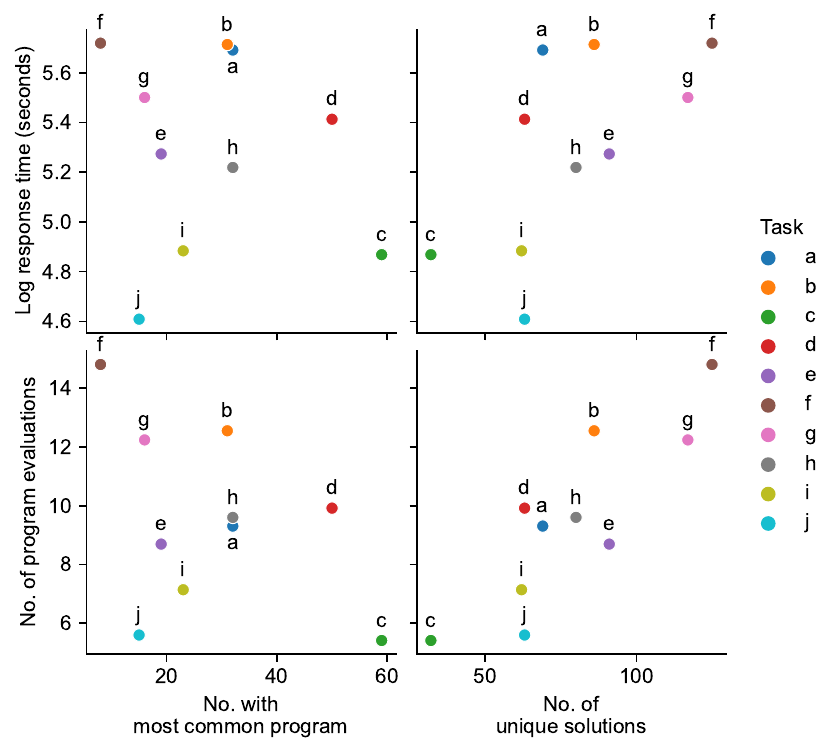}
\caption{
Plotting measures of task difficulty (response times, number of program evaluations) and solution variability (unique solution count, number of participants with most common program).
Task label is a reference to subfigure in Fig.~\ref{fig:exp-task}.
}
\label{fig:solvar_taskperf}
\end{figure}

We first examine whether solution variability can be linked to measures of task difficulty previously examined: response times and number of program evaluations (Fig.~\ref{fig:solvar_taskperf}).
We report Spearman's rank correlation coefficient between measures of difficulty and variance, using a permutation test to evaluate statistical significance.
We found that unique solution count had a positive relationship with increased task difficulty
(program evaluation count: $\rho=0.78$, $p = .01$, $N=10$,
response times: $\rho=0.73$, $p = .021$, $N=10$).
We found a negative relationship (that did not reach statistical significance) between task difficulty and the number of participants who wrote the most common program
(program evaluation count: $\rho=-0.29$, $p = .414$, $N=10$,
response times: $\rho=-0.20$, $p = .572$, $N=10$).
These results both suggest that tasks with greater solution variability
also tend to be more difficult.

\begin{figure}[t]
\centering
\includegraphics[scale=.7]{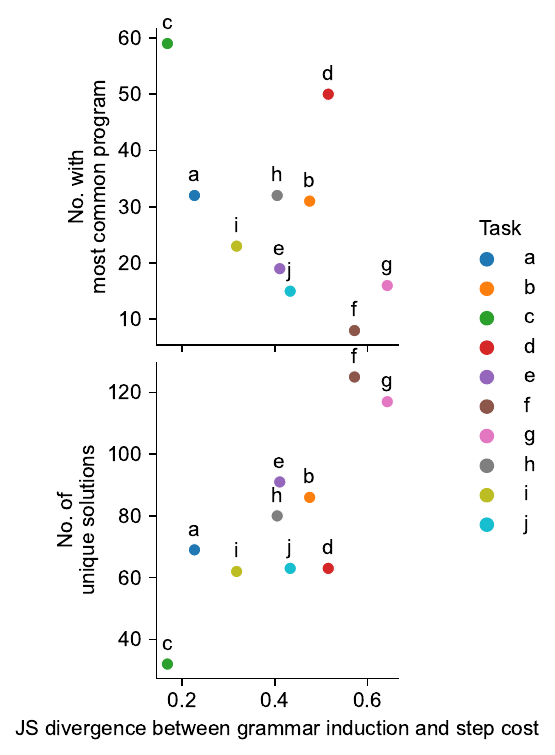}
\caption{
Plotting a measure of theoretical solution variability (the JS divergence between the grammar induction and step cost priors, using fitted parameters from Tab.~\ref{tab:parameters}) against two measures of empirical solution variability (number of unique solutions, and number of participants that wrote the most common program).
Task label is a reference to subfigure in Fig.~\ref{fig:exp-task}.
}
\label{fig:jsdiv_solvar}
\end{figure}

Can we predict the empirical variability we see, using the predictions of our models? The variability could arise from the competing objectives that people are balancing. For example, our best fitting model combines two objectives: the grammar induction prior and the step cost prior. In cases where these two priors disagree, differences in their relative weight could lead to considerably different model predictions. So any variability in their relative weight in our research participants could be a source of solution variability.

We test this idea by seeing whether our measures of empirical solution variance are related to a theoretical measure about the level of disagreement between our theories of program writing.
We quantify the disagreement between two theories as the Jensen-Shannon (JS) divergence of their distributions over programs, using the fitted parameters in Table~\ref{tab:parameters}.
The JS divergence is
$$
D_{JS}(p \mid\mid q) = \frac{1}{2} \left[ D_{KL}(p \mid\mid m) + D_{KL}(q \mid\mid m) \right]
$$
where $m(\rho) = \frac{1}{2}\left[p(\rho) + q(\rho)\right]$ is a mixture of the two distributions and
$D_{KL}(p \mid\mid m) = \sum_\rho p(\rho) \log \left( \frac{p(\rho)}{m(\rho)} \right)$
is the Kullback-Leibler divergence.
The JS divergence is a non-negative measurement of how different two distributions are, taking the value of 0 when the two distributions are the same.

Since our best model combines the grammar induction and step cost priors, we used the JS divergence between these two priors as a measure of theoretical disagreement, and compared them to our measures of empirical solution variability (Fig.~\ref{fig:jsdiv_solvar}).
We found a positive relationship between the JS divergence and the number of unique solutions ($\rho=0.69$, $p = .034$, $N=10$) and a negative (but not statistically significant) relationship between the JS divergence and the count of participants that wrote the most common program ($\rho=-0.58$, $p = .083$, $N=10$). The direction of these correlations is consistent with the idea that disagreement between priors could be a source of increased solution variance.

\subsection{The influence of program evaluation on subroutine creation}

\begin{figure}[t]
\centering
\includegraphics[scale=.5]{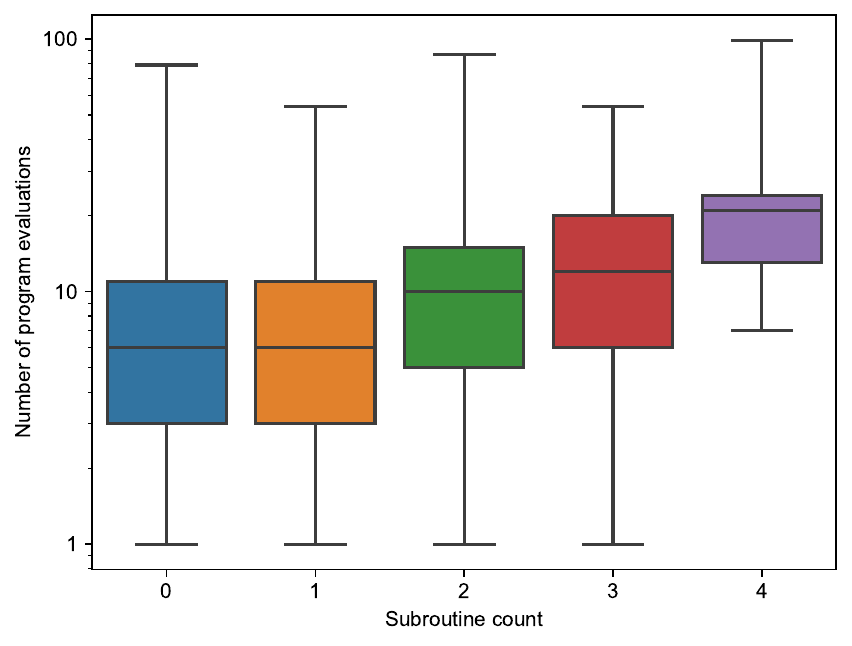}
\caption{
Plotting the number of program evaluations for trials with varying numbers of subroutines. The box shows data quartiles, while whiskers show range of the data.
}
\label{fig:program-tweaking}
\end{figure}

In contrast to typical studies of planning,
where a plan might be updated in the course of its execution,
our task permits the testing and revision of a plan.
A natural question is whether the ability to evaluate programs in Lightbot has an influence on the programs people write.
One simple test is to see whether the number of subroutines participants used was associated with the number of times they evaluated their program.
A positive relationship might suggest that the use of subroutines requires validating the subroutine has the expected result. A negative relationship might suggest that participants avoid validating subroutines because it is easy to reason about their expected results.
We find a positive relationship between program evaluations and subroutine counts ($\rho=0.21$, $p < .001$, $N=1668$, Fig.~\ref{fig:program-tweaking}), which holds even after excluding outlier trials where participants evaluated programs more than 40 times ($\rho=0.20$, $p < .001$, $N=1649$).
While we avoid any causal interpretation of these results, we hope future research can continue to examine the influence of process-tracing paradigms on behavior.

\subsection{Individual differences in subroutine use}

\begin{figure}[t]
\centering
\begin{subfigure}[t]{.45\textwidth}
    \includegraphics[width=\textwidth]{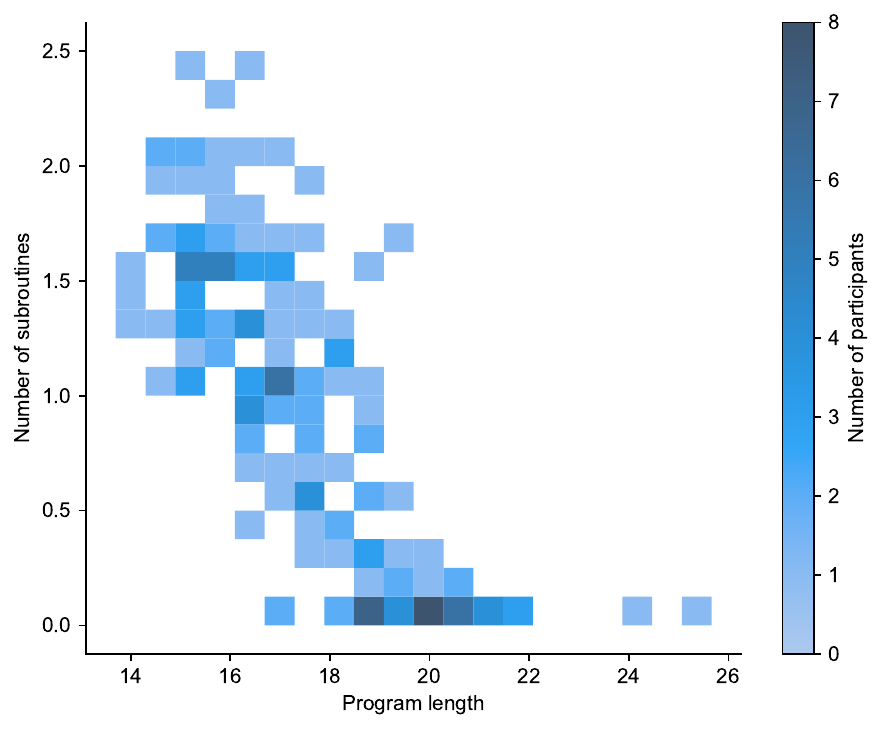}
    \caption{}
    \label{fig:indiv_diff:program_length-sr_count}
\end{subfigure}
\begin{subfigure}[t]{.45\textwidth}
    \includegraphics[width=\textwidth]{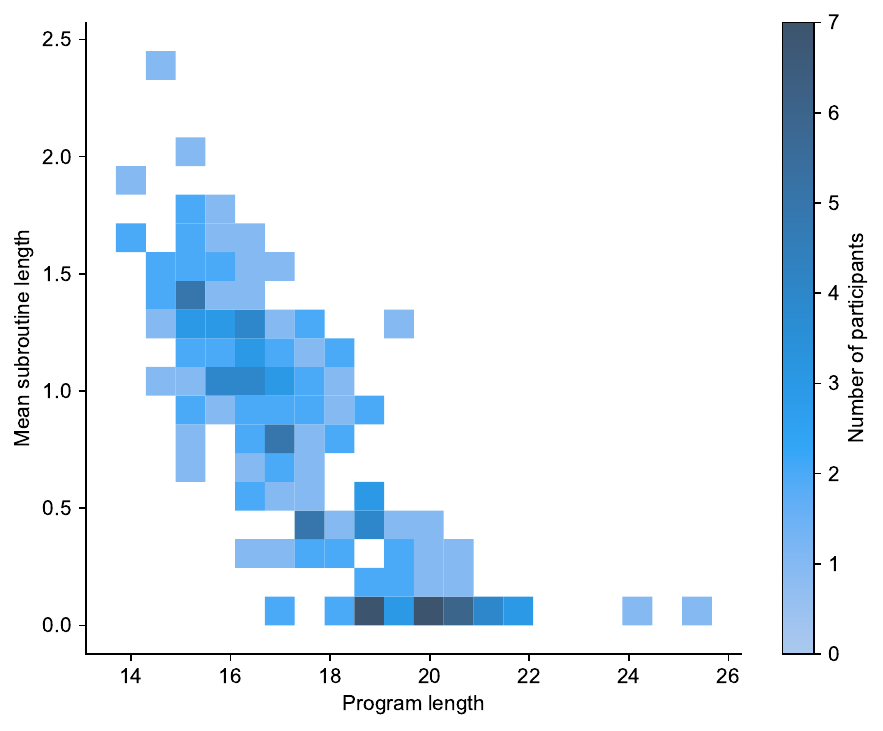}
    \caption{}
    \label{fig:indiv_diff:program_length-sr_len}
\end{subfigure}
\caption{
Showing individual differences between participants by plotting program length against a) number of created subroutines and b) the average length of created subroutines. Participant-level values are averaged across tasks.
}
\label{fig:indiv_diff}
\end{figure}

Our analyses have largely focused on characterizing behavior at the group level. However, there is considerable variety in the programs that people write.
Are these programs merely superficially different, with similar structural characteristics, or do participants have different structural preferences?
We investigate this at the individual level (averaging over tasks) by plotting program length against attributes related to subroutines\textemdash the number of subroutines (Fig.~\ref{fig:indiv_diff:program_length-sr_count}), and their average length (Fig.~\ref{fig:indiv_diff:program_length-sr_len}).
Testing the relationship between these variables, we find a significant negative correlation between
program length and subroutine count
($\rho=-0.79$, $p < .001$, $N=171$), as well as between program length and mean subroutine length ($\rho=-0.83$, $p < .001$, $N=171$).
These findings suggest that the variety in participant programs is fit well by a single dimension, with short programs and more/longer subroutines at one extreme, and long programs with fewer/shorter subroutines at the other.

\subsection{Experiment tutorial}

The complete experiment tutorial is presented in Fig.~\ref{fig:experiment-tutorial1} to Fig.~\ref{fig:experiment-tutorial4}. It explains the task (Fig.~\ref{fig:experiment-tutorial-inst-1}--\ref{fig:experiment-tutorial-inst-3}), how to use each of the program instructions (Fig.~\ref{fig:experiment-tutorial-inst-4}--\ref{fig:experiment-tutorial-inst-13}), and how to use and create subroutines (Fig.~\ref{fig:experiment-tutorial-inst-14}--\ref{fig:experiment-tutorial-inst-15}). Screenshots with a \emph{Run} button at upper left require the solution of a task (e.g., Fig.~\ref{fig:experiment-tutorial-inst-7}), while others are simply informational. Then, participants complete a more complex practice problem (Fig.~\ref{fig:experiment-tutorial-practice}). Finally, instructions for the incentive for minimizing instruction count (i.e. program length) are shown in Fig.~\ref{fig:experiment-tutorial4}. It shows a number of example programs for a given trace, in order to demonstrate how program length varies with the choice of subroutines.

\newcommand\tutorialfig[1]{\begin{subfigure}[t]{.49\textwidth}
\fbox{\includegraphics[scale=.18]{figures/tutorial/#1}}
    \caption{}
    \label{fig:experiment-#1}
\end{subfigure}}

\begin{figure}[hp!]
\centering

\tutorialfig{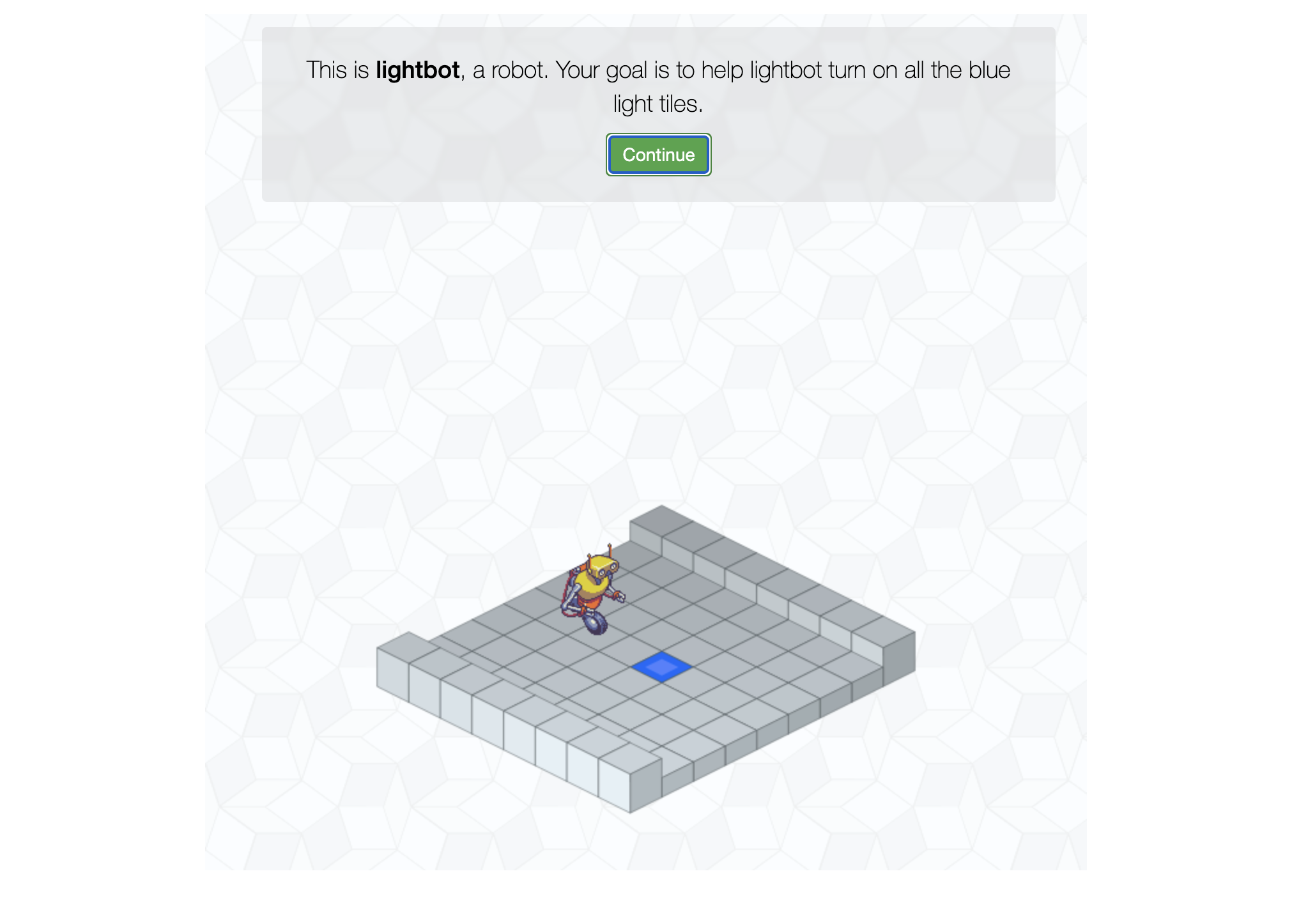}
\tutorialfig{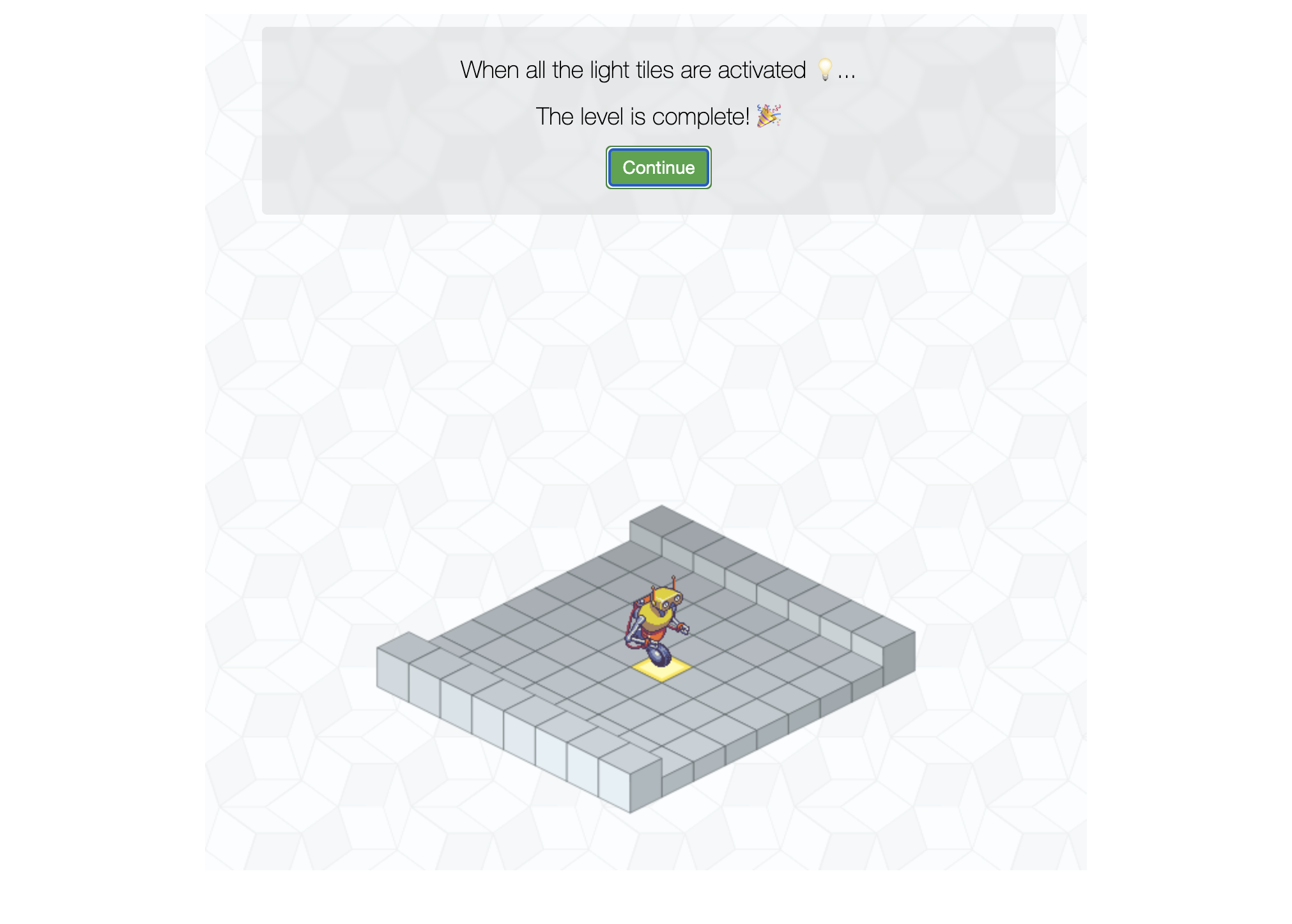}

\tutorialfig{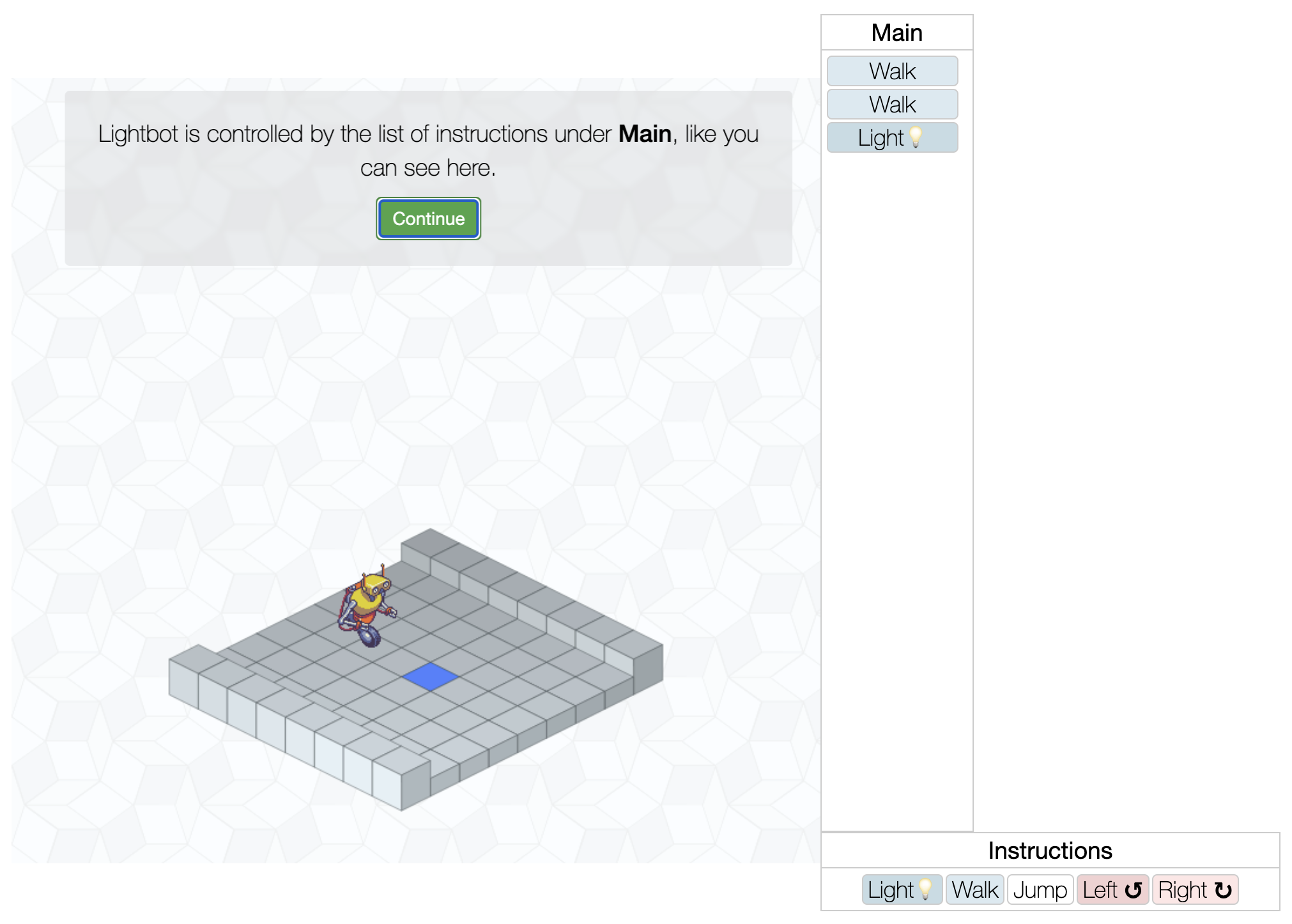}
\tutorialfig{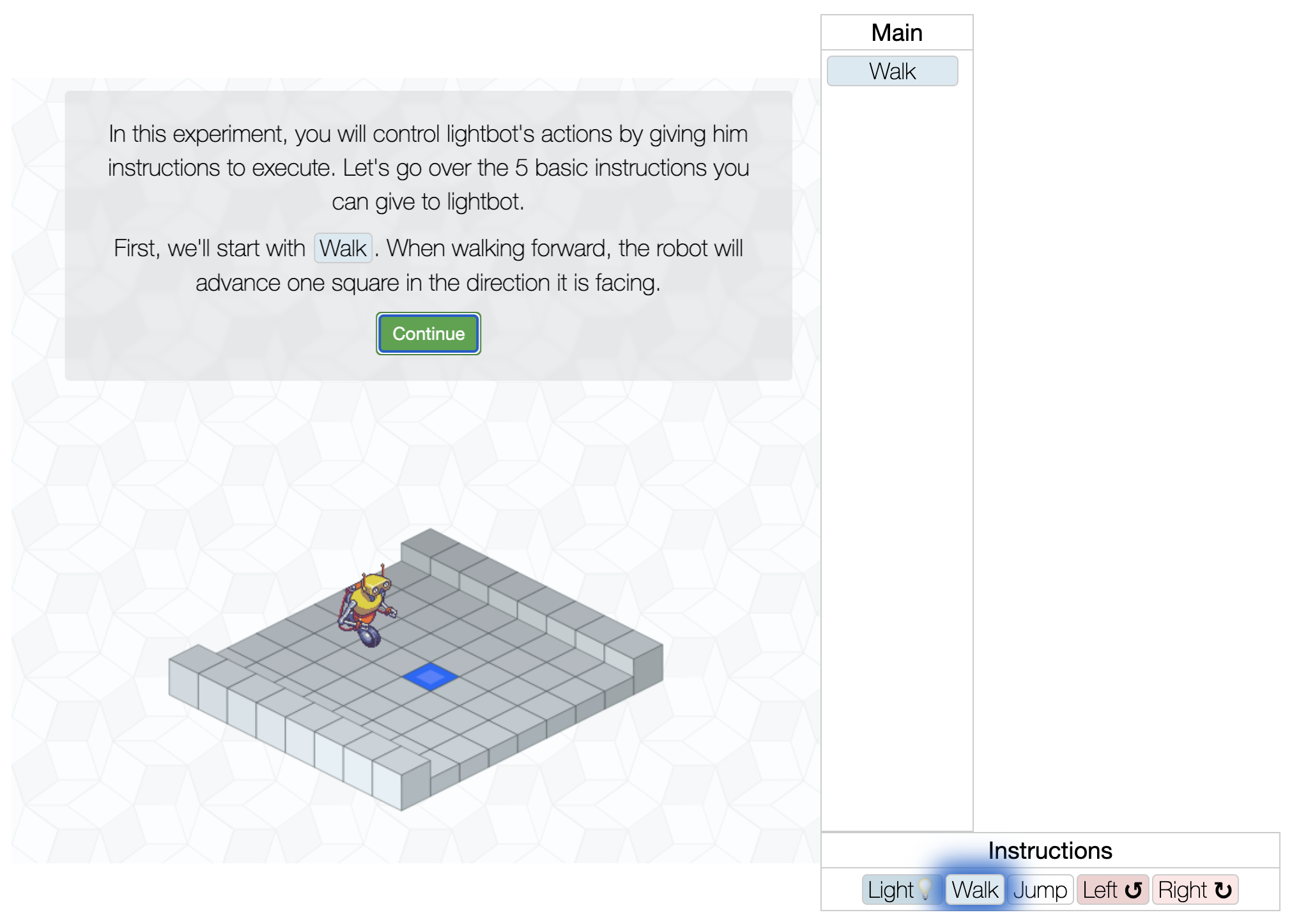}

\tutorialfig{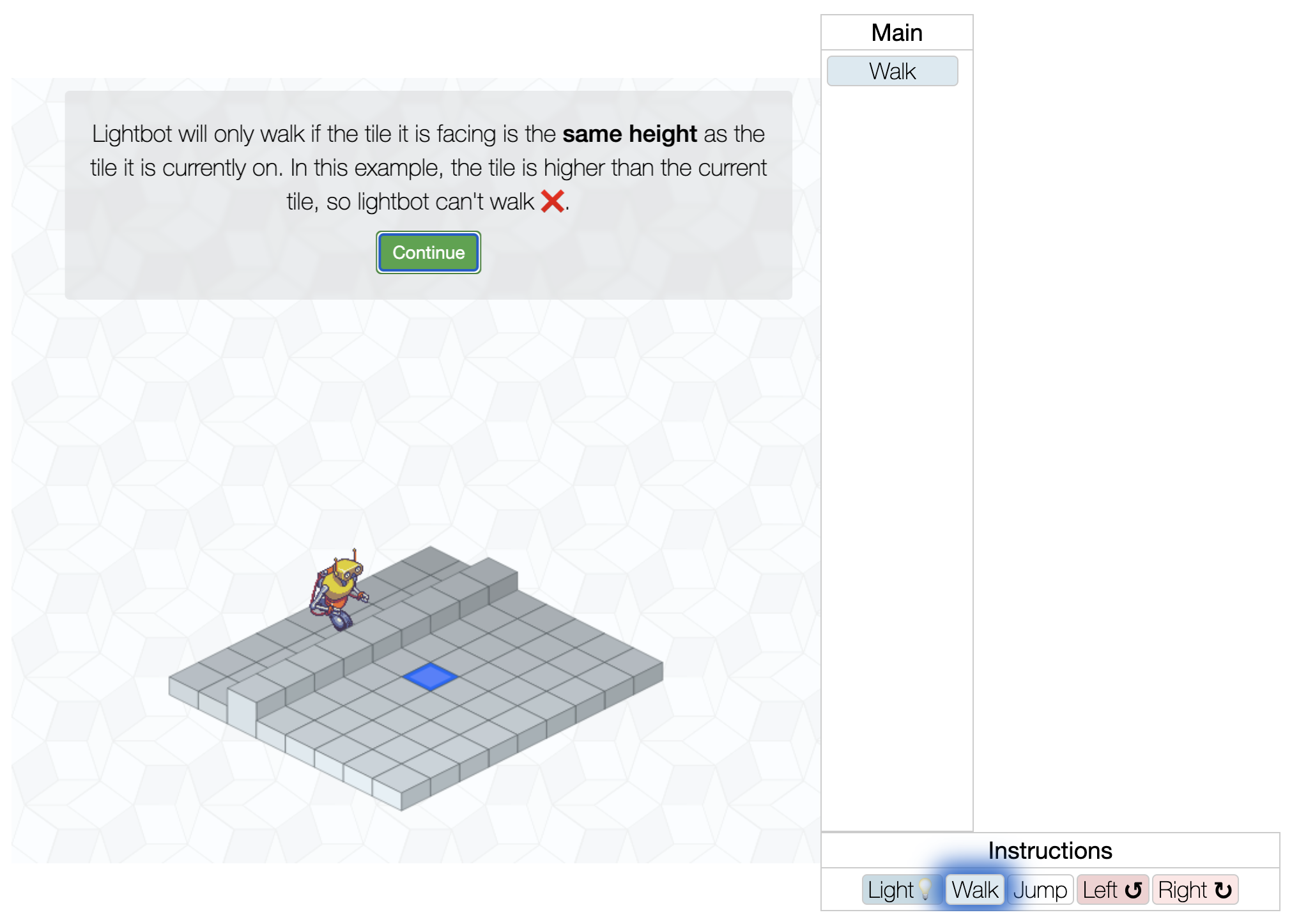}
\tutorialfig{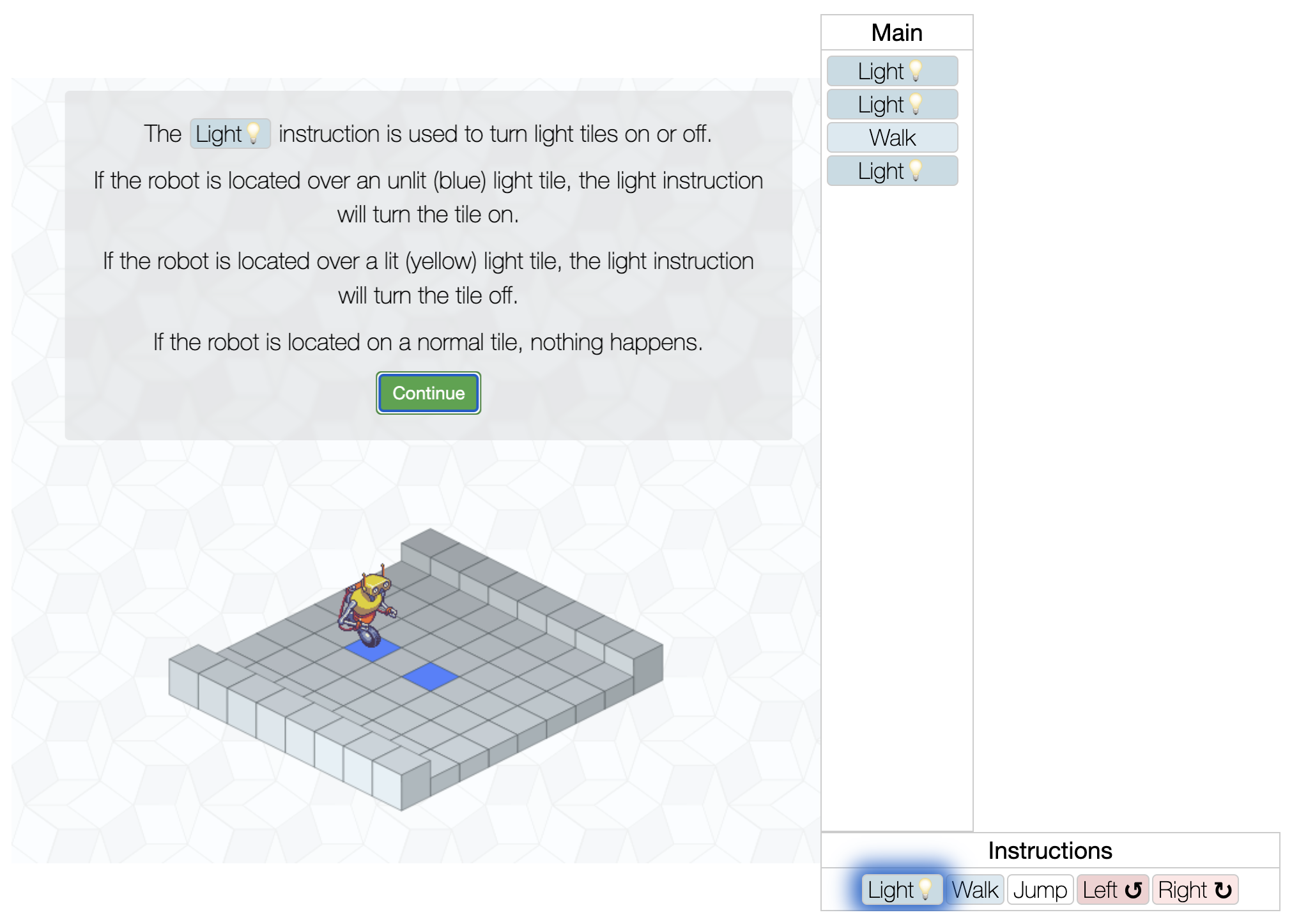}

\caption{
Screenshots of the experiment tutorial, in the same order as presented to research participants.
}
\label{fig:experiment-tutorial1}
\end{figure}

\begin{figure}[hp!]
\centering

\tutorialfig{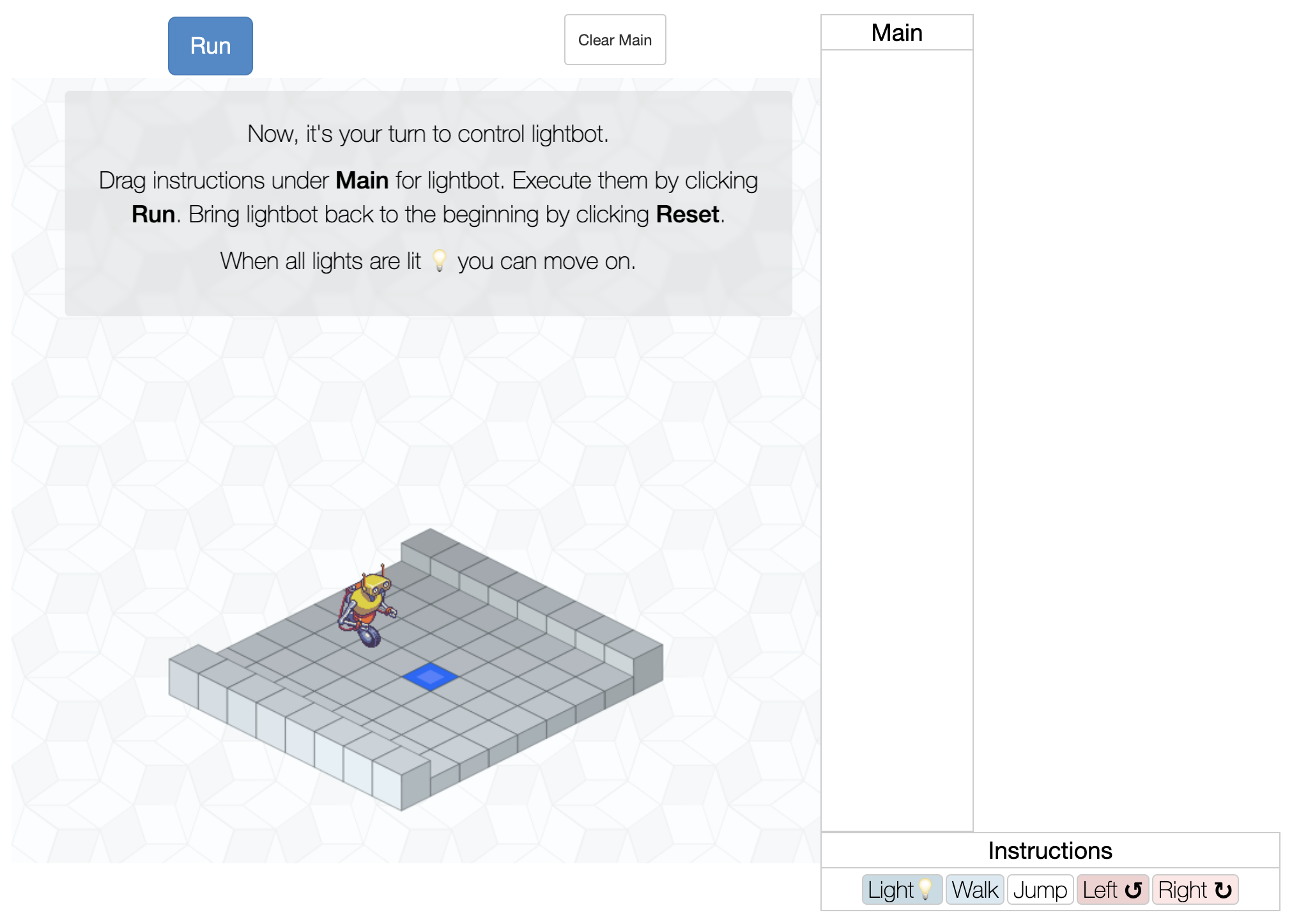}
\tutorialfig{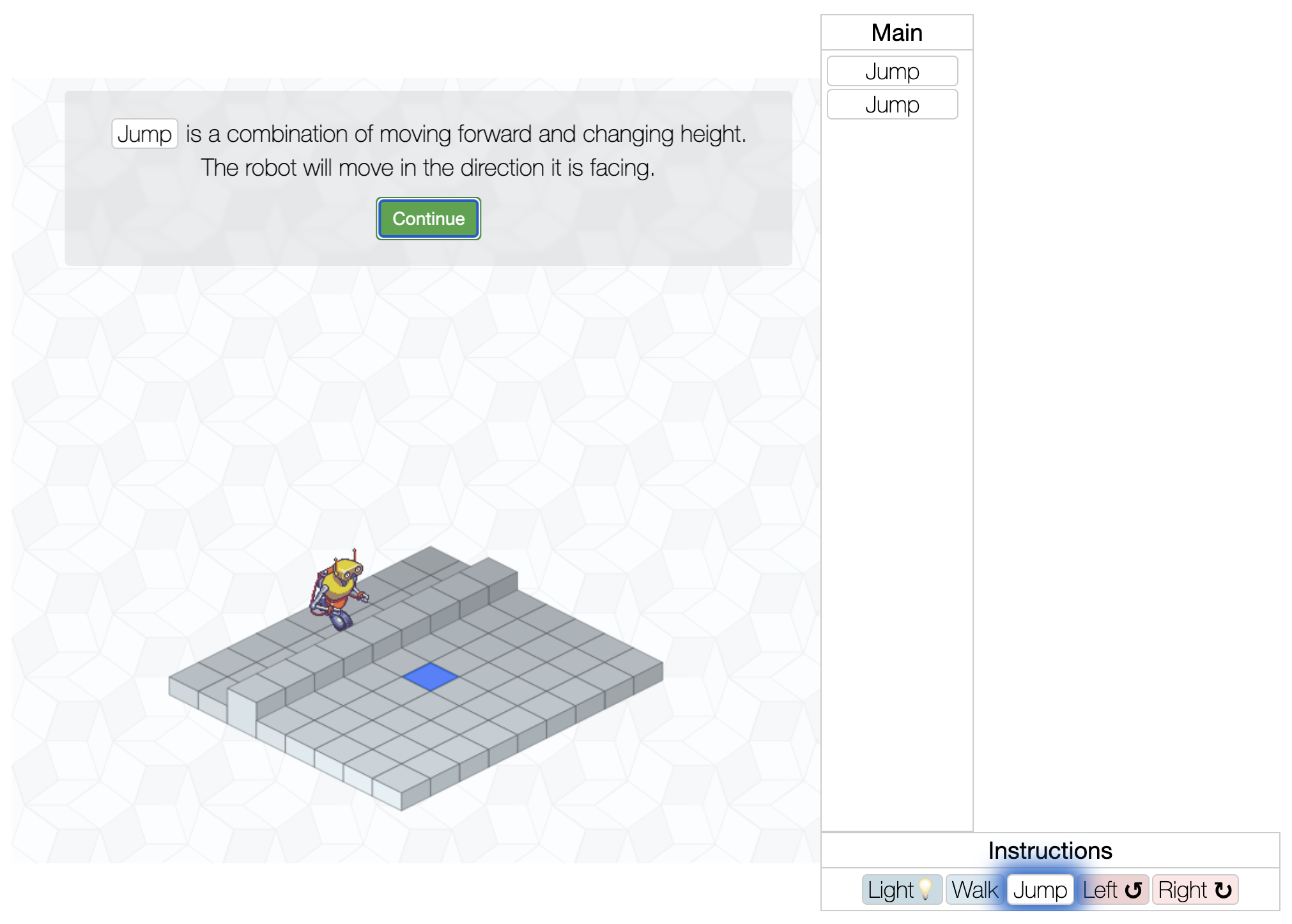}

\tutorialfig{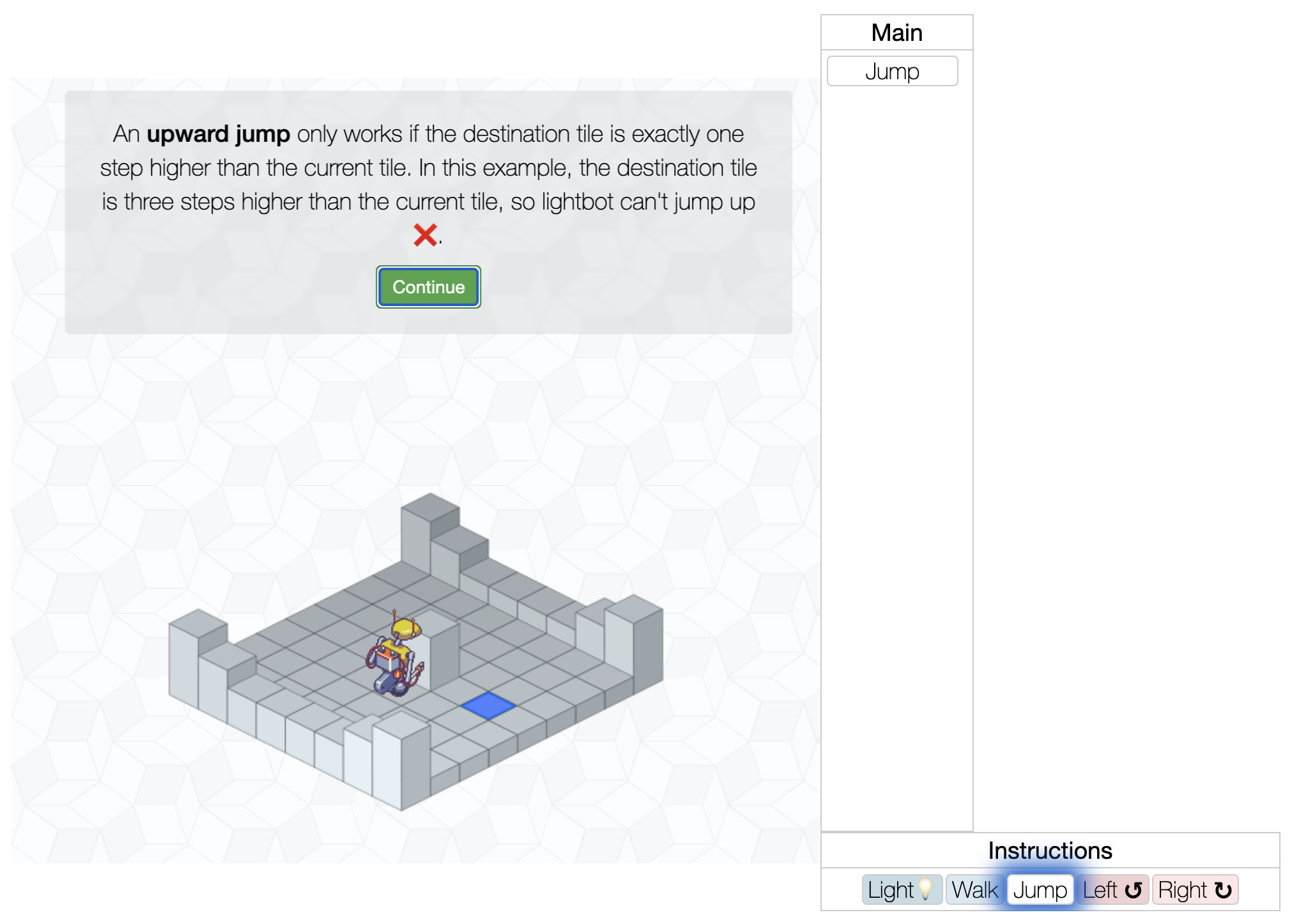}
\tutorialfig{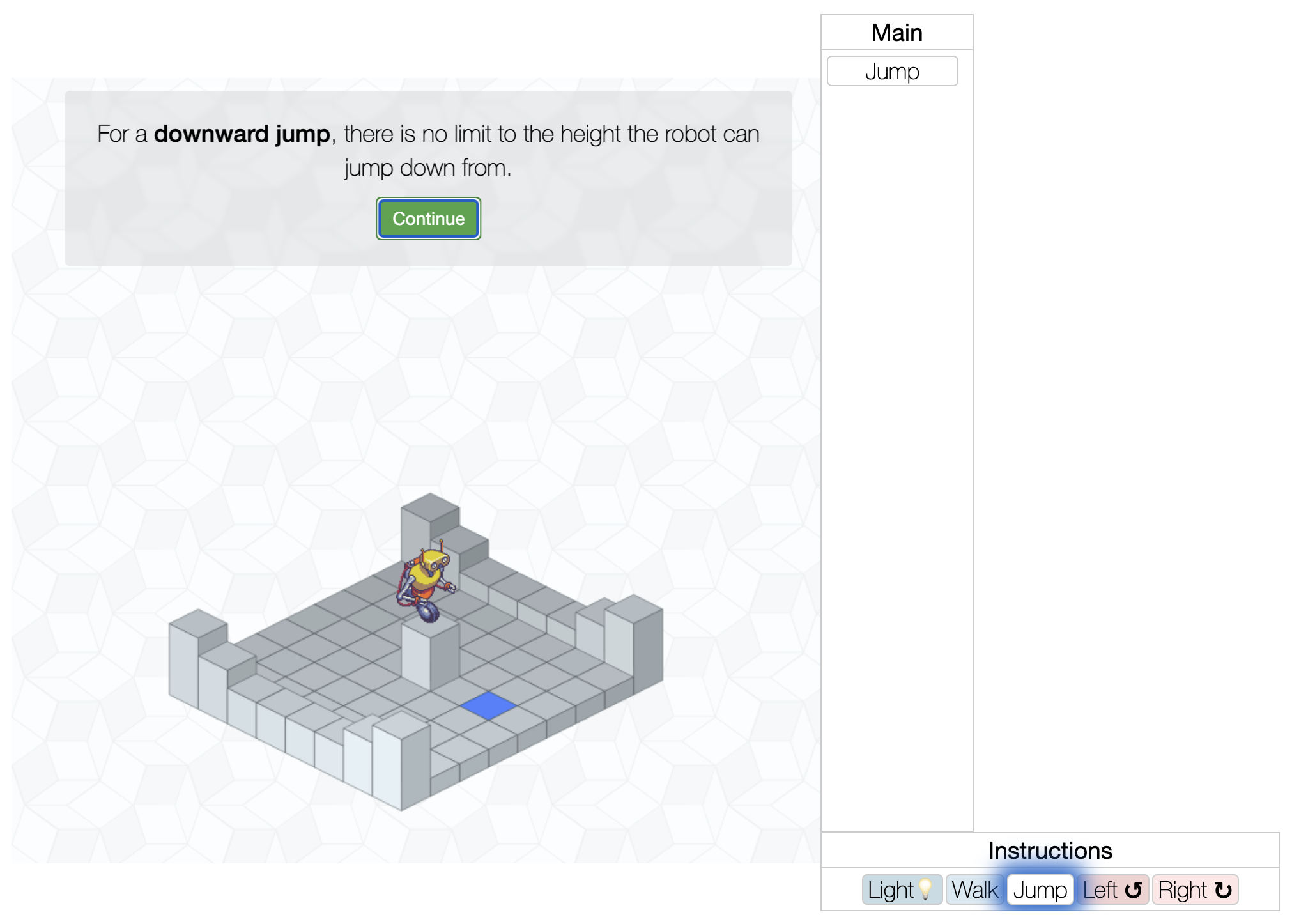}

\tutorialfig{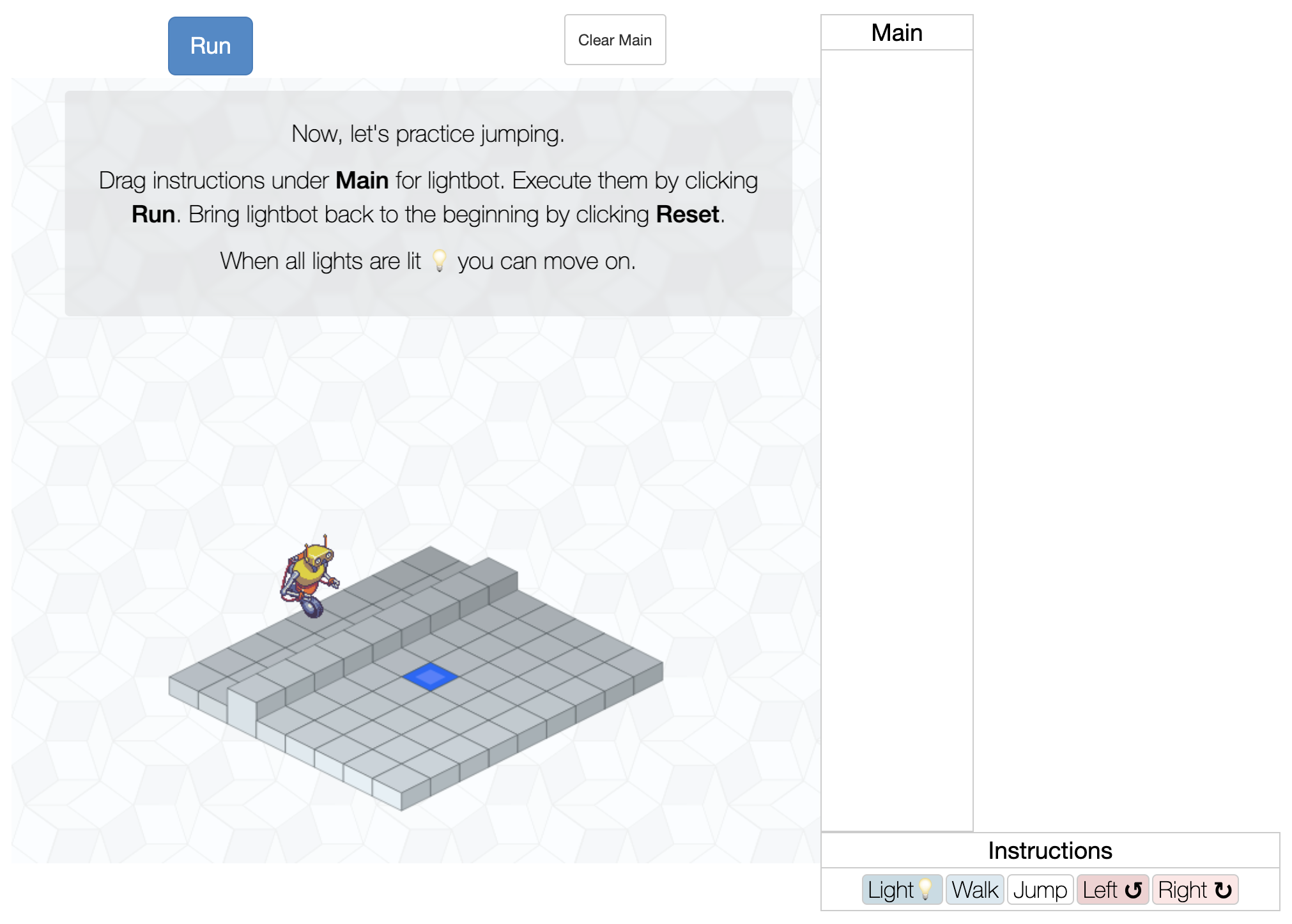}
\tutorialfig{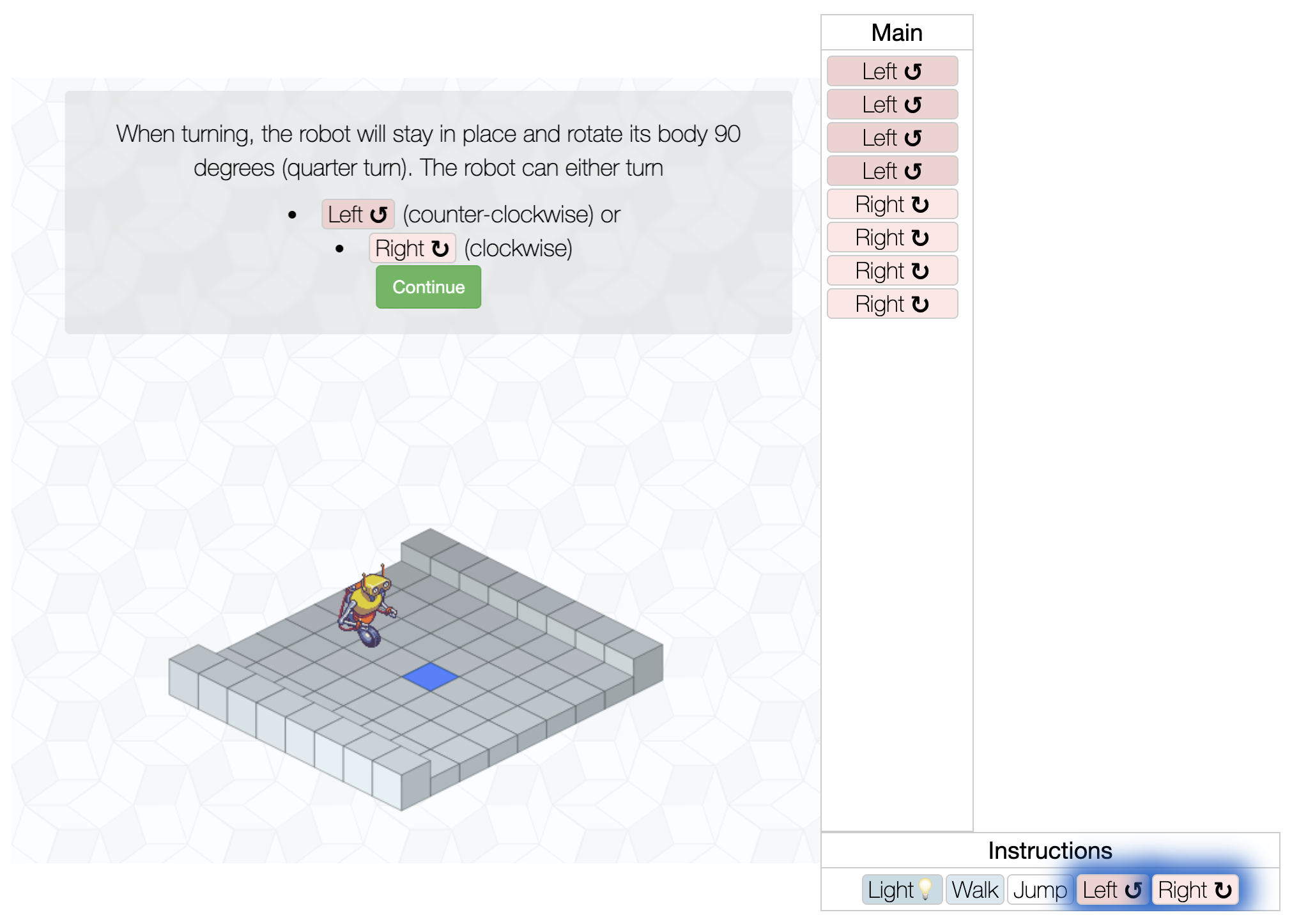}

\caption{Screenshots of the experiment tutorial, continued from Fig.~\ref{fig:experiment-tutorial1}.}
\label{fig:experiment-tutorial2}
\end{figure}

\begin{figure}[hp!]
\centering

\tutorialfig{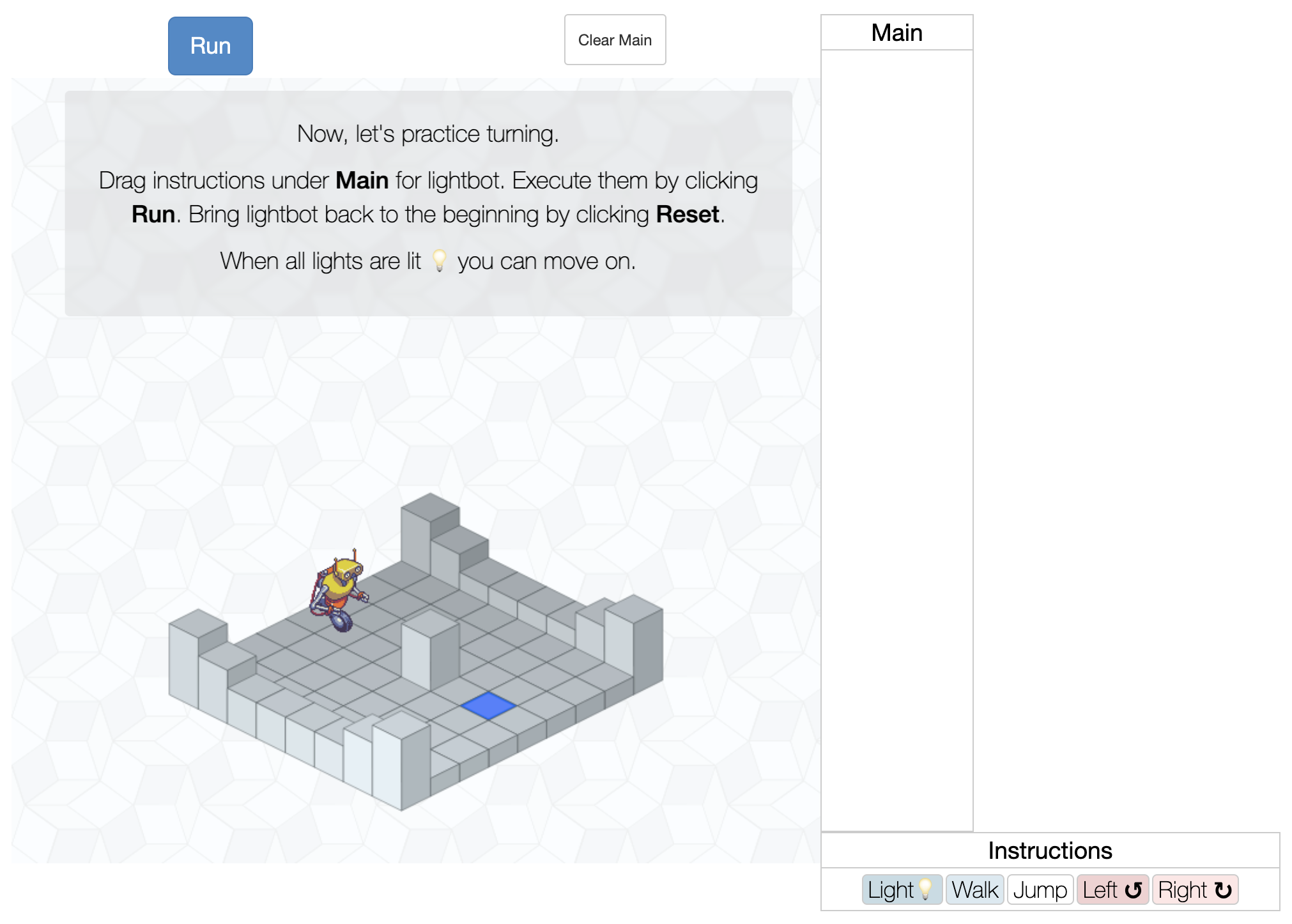}
\tutorialfig{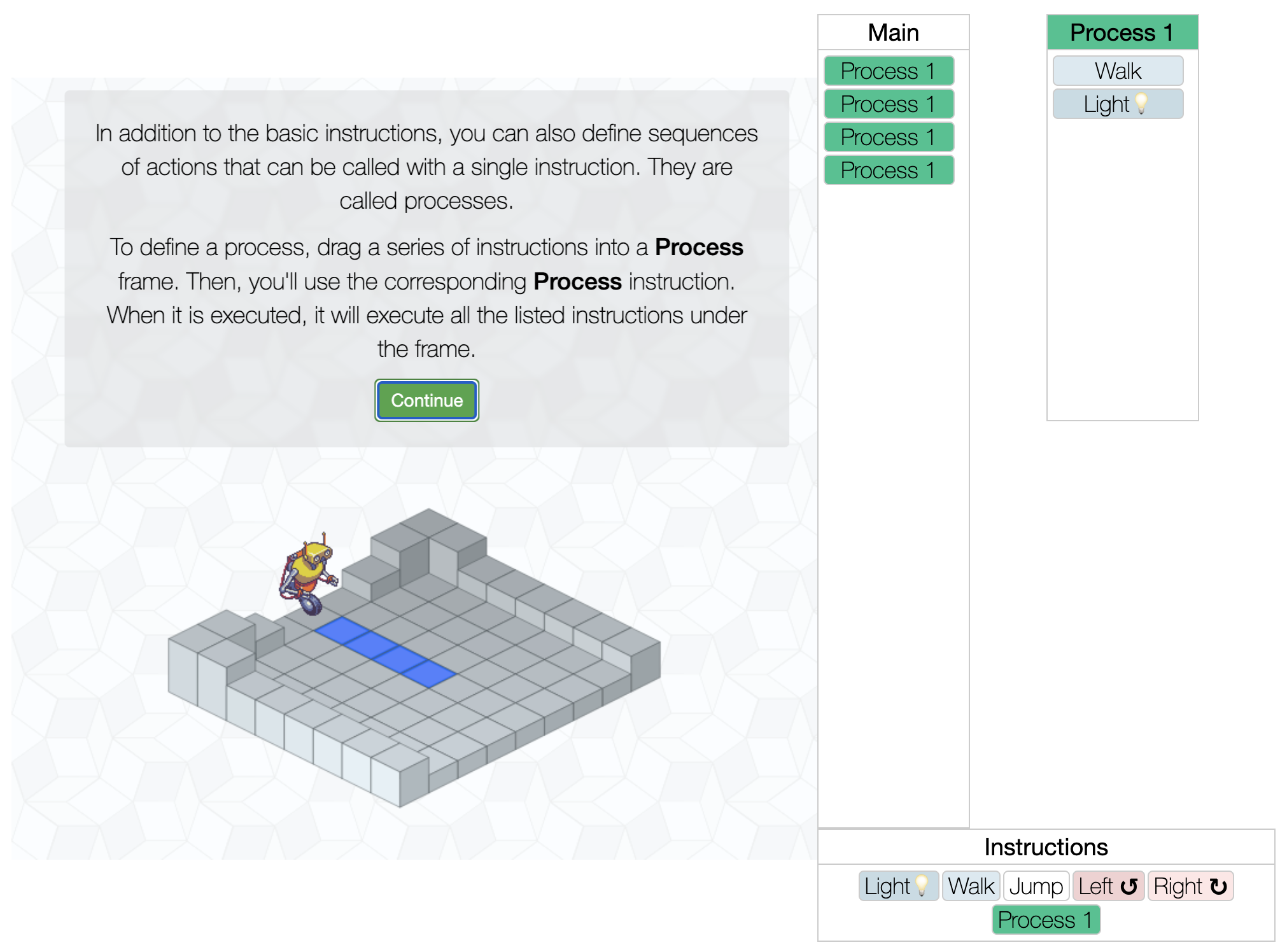}

\tutorialfig{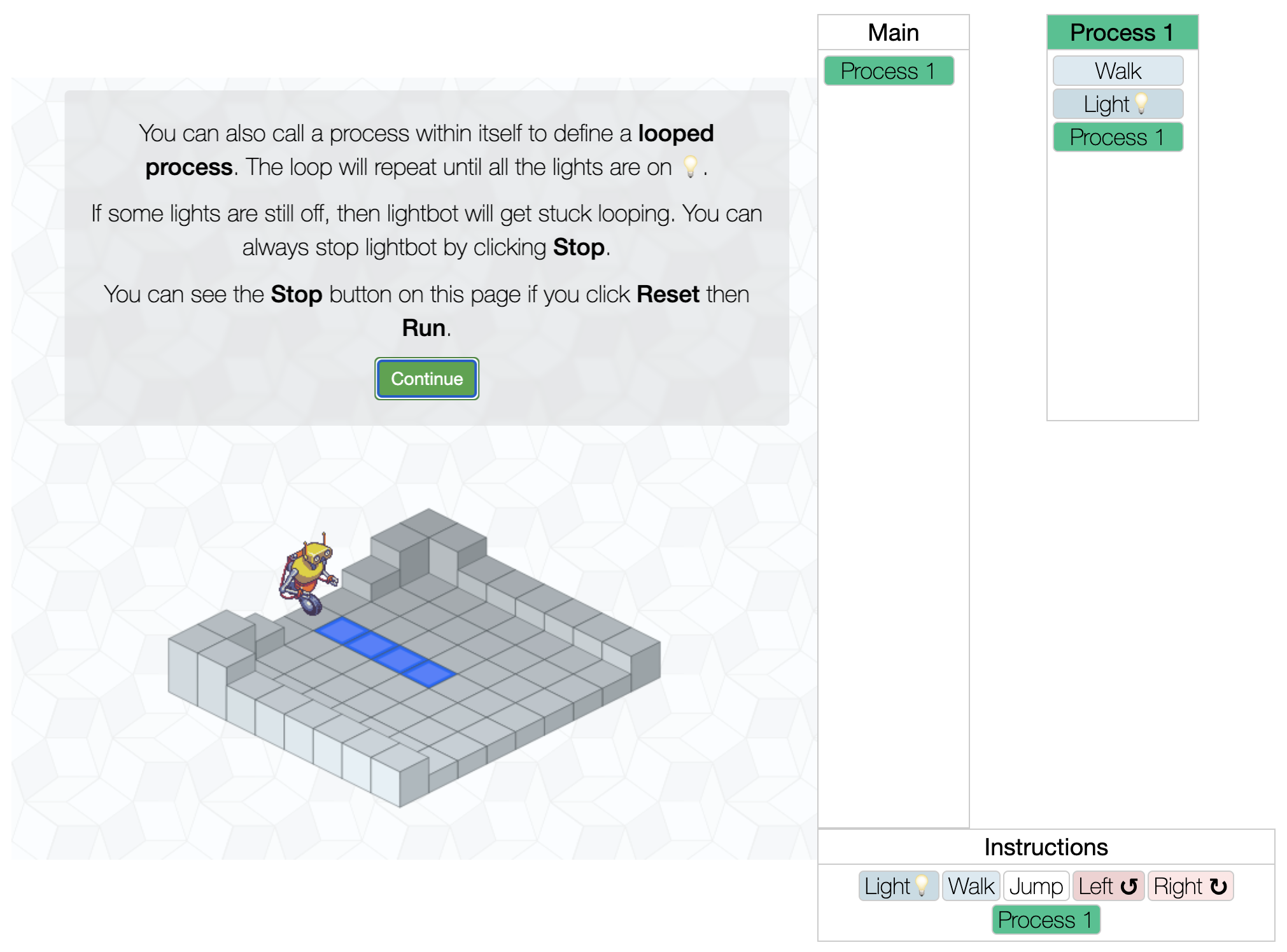}
\tutorialfig{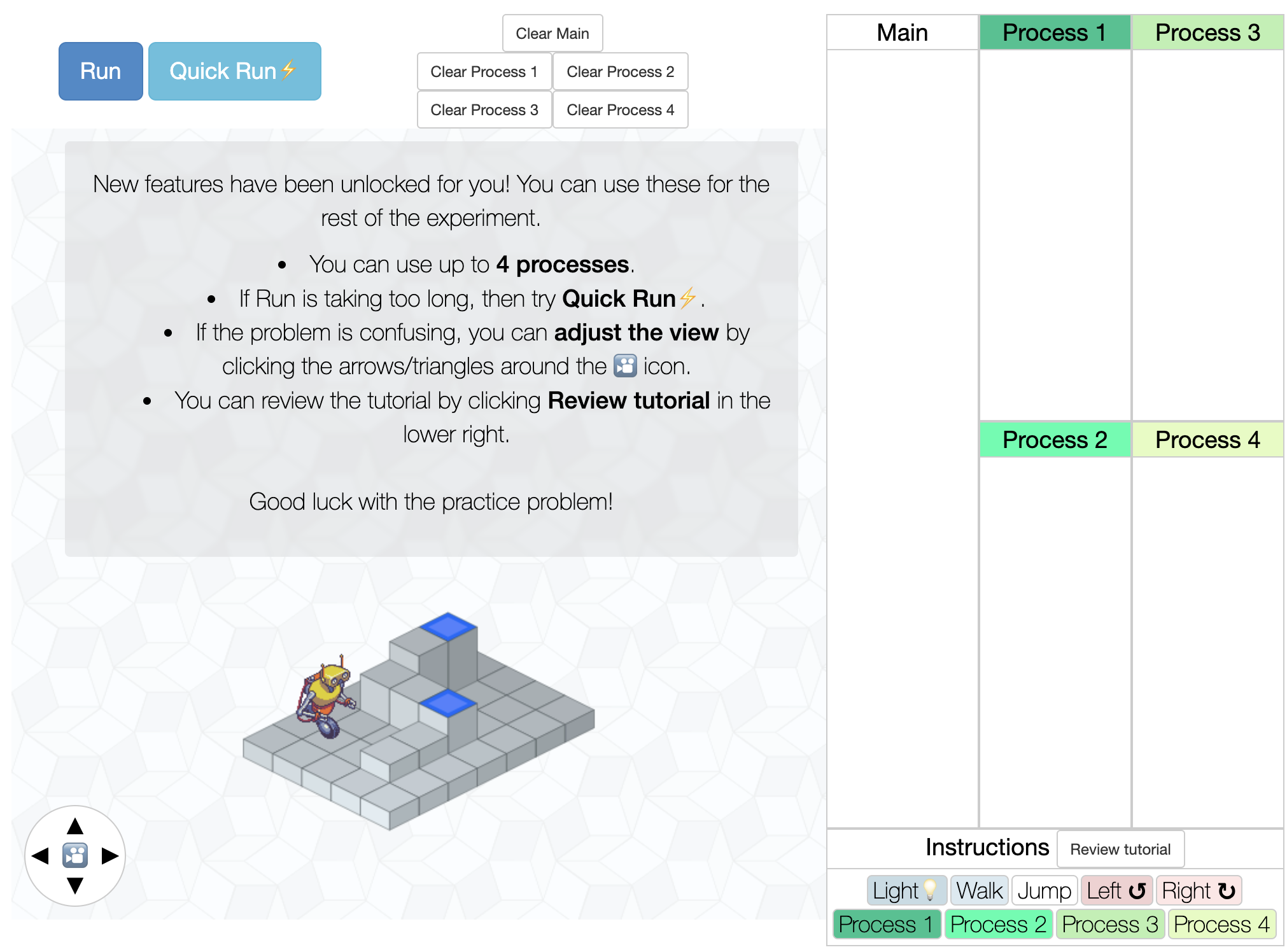}

\caption{Screenshots of the experiment tutorial, continued from Fig.~\ref{fig:experiment-tutorial2}.}
\label{fig:experiment-tutorial3}
\end{figure}

\begin{figure}[hp!]
\centering

\tutorialfig{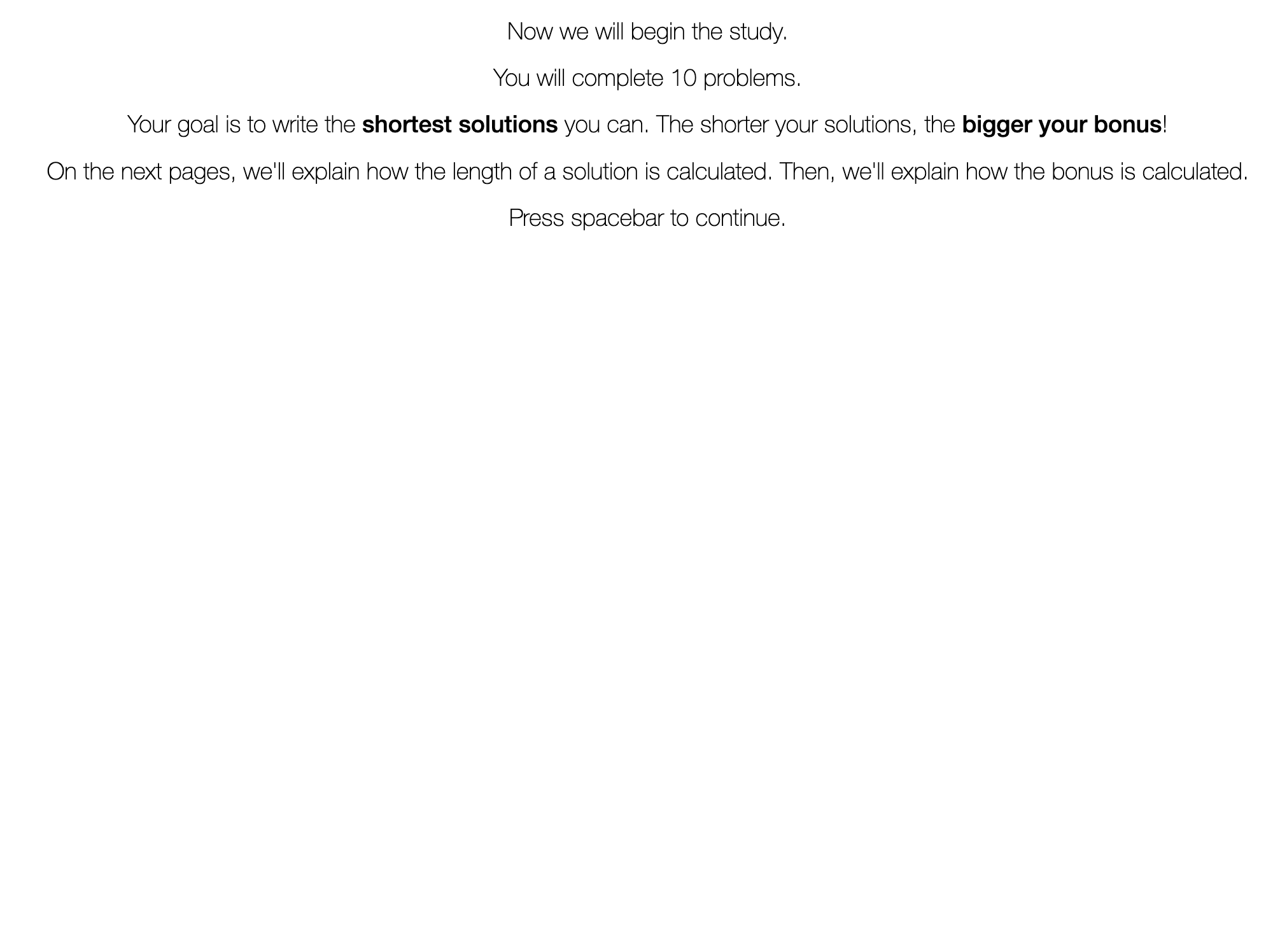}
\tutorialfig{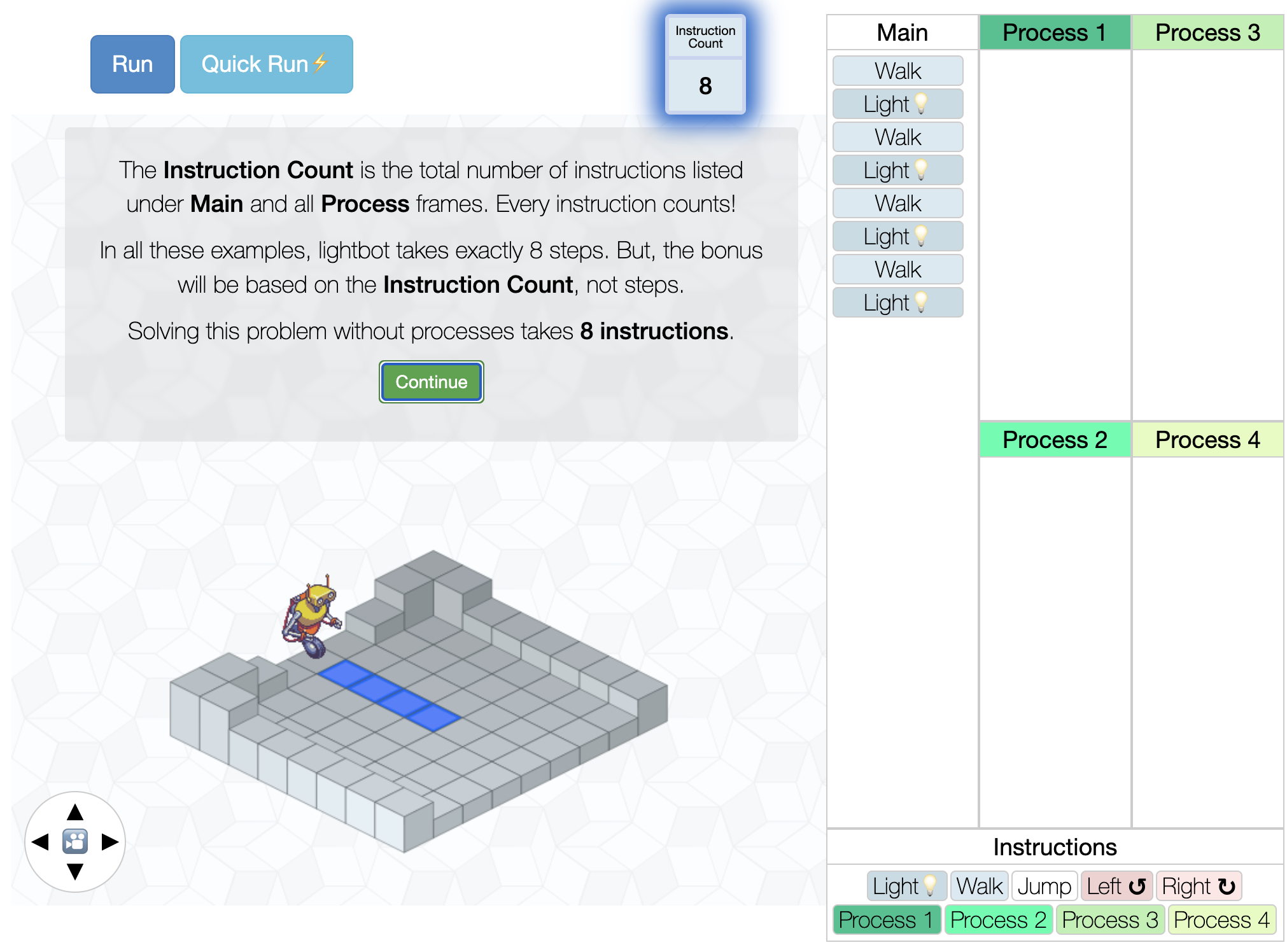}

\tutorialfig{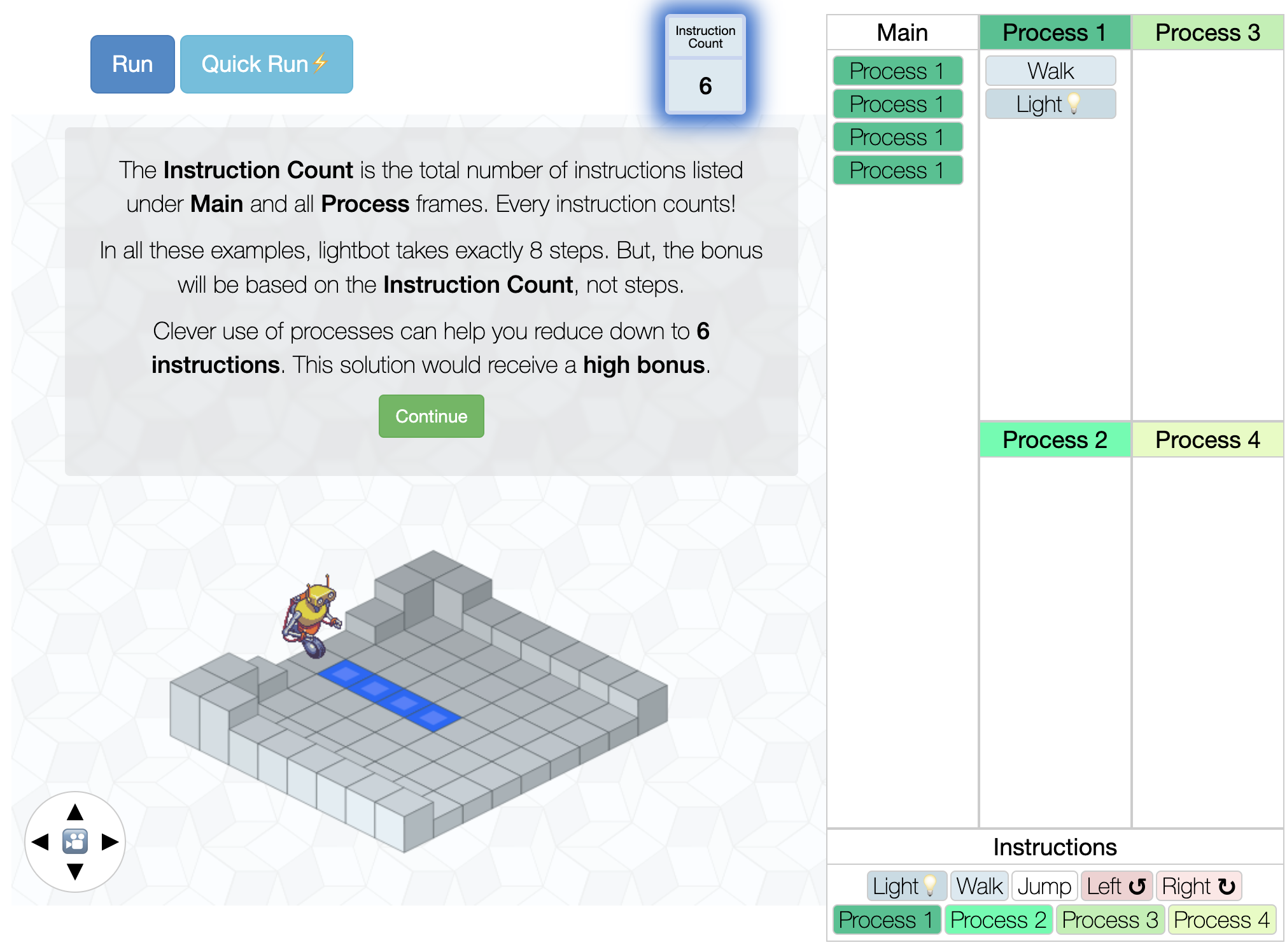}
\tutorialfig{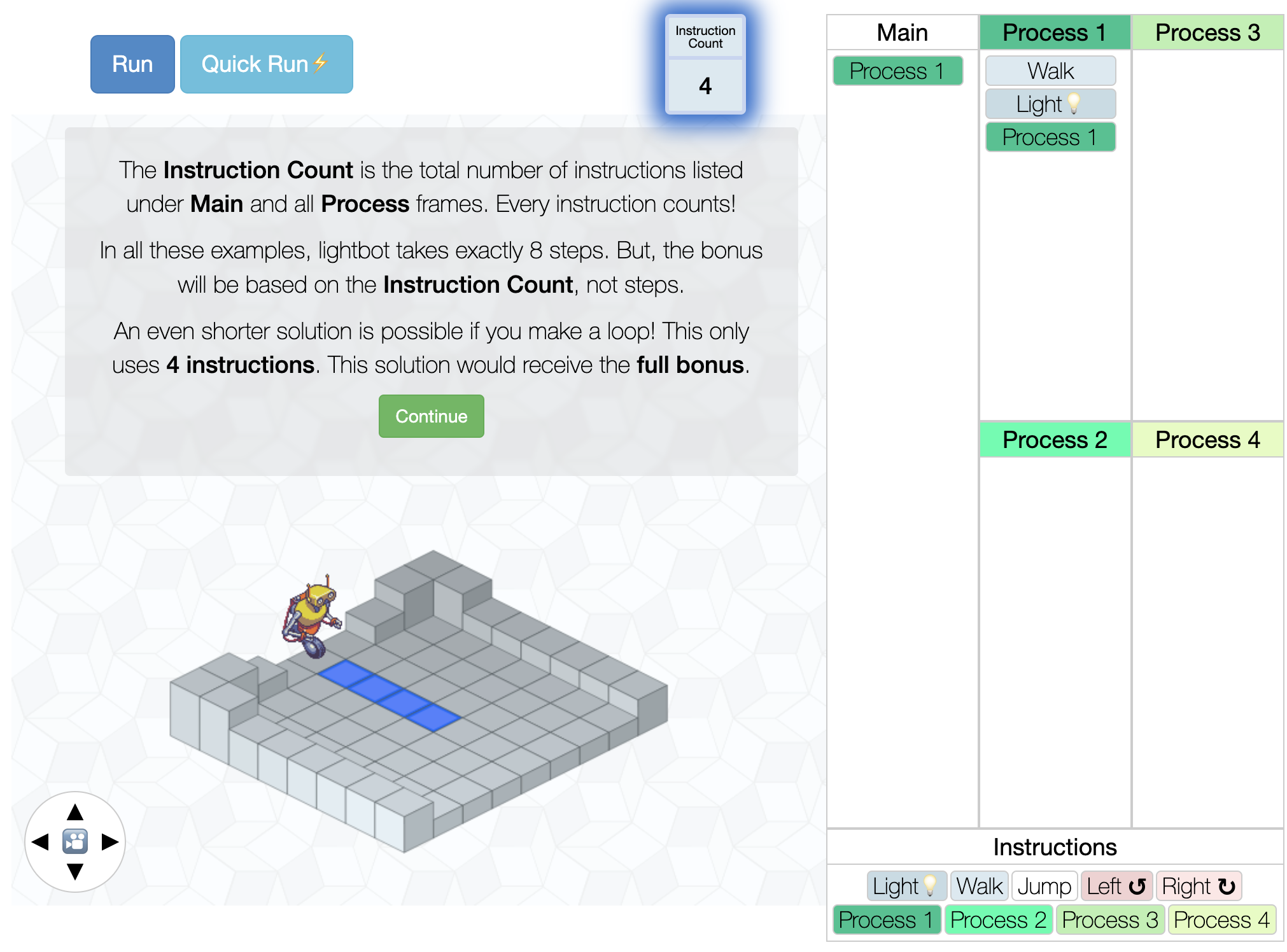}

\tutorialfig{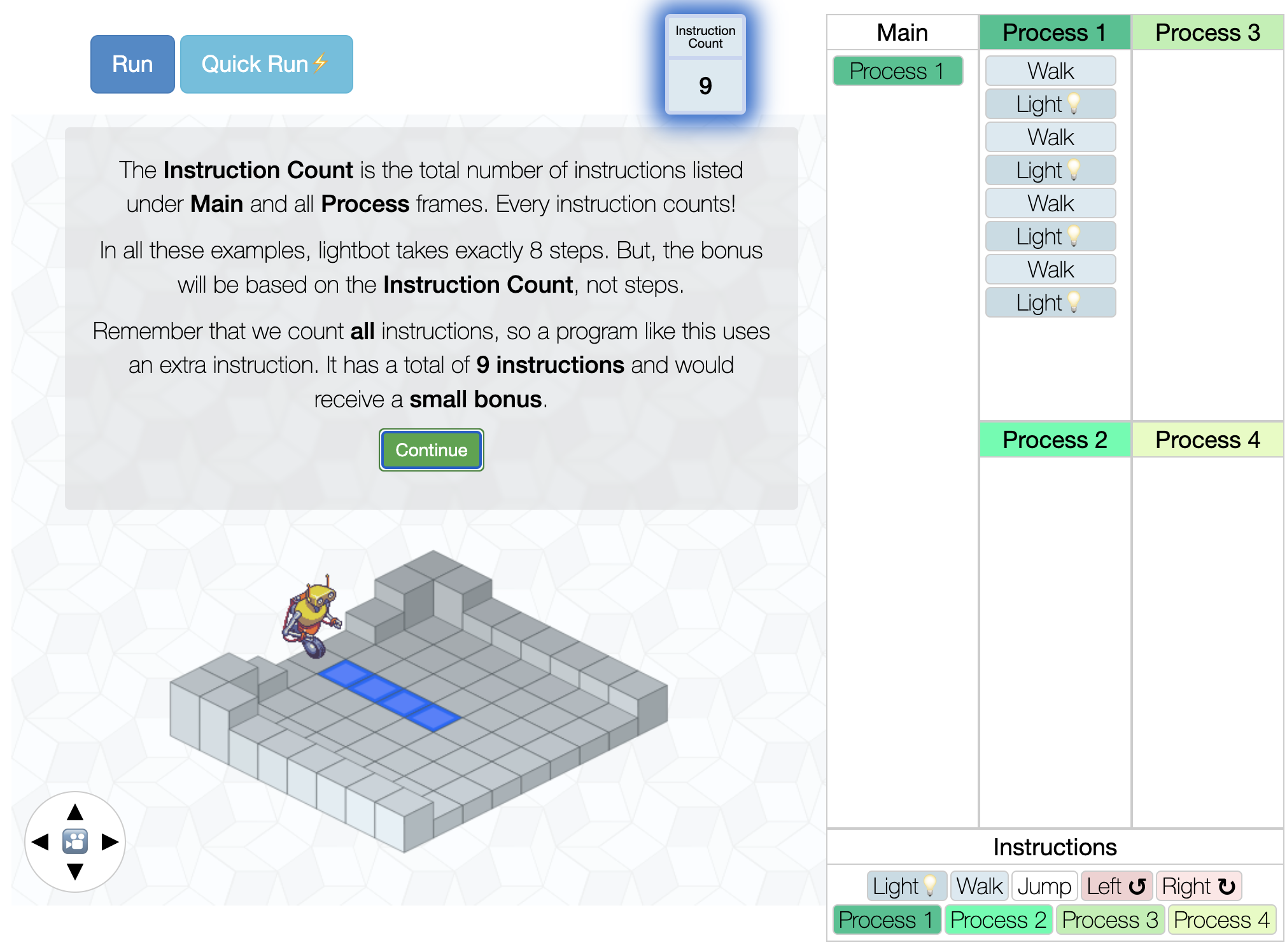}
\tutorialfig{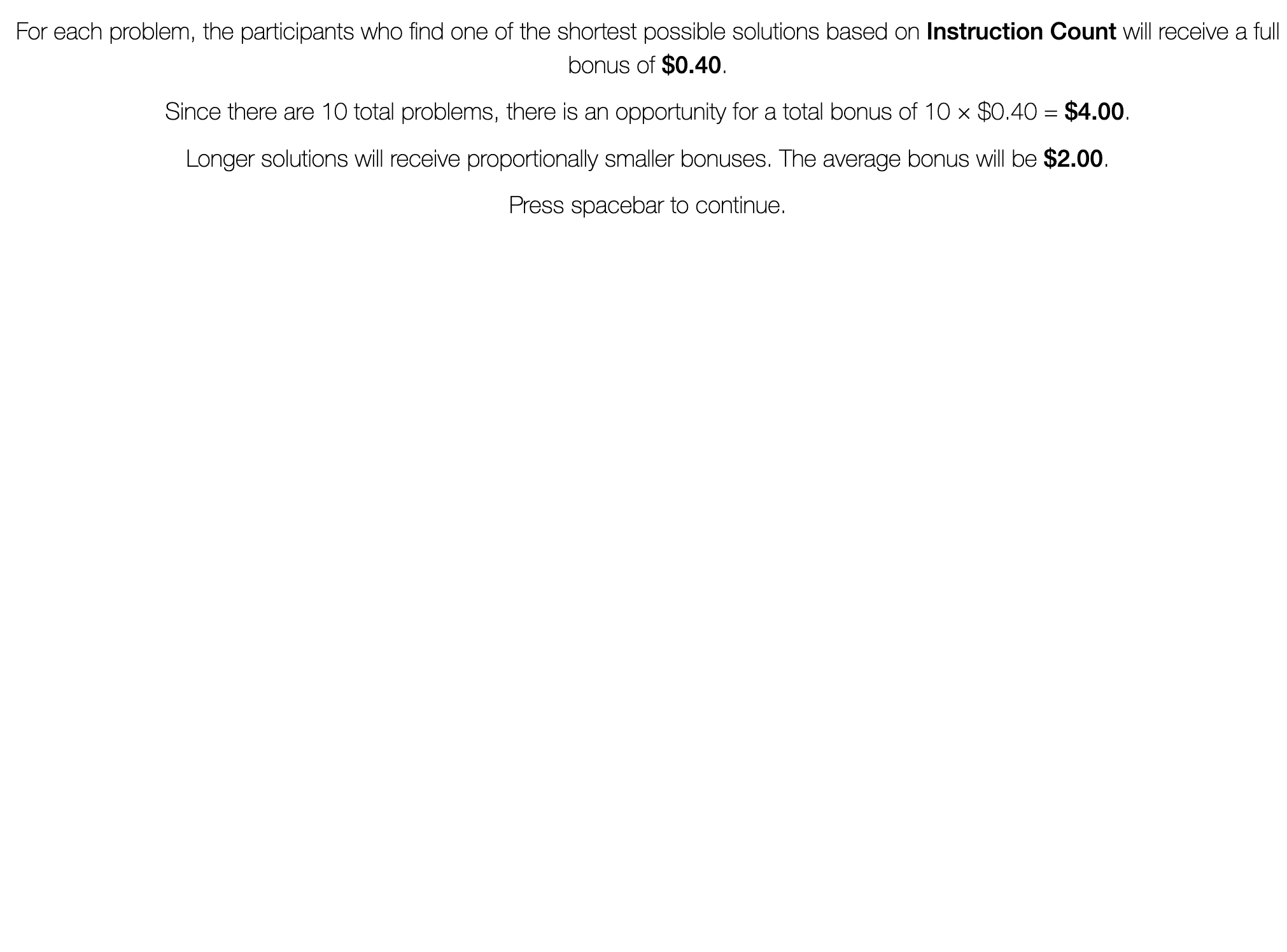}

\caption{Screenshots of the experiment tutorial, continued from Fig.~\ref{fig:experiment-tutorial3}.}
\label{fig:experiment-tutorial4}
\end{figure}

\clearpage

\bibliographystyle{elsarticle-harv}

\begin{thebibliography}{73}
\expandafter\ifx\csname natexlab\endcsname\relax\def\natexlab#1{#1}\fi
\providecommand{\url}[1]{\texttt{#1}}
\providecommand{\href}[2]{#2}
\providecommand{\path}[1]{#1}
\providecommand{\DOIprefix}{doi:}
\providecommand{\ArXivprefix}{arXiv:}
\providecommand{\URLprefix}{URL: }
\providecommand{\Pubmedprefix}{pmid:}
\providecommand{\doi}[1]{\href{http://dx.doi.org/#1}{\path{#1}}}
\providecommand{\Pubmed}[1]{\href{pmid:#1}{\path{#1}}}
\providecommand{\bibinfo}[2]{#2}
\ifx\xfnm\relax \def\xfnm[#1]{\unskip,\space#1}\fi
\bibitem[{Acuna et~al.(2014)Acuna, Wymbs, Reynolds, Picard, Turner, Strick,
  Grafton and Kording}]{acuna2014multifaceted}
\bibinfo{author}{Acuna, D.E.}, \bibinfo{author}{Wymbs, N.F.},
  \bibinfo{author}{Reynolds, C.A.}, \bibinfo{author}{Picard, N.},
  \bibinfo{author}{Turner, R.S.}, \bibinfo{author}{Strick, P.L.},
  \bibinfo{author}{Grafton, S.T.}, \bibinfo{author}{Kording, K.P.},
  \bibinfo{year}{2014}.
\newblock \bibinfo{title}{Multifaceted aspects of chunking enable robust
  algorithms}.
\newblock \bibinfo{journal}{Journal of Neurophysiology} \bibinfo{volume}{112},
  \bibinfo{pages}{1849--1856}.
\bibitem[{Aldous(1985)}]{aldous1985exchangeability}
\bibinfo{author}{Aldous, D.J.}, \bibinfo{year}{1985}.
\newblock \bibinfo{title}{Exchangeability and related topics}, in:
  \bibinfo{editor}{Aldous, D.J.}, \bibinfo{editor}{Ibragimov, I.A.},
  \bibinfo{editor}{Jacod, J.}, \bibinfo{editor}{Hennequin, P.L.} (Eds.),
  \bibinfo{booktitle}{École d’Été de Probabilités de Saint-Flour
  XIII---1983}, \bibinfo{publisher}{Springer}, \bibinfo{address}{Berlin,
  Heidelberg}. pp. \bibinfo{pages}{1--198}.
\bibitem[{Anderson(1991)}]{anderson1991adaptive}
\bibinfo{author}{Anderson, J.R.}, \bibinfo{year}{1991}.
\newblock \bibinfo{title}{The adaptive nature of human categorization}.
\newblock \bibinfo{journal}{Psychological Review} \bibinfo{volume}{98},
  \bibinfo{pages}{409–429}.
\bibitem[{Balaguer et~al.(2016)Balaguer, Spiers, Hassabis and
  Summerfield}]{balaguer2016neural}
\bibinfo{author}{Balaguer, J.}, \bibinfo{author}{Spiers, H.},
  \bibinfo{author}{Hassabis, D.}, \bibinfo{author}{Summerfield, C.},
  \bibinfo{year}{2016}.
\newblock \bibinfo{title}{Neural mechanisms of hierarchical planning in a
  virtual subway network}.
\newblock \bibinfo{journal}{Neuron} \bibinfo{volume}{90},
  \bibinfo{pages}{893--903}.
\bibitem[{Bear et~al.(2020)Bear, Bensinger, Jara-Ettinger, Knobe and
  Cushman}]{bear2020what}
\bibinfo{author}{Bear, A.}, \bibinfo{author}{Bensinger, S.},
  \bibinfo{author}{Jara-Ettinger, J.}, \bibinfo{author}{Knobe, J.},
  \bibinfo{author}{Cushman, F.}, \bibinfo{year}{2020}.
\newblock \bibinfo{title}{What comes to mind?}
\newblock \bibinfo{journal}{Cognition} \bibinfo{volume}{194},
  \bibinfo{pages}{104057}.
\bibitem[{Blei and Frazier(2011)}]{blei2011distance}
\bibinfo{author}{Blei, D.M.}, \bibinfo{author}{Frazier, P.I.},
  \bibinfo{year}{2011}.
\newblock \bibinfo{title}{Distance dependent chinese restaurant processes}.
\newblock \bibinfo{journal}{The Journal of Machine Learning Research}
  \bibinfo{volume}{12}, \bibinfo{pages}{2461–2488}.
\bibitem[{Botvinick et~al.(2009)Botvinick, Niv and Barto}]{botvinick2009}
\bibinfo{author}{Botvinick, M.M.}, \bibinfo{author}{Niv, Y.},
  \bibinfo{author}{Barto, A.G.}, \bibinfo{year}{2009}.
\newblock \bibinfo{title}{Hierarchically organized behavior and its neural
  foundations: A reinforcement learning perspective}.
\newblock \bibinfo{journal}{Cognition} \bibinfo{volume}{113},
  \bibinfo{pages}{262--280}.
\bibitem[{Buchsbaum et~al.(2015)Buchsbaum, Griffiths, Plunkett, Gopnik and
  Baldwin}]{buchsbaum2015inferring}
\bibinfo{author}{Buchsbaum, D.}, \bibinfo{author}{Griffiths, T.L.},
  \bibinfo{author}{Plunkett, D.}, \bibinfo{author}{Gopnik, A.},
  \bibinfo{author}{Baldwin, D.}, \bibinfo{year}{2015}.
\newblock \bibinfo{title}{Inferring action structure and causal relationships
  in continuous sequences of human action}.
\newblock \bibinfo{journal}{Cognitive Psychology} \bibinfo{volume}{76},
  \bibinfo{pages}{30--77}.
\bibitem[{Chater(1999)}]{chater1999search}
\bibinfo{author}{Chater, N.}, \bibinfo{year}{1999}.
\newblock \bibinfo{title}{The search for simplicity: A fundamental cognitive
  principle?}
\newblock \bibinfo{journal}{The Quarterly Journal of Experimental Psychology
  Section A} \bibinfo{volume}{52}, \bibinfo{pages}{273--302}.
\bibitem[{Correa et~al.(2023)Correa, Ho, Callaway, Daw and
  Griffiths}]{correa2023humans}
\bibinfo{author}{Correa, C.G.}, \bibinfo{author}{Ho, M.K.},
  \bibinfo{author}{Callaway, F.}, \bibinfo{author}{Daw, N.D.},
  \bibinfo{author}{Griffiths, T.L.}, \bibinfo{year}{2023}.
\newblock \bibinfo{title}{Humans decompose tasks by trading off utility and
  computational cost}.
\newblock \bibinfo{journal}{PLOS Computational Biology} \bibinfo{volume}{19},
  \bibinfo{pages}{1--31}.
\bibitem[{Dasgupta and Griffiths(2022)}]{dasgupta2022clustering}
\bibinfo{author}{Dasgupta, I.}, \bibinfo{author}{Griffiths, T.L.},
  \bibinfo{year}{2022}.
\newblock \bibinfo{title}{Clustering and the efficient use of cognitive
  resources}.
\newblock \bibinfo{journal}{Journal of Mathematical Psychology}
  \bibinfo{volume}{109}, \bibinfo{pages}{102675}.
\bibitem[{Daw and Courville(2007)}]{daw2007pigeon}
\bibinfo{author}{Daw, N.D.}, \bibinfo{author}{Courville, A.C.},
  \bibinfo{year}{2007}.
\newblock \bibinfo{title}{The pigeon as particle filter}, in:
  \bibinfo{booktitle}{Proceedings of the 20th International Conference on
  Neural Information Processing Systems}, pp. \bibinfo{pages}{369--376}.
\bibitem[{Dezfouli and Balleine(2012)}]{dezfouli2012habits}
\bibinfo{author}{Dezfouli, A.}, \bibinfo{author}{Balleine, B.W.},
  \bibinfo{year}{2012}.
\newblock \bibinfo{title}{Habits, action sequences and reinforcement learning}.
\newblock \bibinfo{journal}{European Journal of Neuroscience}
  \bibinfo{volume}{35}, \bibinfo{pages}{1036--1051}.
\bibitem[{Doucet et~al.(2001)Doucet, De~Freitas, Gordon
  et~al.}]{doucet2001sequential}
\bibinfo{author}{Doucet, A.}, \bibinfo{author}{De~Freitas, N.},
  \bibinfo{author}{Gordon, N.J.}, et~al., \bibinfo{year}{2001}.
\newblock \bibinfo{title}{Sequential Monte Carlo Methods in Practice}.
\newblock \bibinfo{publisher}{Springer}.
\bibitem[{Duncan and Shohamy(2016)}]{duncan2016memory}
\bibinfo{author}{Duncan, K.D.}, \bibinfo{author}{Shohamy, D.},
  \bibinfo{year}{2016}.
\newblock \bibinfo{title}{Memory states influence value-based decisions.}
\newblock \bibinfo{journal}{Journal of Experimental Psychology: General}
  \bibinfo{volume}{145}, \bibinfo{pages}{1420--1426}.
\bibitem[{Eckstein and Collins(2020)}]{eckstein2020computational}
\bibinfo{author}{Eckstein, M.K.}, \bibinfo{author}{Collins, A.G.E.},
  \bibinfo{year}{2020}.
\newblock \bibinfo{title}{Computational evidence for hierarchically structured
  reinforcement learning in humans}.
\newblock \bibinfo{journal}{Proceedings of the National Academy of Sciences}
  \bibinfo{volume}{117}, \bibinfo{pages}{29381--29389}.
\bibitem[{Ellis et~al.(2021)Ellis, Wong, Nye, Sabl\'{e}-Meyer, Morales, Hewitt,
  Cary, Solar-Lezama and Tenenbaum}]{ellis2021dreamcoder}
\bibinfo{author}{Ellis, K.}, \bibinfo{author}{Wong, C.}, \bibinfo{author}{Nye,
  M.}, \bibinfo{author}{Sabl\'{e}-Meyer, M.}, \bibinfo{author}{Morales, L.},
  \bibinfo{author}{Hewitt, L.}, \bibinfo{author}{Cary, L.},
  \bibinfo{author}{Solar-Lezama, A.}, \bibinfo{author}{Tenenbaum, J.B.},
  \bibinfo{year}{2021}.
\newblock \bibinfo{title}{Dreamcoder: Bootstrapping inductive program synthesis
  with wake-sleep library learning}, in: \bibinfo{booktitle}{Proceedings of the
  42nd ACM SIGPLAN International Conference on Programming Language Design and
  Implementation}, \bibinfo{publisher}{Association for Computing Machinery},
  \bibinfo{address}{New York, NY, USA}. pp. \bibinfo{pages}{835--850}.
\bibitem[{Flerova et~al.(2016)Flerova, Marinescu and
  Dechter}]{flerova2016searching}
\bibinfo{author}{Flerova, N.}, \bibinfo{author}{Marinescu, R.},
  \bibinfo{author}{Dechter, R.}, \bibinfo{year}{2016}.
\newblock \bibinfo{title}{Searching for the m best solutions in graphical
  models}.
\newblock \bibinfo{journal}{Journal of Artificial Intelligence Research}
  \bibinfo{volume}{55}, \bibinfo{pages}{889--952}.
\bibitem[{Fränken et~al.(2022)Fränken, Theodoropoulos and
  Bramley}]{franken2022algorithms}
\bibinfo{author}{Fränken, J.P.}, \bibinfo{author}{Theodoropoulos, N.C.},
  \bibinfo{author}{Bramley, N.R.}, \bibinfo{year}{2022}.
\newblock \bibinfo{title}{Algorithms of adaptation in inductive inference}.
\newblock \bibinfo{journal}{Cognitive Psychology} \bibinfo{volume}{137},
  \bibinfo{pages}{101506}.
\bibitem[{Gershman et~al.(2010)Gershman, Blei and Niv}]{gershman2010context}
\bibinfo{author}{Gershman, S.J.}, \bibinfo{author}{Blei, D.M.},
  \bibinfo{author}{Niv, Y.}, \bibinfo{year}{2010}.
\newblock \bibinfo{title}{Context, learning, and extinction}.
\newblock \bibinfo{journal}{Psychological Review} \bibinfo{volume}{117},
  \bibinfo{pages}{197–209}.
\bibitem[{Goldwater et~al.(2009)Goldwater, Griffiths and
  Johnson}]{goldwater2009bayesian}
\bibinfo{author}{Goldwater, S.}, \bibinfo{author}{Griffiths, T.L.},
  \bibinfo{author}{Johnson, M.}, \bibinfo{year}{2009}.
\newblock \bibinfo{title}{A bayesian framework for word segmentation: Exploring
  the effects of context}.
\newblock \bibinfo{journal}{Cognition} \bibinfo{volume}{112},
  \bibinfo{pages}{21--54}.
\bibitem[{Goodman et~al.(2008a)Goodman, Mansinghka, Roy, Bonawitz and
  Tenenbaum}]{goodman2008church}
\bibinfo{author}{Goodman, N.D.}, \bibinfo{author}{Mansinghka, V.K.},
  \bibinfo{author}{Roy, D.}, \bibinfo{author}{Bonawitz, K.},
  \bibinfo{author}{Tenenbaum, J.B.}, \bibinfo{year}{2008}a.
\newblock \bibinfo{title}{Church: A language for generative models}, in:
  \bibinfo{booktitle}{Proceedings of the Twenty-Fourth Conference on
  Uncertainty in Artificial Intelligence}, \bibinfo{publisher}{AUAI Press}. pp.
  \bibinfo{pages}{220--229}.
\bibitem[{Goodman et~al.(2008b)Goodman, Tenenbaum, Feldman and
  Griffiths}]{goodman2008rational}
\bibinfo{author}{Goodman, N.D.}, \bibinfo{author}{Tenenbaum, J.B.},
  \bibinfo{author}{Feldman, J.}, \bibinfo{author}{Griffiths, T.L.},
  \bibinfo{year}{2008}b.
\newblock \bibinfo{title}{A rational analysis of rule-based concept learning}.
\newblock \bibinfo{journal}{Cognitive Science} \bibinfo{volume}{32},
  \bibinfo{pages}{108--154}.
\bibitem[{Hamrick et~al.(2015)Hamrick, Smith, Griffiths and
  Vul}]{hamrick2015think}
\bibinfo{author}{Hamrick, J.B.}, \bibinfo{author}{Smith, K.A.},
  \bibinfo{author}{Griffiths, T.L.}, \bibinfo{author}{Vul, E.},
  \bibinfo{year}{2015}.
\newblock \bibinfo{title}{Think again? the amount of mental simulation tracks
  uncertainty in the outcome.}, in: \bibinfo{booktitle}{Proceedings of the 37th
  Annual Conference of the Cognitive Science Society}.
\bibitem[{Hart et~al.(1968)Hart, Nilsson and Raphael}]{hart1968formal}
\bibinfo{author}{Hart, P.E.}, \bibinfo{author}{Nilsson, N.J.},
  \bibinfo{author}{Raphael, B.}, \bibinfo{year}{1968}.
\newblock \bibinfo{title}{A formal basis for the heuristic determination of
  minimum cost paths}.
\newblock \bibinfo{journal}{IEEE Transactions on Systems Science and
  Cybernetics} \bibinfo{volume}{4}, \bibinfo{pages}{100--107}.
\bibitem[{Ho et~al.(2018)Ho, Sanborn, Callaway, Bourgin and
  Griffiths}]{ho2018human}
\bibinfo{author}{Ho, M.K.}, \bibinfo{author}{Sanborn, S.},
  \bibinfo{author}{Callaway, F.}, \bibinfo{author}{Bourgin, D.},
  \bibinfo{author}{Griffiths, T.}, \bibinfo{year}{2018}.
\newblock \bibinfo{title}{Human priors in hierarchical program induction}, in:
  \bibinfo{booktitle}{Proceedings of the 2018 Conference on Cognitive
  Computational Neuroscience}.
\bibitem[{Huys et~al.(2015)Huys, Lally, Faulkner, Eshel, Seifritz, Gershman,
  Dayan and Roiser}]{huys_interplay_2015}
\bibinfo{author}{Huys, Q.J.M.}, \bibinfo{author}{Lally, N.},
  \bibinfo{author}{Faulkner, P.}, \bibinfo{author}{Eshel, N.},
  \bibinfo{author}{Seifritz, E.}, \bibinfo{author}{Gershman, S.J.},
  \bibinfo{author}{Dayan, P.}, \bibinfo{author}{Roiser, J.P.},
  \bibinfo{year}{2015}.
\newblock \bibinfo{title}{Interplay of approximate planning strategies}.
\newblock \bibinfo{journal}{Proc. of the National Academy of Sciences}
  \bibinfo{volume}{112}, \bibinfo{pages}{3098--3103}.
\bibitem[{Johnson et~al.(2006)Johnson, Griffiths and
  Goldwater}]{johnson2007adaptor}
\bibinfo{author}{Johnson, M.}, \bibinfo{author}{Griffiths, T.L.},
  \bibinfo{author}{Goldwater, S.}, \bibinfo{year}{2006}.
\newblock \bibinfo{title}{Adaptor grammars: A framework for specifying
  compositional nonparametric bayesian models}.
\newblock \bibinfo{note}{{Advances in Neural Information Processing Systems}}.
\bibitem[{Kemp et~al.(2010)Kemp, Goodman and Tenenbaum}]{kemp2010learning}
\bibinfo{author}{Kemp, C.}, \bibinfo{author}{Goodman, N.D.},
  \bibinfo{author}{Tenenbaum, J.B.}, \bibinfo{year}{2010}.
\newblock \bibinfo{title}{Learning to learn causal models}.
\newblock \bibinfo{journal}{Cognitive Science} \bibinfo{volume}{34},
  \bibinfo{pages}{1185--1243}.
\bibitem[{Klir and Simon(1991)}]{klir1991architecture}
\bibinfo{author}{Klir, G.J.}, \bibinfo{author}{Simon, H.A.},
  \bibinfo{year}{1991}.
\newblock \bibinfo{title}{The architecture of complexity}.
\newblock \bibinfo{publisher}{Springer}.
\bibitem[{Lai et~al.(2022)Lai, Huang and Gershman}]{lai2022chunking}
\bibinfo{author}{Lai, L.}, \bibinfo{author}{Huang, A.Z.},
  \bibinfo{author}{Gershman, S.J.}, \bibinfo{year}{2022}.
\newblock \bibinfo{title}{Action chunking as policy compression}.
\newblock \URLprefix \url{osf.io/preprints/psyarxiv/z8yrv}.
\bibitem[{Lake and Piantadosi(2020)}]{lake2020people}
\bibinfo{author}{Lake, B.M.}, \bibinfo{author}{Piantadosi, S.T.},
  \bibinfo{year}{2020}.
\newblock \bibinfo{title}{People infer recursive visual concepts from just a
  few examples}.
\newblock \bibinfo{journal}{Computational Brain \& Behavior}
  \bibinfo{volume}{3}, \bibinfo{pages}{54--65}.
\bibitem[{Lake et~al.(2015)Lake, Salakhutdinov and Tenenbaum}]{lake2015human}
\bibinfo{author}{Lake, B.M.}, \bibinfo{author}{Salakhutdinov, R.},
  \bibinfo{author}{Tenenbaum, J.B.}, \bibinfo{year}{2015}.
\newblock \bibinfo{title}{Human-level concept learning through probabilistic
  program induction}.
\newblock \bibinfo{journal}{Science} \bibinfo{volume}{350},
  \bibinfo{pages}{1332--1338}.
\bibitem[{Levine(2018)}]{levine2018rlinference}
\bibinfo{author}{Levine, S.}, \bibinfo{year}{2018}.
\newblock \bibinfo{title}{Reinforcement learning and control as probabilistic
  inference: Tutorial and review}.
\newblock \href{http://arxiv.org/abs/1805.00909}{{\tt arXiv:1805.00909}}.
\bibitem[{Lieder et~al.(2018)Lieder, Griffiths and
  Hsu}]{lieder2018overrepresentation}
\bibinfo{author}{Lieder, F.}, \bibinfo{author}{Griffiths, T.L.},
  \bibinfo{author}{Hsu, M.}, \bibinfo{year}{2018}.
\newblock \bibinfo{title}{Overrepresentation of extreme events in decision
  making reflects rational use of cognitive resources}.
\newblock \bibinfo{journal}{Psychological Review} \bibinfo{volume}{125},
  \bibinfo{pages}{1--32}.
\bibitem[{Luce(1959)}]{luce1959choice}
\bibinfo{author}{Luce, R.D.}, \bibinfo{year}{1959}.
\newblock \bibinfo{title}{On the possible psychophysical laws}.
\newblock \bibinfo{journal}{Psychological review} \bibinfo{volume}{66},
  \bibinfo{pages}{81--95}.
\bibitem[{Luong et~al.(2013)Luong, Frank and Johnson}]{luong2013parsing}
\bibinfo{author}{Luong, M.T.}, \bibinfo{author}{Frank, M.C.},
  \bibinfo{author}{Johnson, M.}, \bibinfo{year}{2013}.
\newblock \bibinfo{title}{{Parsing entire discourses as very long strings:
  Capturing topic continuity in grounded language learning}}.
\newblock \bibinfo{journal}{Transactions of the Association for Computational
  Linguistics} \bibinfo{volume}{1}, \bibinfo{pages}{315--326}.
\bibitem[{Maisto et~al.(2015)Maisto, Donnarumma and Pezzulo}]{maisto2015divide}
\bibinfo{author}{Maisto, D.}, \bibinfo{author}{Donnarumma, F.},
  \bibinfo{author}{Pezzulo, G.}, \bibinfo{year}{2015}.
\newblock \bibinfo{title}{\textit{Divide et impera}: subgoaling reduces the
  complexity of probabilistic inference and problem solving}.
\newblock \bibinfo{journal}{Journal of The Royal Society Interface}
  \bibinfo{volume}{12}, \bibinfo{pages}{20141335}.
\bibitem[{Marr(1982)}]{marr1982vision}
\bibinfo{author}{Marr, D.}, \bibinfo{year}{1982}.
\newblock \bibinfo{title}{Vision}.
\newblock \bibinfo{publisher}{W. H. Freeman}.
\bibitem[{Mattar and Daw(2018)}]{mattar2018prioritized}
\bibinfo{author}{Mattar, M.G.}, \bibinfo{author}{Daw, N.D.},
  \bibinfo{year}{2018}.
\newblock \bibinfo{title}{Prioritized memory access explains planning and
  hippocampal replay}.
\newblock \bibinfo{journal}{Nature Neuroscience} \bibinfo{volume}{21},
  \bibinfo{pages}{1609--1617}.
\bibitem[{McNamee et~al.(2016)McNamee, Wolpert and
  Lengyel}]{mcnamee2016efficient}
\bibinfo{author}{McNamee, D.}, \bibinfo{author}{Wolpert, D.M.},
  \bibinfo{author}{Lengyel, M.}, \bibinfo{year}{2016}.
\newblock \bibinfo{title}{Efficient state-space modularization for planning:
  theory, behavioral and neural signatures}, in: \bibinfo{booktitle}{{Advances
  in Neural Information Processing Systems}}.
\bibitem[{Miller et~al.(1960)Miller, Galanter and Pribram}]{miller1960plans}
\bibinfo{author}{Miller, G.}, \bibinfo{author}{Galanter, E.},
  \bibinfo{author}{Pribram, K.}, \bibinfo{year}{1960}.
\newblock \bibinfo{title}{Plans and the Structure of Behavior}.
\newblock \bibinfo{publisher}{Henry Holt and Co}.
\bibitem[{Miller(1956)}]{miller1956magical}
\bibinfo{author}{Miller, G.A.}, \bibinfo{year}{1956}.
\newblock \bibinfo{title}{The magical number seven, plus or minus two: Some
  limits on our capacity for processing information}.
\newblock \bibinfo{journal}{Psychological Review} \bibinfo{volume}{63},
  \bibinfo{pages}{81--97}.
\bibitem[{Morris et~al.(2021)Morris, Phillips, Huang and
  Cushman}]{morris2021generating}
\bibinfo{author}{Morris, A.}, \bibinfo{author}{Phillips, J.},
  \bibinfo{author}{Huang, K.}, \bibinfo{author}{Cushman, F.},
  \bibinfo{year}{2021}.
\newblock \bibinfo{title}{Generating options and choosing between them depend
  on distinct forms of value representation}.
\newblock \bibinfo{journal}{Psychological Science} \bibinfo{volume}{32},
  \bibinfo{pages}{1731--1746}.
\bibitem[{Newell and Simon(1972)}]{newell1972human}
\bibinfo{author}{Newell, A.}, \bibinfo{author}{Simon, H.A.},
  \bibinfo{year}{1972}.
\newblock \bibinfo{title}{Human problem solving}.
\newblock \bibinfo{publisher}{Prentice-Hall}.
\bibitem[{O'Donnell(2015)}]{odonnell2015productivity}
\bibinfo{author}{O'Donnell, T.J.}, \bibinfo{year}{2015}.
\newblock \bibinfo{title}{{Productivity and Reuse In Language: A Theory of
  Linguistic Computation and Storage}}.
\newblock \bibinfo{publisher}{The MIT Press}.
\bibitem[{Perruchet and Pacton(2006)}]{perruchet2006implicit}
\bibinfo{author}{Perruchet, P.}, \bibinfo{author}{Pacton, S.},
  \bibinfo{year}{2006}.
\newblock \bibinfo{title}{Implicit learning and statistical learning: one
  phenomenon, two approaches}.
\newblock \bibinfo{journal}{Trends in Cognitive Sciences} \bibinfo{volume}{10},
  \bibinfo{pages}{233--238}.
\bibitem[{Piantadosi et~al.(2012)Piantadosi, Tenenbaum and
  Goodman}]{piantadosi2012bootstrapping}
\bibinfo{author}{Piantadosi, S.T.}, \bibinfo{author}{Tenenbaum, J.B.},
  \bibinfo{author}{Goodman, N.D.}, \bibinfo{year}{2012}.
\newblock \bibinfo{title}{Bootstrapping in a language of thought: A formal
  model of numerical concept learning}.
\newblock \bibinfo{journal}{Cognition} \bibinfo{volume}{123},
  \bibinfo{pages}{199--217}.
\bibitem[{Planton et~al.(2021)Planton, van Kerkoerle, Abbih, Maheu, Meyniel,
  Sigman, Wang, Figueira, Romano and Dehaene}]{planton2021theory}
\bibinfo{author}{Planton, S.}, \bibinfo{author}{van Kerkoerle, T.},
  \bibinfo{author}{Abbih, L.}, \bibinfo{author}{Maheu, M.},
  \bibinfo{author}{Meyniel, F.}, \bibinfo{author}{Sigman, M.},
  \bibinfo{author}{Wang, L.}, \bibinfo{author}{Figueira, S.},
  \bibinfo{author}{Romano, S.}, \bibinfo{author}{Dehaene, S.},
  \bibinfo{year}{2021}.
\newblock \bibinfo{title}{A theory of memory for binary sequences: Evidence for
  a mental compression algorithm in humans}.
\newblock \bibinfo{journal}{PLOS Computational Biology} \bibinfo{volume}{17},
  \bibinfo{pages}{1--43}.
\bibitem[{Poesia and Goodman(2023)}]{poesia2023peano}
\bibinfo{author}{Poesia, G.}, \bibinfo{author}{Goodman, N.D.},
  \bibinfo{year}{2023}.
\newblock \bibinfo{title}{Peano: learning formal mathematical reasoning}.
\newblock \bibinfo{journal}{Philosophical Transactions of the Royal Society A:
  Mathematical, Physical and Engineering Sciences} \bibinfo{volume}{381},
  \bibinfo{pages}{20220044}.
\bibitem[{Ramkumar et~al.(2016)Ramkumar, Acuna, Berniker, Grafton, Turner and
  Kording}]{ramkumar2016chunking}
\bibinfo{author}{Ramkumar, P.}, \bibinfo{author}{Acuna, D.E.},
  \bibinfo{author}{Berniker, M.}, \bibinfo{author}{Grafton, S.T.},
  \bibinfo{author}{Turner, R.S.}, \bibinfo{author}{Kording, K.P.},
  \bibinfo{year}{2016}.
\newblock \bibinfo{title}{Chunking as the result of an efficiency computation
  trade-off}.
\newblock \bibinfo{journal}{Nature communications} \bibinfo{volume}{7},
  \bibinfo{pages}{1--11}.
\bibitem[{Ribas-Fernandes et~al.(2011)Ribas-Fernandes, Solway, Diuk, McGuire,
  Barto, Niv and Botvinick}]{ribasfernandes2011neural}
\bibinfo{author}{Ribas-Fernandes, J.J.}, \bibinfo{author}{Solway, A.},
  \bibinfo{author}{Diuk, C.}, \bibinfo{author}{McGuire, J.T.},
  \bibinfo{author}{Barto, A.G.}, \bibinfo{author}{Niv, Y.},
  \bibinfo{author}{Botvinick, M.M.}, \bibinfo{year}{2011}.
\newblock \bibinfo{title}{A neural signature of hierarchical reinforcement
  learning}.
\newblock \bibinfo{journal}{Neuron} \bibinfo{volume}{71},
  \bibinfo{pages}{370--379}.
\bibitem[{Rosenbaum et~al.(1983)Rosenbaum, Kenny and
  Derr}]{rosenbaum1983hierarchical}
\bibinfo{author}{Rosenbaum, D.A.}, \bibinfo{author}{Kenny, S.B.},
  \bibinfo{author}{Derr, M.A.}, \bibinfo{year}{1983}.
\newblock \bibinfo{title}{Hierarchical control of rapid movement sequences}.
\newblock \bibinfo{journal}{Journal of experimental psychology. Human
  perception and performance} \bibinfo{volume}{9}, \bibinfo{pages}{86--102}.
\bibitem[{Rule et~al.(2018)Rule, Schulz, Piantadosi and
  Tenenbaum}]{rule2018learning}
\bibinfo{author}{Rule, J.}, \bibinfo{author}{Schulz, E.},
  \bibinfo{author}{Piantadosi, S.T.}, \bibinfo{author}{Tenenbaum, J.B.},
  \bibinfo{year}{2018}.
\newblock \bibinfo{title}{Learning list concepts through program induction},
  in: \bibinfo{booktitle}{Proceedings of the Annual Meeting of the Cognitive
  Science Society}.
\bibitem[{Rule et~al.(2020)Rule, Tenenbaum and Piantadosi}]{rule2020child}
\bibinfo{author}{Rule, J.S.}, \bibinfo{author}{Tenenbaum, J.B.},
  \bibinfo{author}{Piantadosi, S.T.}, \bibinfo{year}{2020}.
\newblock \bibinfo{title}{The child as hacker}.
\newblock \bibinfo{journal}{Trends in Cognitive Sciences} \bibinfo{volume}{24},
  \bibinfo{pages}{900--915}.
\bibitem[{Russell and Norvig(2021)}]{russell2021artificial}
\bibinfo{author}{Russell, S.J.}, \bibinfo{author}{Norvig, P.},
  \bibinfo{year}{2021}.
\newblock \bibinfo{title}{Artificial Intelligence: A Modern Approach}.
\newblock \bibinfo{publisher}{Pearson Education, Inc.}
\bibitem[{Sanborn et~al.(2018)Sanborn, Bourgin, Chang and
  Griffiths}]{sanborn2018representational}
\bibinfo{author}{Sanborn, S.}, \bibinfo{author}{Bourgin, D.D.},
  \bibinfo{author}{Chang, M.}, \bibinfo{author}{Griffiths, T.L.},
  \bibinfo{year}{2018}.
\newblock \bibinfo{title}{Representational efficiency outweighs action
  efficiency in human program induction}, in: \bibinfo{booktitle}{Proceedings
  of the 40th Annual Conference of the Cognitive Science Society}.
\bibitem[{Schwarz(1978)}]{schwarz1978bic}
\bibinfo{author}{Schwarz, G.}, \bibinfo{year}{1978}.
\newblock \bibinfo{title}{{Estimating the Dimension of a Model}}.
\newblock \bibinfo{journal}{The Annals of Statistics} \bibinfo{volume}{6},
  \bibinfo{pages}{461--464}.
\bibitem[{{\c{S}}im{\c{s}}ek and Barto(2009)}]{simsek2009skill}
\bibinfo{author}{{\c{S}}im{\c{s}}ek, {\"O}.}, \bibinfo{author}{Barto, A.G.},
  \bibinfo{year}{2009}.
\newblock \bibinfo{title}{Skill characterization based on betweenness}, in:
  \bibinfo{booktitle}{{Advances in Neural Information Processing Systems}}.
\bibitem[{Solway et~al.(2014)Solway, Diuk, C\'{o}rdova, Yee, Barto, Niv and
  Botvinick}]{solway2014optimal}
\bibinfo{author}{Solway, A.}, \bibinfo{author}{Diuk, C.},
  \bibinfo{author}{C\'{o}rdova, N.}, \bibinfo{author}{Yee, D.},
  \bibinfo{author}{Barto, A.G.}, \bibinfo{author}{Niv, Y.},
  \bibinfo{author}{Botvinick, M.M.}, \bibinfo{year}{2014}.
\newblock \bibinfo{title}{Optimal behavioral hierarchy}.
\newblock \bibinfo{journal}{PLOS Computational Biology} \bibinfo{volume}{10},
  \bibinfo{pages}{1--10}.
\bibitem[{Stolle and Precup(2002)}]{stolle2002learning}
\bibinfo{author}{Stolle, M.}, \bibinfo{author}{Precup, D.},
  \bibinfo{year}{2002}.
\newblock \bibinfo{title}{Learning options in reinforcement learning}, in:
  \bibinfo{booktitle}{Abstraction, Reformulation, and Approximation: 5th
  International Symposium, SARA 2002 Kananaskis, Alberta, Canada August 2--4,
  2002 Proceedings 5}, \bibinfo{organization}{Springer}. pp.
  \bibinfo{pages}{212--223}.
\bibitem[{Sutton et~al.(1999)Sutton, Precup and Singh}]{sutton1999between}
\bibinfo{author}{Sutton, R.S.}, \bibinfo{author}{Precup, D.},
  \bibinfo{author}{Singh, S.}, \bibinfo{year}{1999}.
\newblock \bibinfo{title}{Between mdps and semi-mdps: A framework for temporal
  abstraction in reinforcement learning}.
\newblock \bibinfo{journal}{Artificial Intelligence} \bibinfo{volume}{112},
  \bibinfo{pages}{181--211}.
\bibitem[{Tomov et~al.(2020)Tomov, Yagati, Kumar, Yang and
  Gershman}]{tomov2020discovery}
\bibinfo{author}{Tomov, M.S.}, \bibinfo{author}{Yagati, S.},
  \bibinfo{author}{Kumar, A.}, \bibinfo{author}{Yang, W.},
  \bibinfo{author}{Gershman, S.J.}, \bibinfo{year}{2020}.
\newblock \bibinfo{title}{Discovery of hierarchical representations for
  efficient planning}.
\newblock \bibinfo{journal}{PLOS Computational Biology} \bibinfo{volume}{16},
  \bibinfo{pages}{1--42}.
\bibitem[{Tosatto et~al.(2022)Tosatto, Fagot, Nemeth and
  Rey}]{tosatto2022evolution}
\bibinfo{author}{Tosatto, L.}, \bibinfo{author}{Fagot, J.},
  \bibinfo{author}{Nemeth, D.}, \bibinfo{author}{Rey, A.},
  \bibinfo{year}{2022}.
\newblock \bibinfo{title}{The evolution of chunks in sequence learning}.
\newblock \bibinfo{journal}{Cognitive Science} \bibinfo{volume}{46},
  \bibinfo{pages}{e13124}.
\bibitem[{Toussaint and Storkey(2006)}]{toussaint2006probabilistic}
\bibinfo{author}{Toussaint, M.}, \bibinfo{author}{Storkey, A.},
  \bibinfo{year}{2006}.
\newblock \bibinfo{title}{Probabilistic inference for solving discrete and
  continuous state markov decision processes}, in:
  \bibinfo{booktitle}{Proceedings of the 23rd International Conference on
  Machine Learning}, \bibinfo{publisher}{Association for Computing Machinery},
  \bibinfo{address}{New York, NY, USA}. pp. \bibinfo{pages}{945--952}.
\bibitem[{Ullman and Wang(2023)}]{ullman2023resource}
\bibinfo{author}{Ullman, T.D.}, \bibinfo{author}{Wang, Y.},
  \bibinfo{year}{2023}.
\newblock \bibinfo{title}{Resource bounds on mental simulations: Evidence from
  a fluid-reasoning task}.
\newblock \URLprefix \url{osf.io/preprints/psyarxiv/rf367}.
\bibitem[{Verwey(1996)}]{verwey1996buffer}
\bibinfo{author}{Verwey, W.B.}, \bibinfo{year}{1996}.
\newblock \bibinfo{title}{Buffer loading and chunking in sequential
  keypressing}.
\newblock \bibinfo{journal}{Journal of experimental psychology. Human
  perception and performance} \bibinfo{volume}{22}, \bibinfo{pages}{544--562}.
\bibitem[{Wen et~al.(2020)Wen, Precup, Ibrahimi, Barreto, Van~Roy and
  Singh}]{wen2020efficiency}
\bibinfo{author}{Wen, Z.}, \bibinfo{author}{Precup, D.},
  \bibinfo{author}{Ibrahimi, M.}, \bibinfo{author}{Barreto, A.},
  \bibinfo{author}{Van~Roy, B.}, \bibinfo{author}{Singh, S.},
  \bibinfo{year}{2020}.
\newblock \bibinfo{title}{On efficiency in hierarchical reinforcement
  learning}, in: \bibinfo{editor}{Larochelle, H.}, \bibinfo{editor}{Ranzato,
  M.}, \bibinfo{editor}{Hadsell, R.}, \bibinfo{editor}{Balcan, M.},
  \bibinfo{editor}{Lin, H.} (Eds.), \bibinfo{booktitle}{{Advances in Neural
  Information Processing Systems}}, \bibinfo{publisher}{Curran Associates,
  Inc.}. pp. \bibinfo{pages}{6708--6718}.
\bibitem[{Wingate et~al.(2011)Wingate, Goodman, Roy, Kaelbling and
  Tenenbaum}]{wingate2011bayesian}
\bibinfo{author}{Wingate, D.}, \bibinfo{author}{Goodman, N.D.},
  \bibinfo{author}{Roy, D.M.}, \bibinfo{author}{Kaelbling, L.P.},
  \bibinfo{author}{Tenenbaum, J.B.}, \bibinfo{year}{2011}.
\newblock \bibinfo{title}{Bayesian policy search with policy priors}.
\newblock \bibinfo{note}{Proceedings of the Twenty-Second international joint
  conference on Artificial Intelligence}.
\bibitem[{Wu et~al.(2023)Wu, Éltető, Dasgupta and Schulz}]{wu2023chunking}
\bibinfo{author}{Wu, S.}, \bibinfo{author}{Éltető, N.},
  \bibinfo{author}{Dasgupta, I.}, \bibinfo{author}{Schulz, E.},
  \bibinfo{year}{2023}.
\newblock \bibinfo{title}{Chunking as a rational solution to the speed-accuracy
  trade-off in a serial reaction time task}.
\newblock \bibinfo{journal}{Scientific Reports} \bibinfo{volume}{13},
  \bibinfo{pages}{7680}.
\bibitem[{Zhao et~al.(2023)Zhao, Lucas and Bramley}]{zhao2023model}
\bibinfo{author}{Zhao, B.}, \bibinfo{author}{Lucas, C.G.},
  \bibinfo{author}{Bramley, N.R.}, \bibinfo{year}{2023}.
\newblock \bibinfo{title}{A model of conceptual bootstrapping in human
  cognition}.
\newblock \bibinfo{journal}{Nature Human Behaviour} , \bibinfo{pages}{1--12}.
\bibitem[{Éltető and Dayan(2023)}]{elteto2023habits}
\bibinfo{author}{Éltető, N.}, \bibinfo{author}{Dayan, P.},
  \bibinfo{year}{2023}.
\newblock \bibinfo{title}{Habits of mind: Reusing action sequences for
  efficient planning} \bibinfo{note}{ArXiv:2306.05298 [cs]}.
\bibitem[{Éltető et~al.(2022)Éltető, Nemeth, Janacsek and
  Dayan}]{elteto2022tracking}
\bibinfo{author}{Éltető, N.}, \bibinfo{author}{Nemeth, D.},
  \bibinfo{author}{Janacsek, K.}, \bibinfo{author}{Dayan, P.},
  \bibinfo{year}{2022}.
\newblock \bibinfo{title}{Tracking human skill learning with a hierarchical
  bayesian sequence model}.
\newblock \bibinfo{journal}{PLOS Computational Biology} \bibinfo{volume}{18},
  \bibinfo{pages}{1--28}.

\end{thebibliography}

\end{document}